\documentclass[sigconf]{acmart}

\AtBeginDocument{%
  }

\copyrightyear{2023}
\acmYear{2023}
\setcopyright{rightsretained}
\acmConference[IoTDI '23]{International Conference on Internet-of-Things Design and Implementation}{May 9--12, 2023}{San Antonio, TX, USA}
\acmBooktitle{International Conference on Internet-of-Things Design and Implementation (IoTDI '23), May 9--12, 2023, San Antonio, TX, USA}
\acmDOI{10.1145/3576842.3582377}
\acmISBN{979-8-4007-0037-8/23/05}

\usepackage{subcaption}
\usepackage{amsmath,amsfonts}
\usepackage{bm}
\usepackage{bbm}
\usepackage[ruled,vlined]{algorithm2e}
\usepackage{graphicx}
\usepackage{textcomp}
\usepackage{xcolor}
\usepackage[inline]{enumitem}
\usepackage{url}
\usepackage{amsthm}
\usepackage{wrapfig}
\usepackage{multicol}
\usepackage{circledsteps}

\newcommand{\method}{\textit{Async-HFL}}
\newcommand{\norm}[1]{\lVert#1\rVert}
\newcommand{\lin}[1]{\left\langle#1\right\rangle}
\newcommand{\R}{\mathbb{R}}
\newtheorem{assumption}{Assumption}
\theoremstyle{definition}

\newcommand{\E}{\mathbb{E}}
\newcommand{\cO}{\mathcal{O}}

\newcommand{\Ch}{\checkmark}
\newcommand{\X}{$\times$}
\newcommand{\cQ}{\mathcal{Q}}
\newcommand{\revise}[1]{\textcolor{black}{#1}}

\begin{document}

\title{Async-HFL: Efficient and Robust Asynchronous Federated Learning in Hierarchical IoT Networks}

\author{Xiaofan Yu}
\email{x1yu@ucsd.edu}
\orcid{0000-0002-9638-6184}
\affiliation{%
  \institution{University of California San Diego}
  \city{La Jolla}
  \state{California}
  \country{USA}
}

\author{Ludmila Cherkasova}
\email{lucy.cherkasova@gmail.com}
\orcid{0000-0002-9333-4901}
\affiliation{%
  \institution{Arm Research}
  \city{San Jose}
  \state{California}
  \country{USA}
}

\author{Harsh Vardhan}
\email{hharshvardhan@ucsd.edu}
\orcid{0000-0002-4656-3162}
\affiliation{%
  \institution{University of California San Diego}
  \city{La Jolla}
  \state{California}
  \country{USA}
}

\author{Quanling Zhao}
\email{quzhao@ucsd.edu}
\orcid{0000-0003-4699-5149}
\affiliation{%
  \institution{University of California San Diego}
  \city{La Jolla}
  \state{California}
  \country{USA}
}

\author{Emily Ekaireb}
\email{eekaireb@ucsd.edu}
\orcid{0000-0002-5090-1902}
\affiliation{%
  \institution{University of California San Diego}
  \city{La Jolla}
  \state{California}
  \country{USA}
}

\author{Xiyuan Zhang}
\email{xiyuanzh@ucsd.edu}
\orcid{0000-0002-8908-1307}
\affiliation{%
  \institution{University of California San Diego}
  \city{La Jolla}
  \state{California}
  \country{USA}
}

\author{Arya Mazumdar}
\email{arya@ucsd.edu}
\orcid{0000-0003-4605-7996}
\affiliation{%
  \institution{University of California San Diego}
  \city{La Jolla}
  \state{California}
  \country{USA}
}

\author{Tajana \v{S}imuni\'{c} Rosing}
\email{tajana@ucsd.edu}
\orcid{0000-0002-6954-997X}
\affiliation{%
  \institution{University of California San Diego}
  \city{La Jolla}
  \country{USA}
}

\renewcommand{\shortauthors}{Yu et al.}
%The canonical FL algorithm adopts synchronous model updates from distributed clients, where each round of global aggregation waits for the slowest device to return. Such a scheme suffers from long waiting time or suspension under heterogeneous network delays and unexpected failures. 

\begin{abstract}
  Federated Learning (FL) has gained increasing interest in recent years as a distributed on-device learning paradigm. However, multiple challenges remain to be addressed for deploying FL in real-world Internet-of-Things (IoT) networks with hierarchies.
  Although existing works have proposed various approaches to account data heterogeneity, system heterogeneity, unexpected stragglers and scalability, none of them provides a systematic solution to address all of the challenges in a hierarchical and unreliable IoT network.
  In this paper, we propose an asynchronous and hierarchical framework (\method) for performing FL in a common three-tier IoT network architecture. \revise{In response to the largely varied networking and system processing delays, {\method} employs asynchronous aggregations at both the gateway and cloud levels thus avoids long waiting time. To fully unleash the potential of {\method} in converging speed under system heterogeneities and stragglers, we design \textit{device selection} at the gateway level and \textit{device-gateway association}  at the cloud level. Device selection module chooses diverse and fast edge devices to trigger local training in real-time while device-gateway association module determines the efficient network topology periodically after several cloud epochs, with both modules satisfying bandwidth limitations.}
  We evaluate {\method}'s convergence speedup using large-scale simulations based on ns-3 and a network topology from NYCMesh. Our results show that {\method} converges 1.08-1.31x faster in wall-clock time and saves up to 21.6\% total communication cost compared to state-of-the-art asynchronous FL algorithms (with client selection).
  We further validate {\method} on a physical deployment and observe its  robust convergence under unexpected stragglers.
\end{abstract}

%\begin{CCSXML}
%<ccs2012>
%<concept>
%<concept_id>10010520.10010553.10003238</concept_id>
%<concept_desc>Computer systems organization~Sensor networks</concept_desc>
%<concept_significance>500</concept_significance>
%</concept>
%<concept>
%<concept_id>10010147.10010257</concept_id>
%<concept_desc>Computing methodologies~Machine learning</concept_desc>
%<concept_significance>500</concept_significance>
%</concept>
%<concept>
%<concept_id>10003033.10003083</concept_id>
%<concept_desc>Networks~Network properties</concept_desc>
%<concept_significance>300</concept_significance>
%</concept>
%</ccs2012>
%\end{CCSXML}

%\ccsdesc[300]{Networks~Network properties}

\maketitle

%%%%%%%%%%%%%%%%%%%%%%%%%%%%%%%%%%%%%%%%%%%%%%%%%%%%%%%%%%
% Introduction
%%%%%%%%%%%%%%%%%%%%%%%%%%%%%%%%%%%%%%%%%%%%%%%%%%%%%%%%%%
\section{Introduction}
\label{sec:intro}
%The number of Internet-of-Things (IoT) devices has been increasing exponentially in recent years, with an expectation of establishing 26.4 billion connections by 2026~\cite{ericsson}. 
Embedding intelligence into ubiquitous IoT devices can perform more complex tasks, thus benefiting a wide range of applications including personal healthcare~\cite{beniczky2021machine}, smart cities~\cite{jasim2021design}, and self-driving vehicles~\cite{ahmad2020design}.
%On-device distributed learning has become the trend for the next generation of IoT under rapid developments of lightweight machine learning algorithms and powerful edge-computing platforms.
%  (e.g., Raspberry Pi 4~\cite{rpi4b} and NVIDIA Jetson Nano~\cite{jetsonnano})
To enable distributed learning in a large-scale network, Federated Learning (FL) has appeared as a promising paradigm. The learning procedure begins with the central server distributing the global model to selected devices. Then each device trains with gradient descent on its local dataset and sends the updated model back to the server. Finally, the central server aggregates the received models to obtain a new global model.
%, only edge devices have access to the local dataset. Each round of the learning process consists of (1) local training on device, (2) device sending model updates to the cloud, and (3) the cloud aggregating models. 
Edge devices do not reveal the local dataset but only share the updated model.
Hence FL collaboratively learns from distributed devices while preserving users' privacy. The canonical baseline in FL is Federated Averaging (FedAvg)~\cite{mcmahan2017communication} which employs synchronous global aggregation - the central server performs aggregation after the slowest device returns, thus is impeded by unacceptable long delays or stragglers.
Recent contributions on semi-asynchronous FL~\cite{chai2020fedat,wu2020safa,zhang2021csafl,chen2022heterogeneous,nguyen2022federated} alleviate the issue by aggregating updates that arrive within a certain period and dealing with late model updates asynchronously.
However, the semi-asynchronous scheme still suffers from untimely updates with hard-to-tune waiting periods on heterogeneous and unreliable networks.
%However, multiple challenges need to be addressed for deploying FL in real world.
%While traditional methods perform training and inference on the central cloud, recent research has attempted to push the intelligence to the edge via Federated Learning (FL). With FL, only edge devices have access to the local dataset and learning is performed by sending model updates to the cloud. The FL scheme saves communication resources in data-intensive applications (e.g., video streaming) while also preserves user privacy~\cite{brik2020federated}.
%, including expensive communication, system heterogeneity, data heterogeneity and stragglers~\cite{li2020federated}.

\iffalse
\begin{figure}[tb]
  \centering
  %≈
  \includegraphics[width=0.42\textwidth]{fig/intro.png}
  \vspace{-2mm}
  \caption{\small System architecture for a common three-tier IoT networks with FL applications. The nature of heterogeneous delays and unreliable networks in hierarchy have not been systematically considered in previous works that use synchronous and semi-asynchronous schemes~\cite{lai2021oort,li2022pyramidfl,nguyen2022federated}.}
  \vspace{-4mm}
  \label{fig:intro}
\end{figure}
\fi

\begin{figure}%1   %{r}{0.54\textwidth}
\centering
%\vspace{-6mm}
\includegraphics[width=0.9\columnwidth]{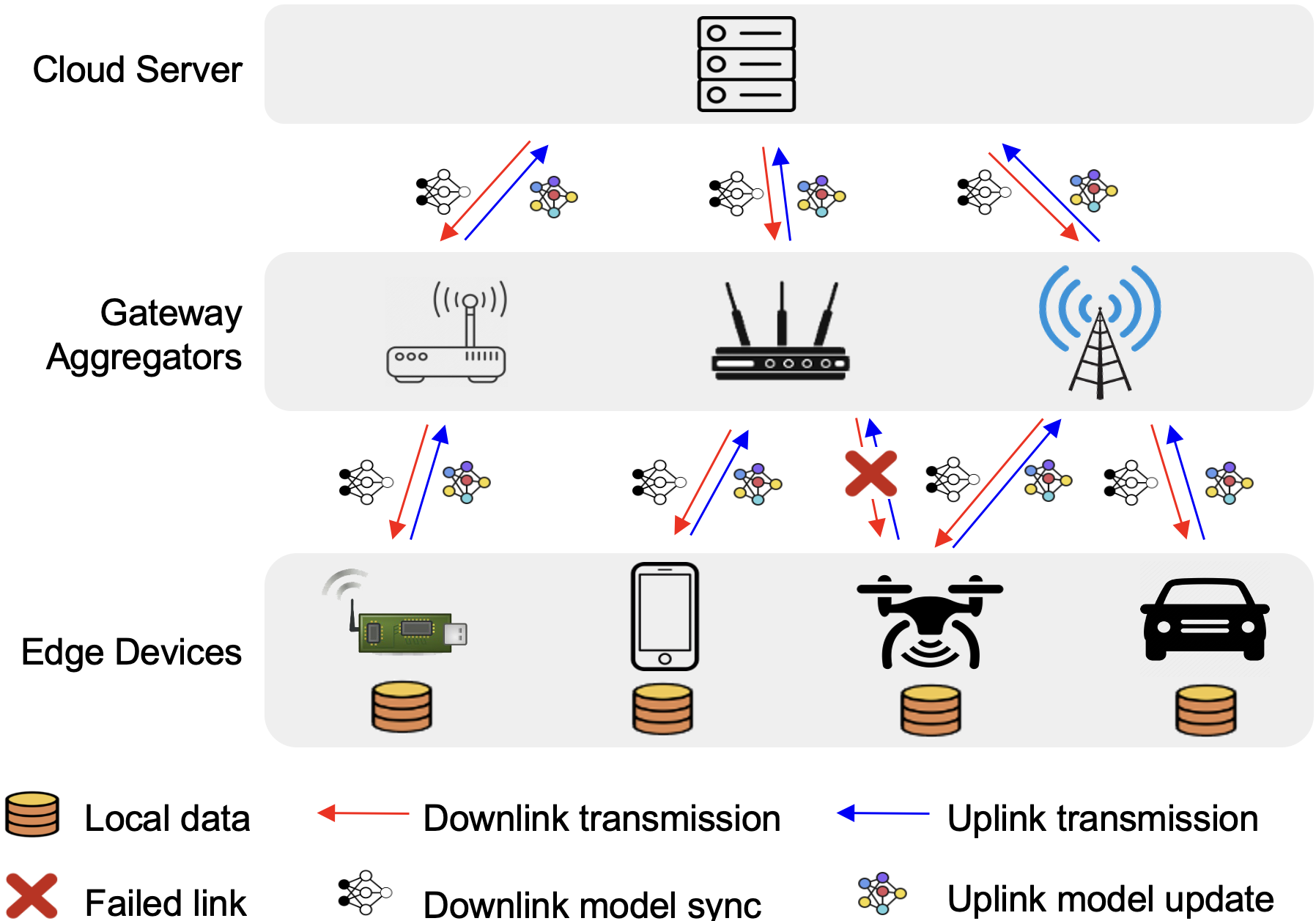}
%\vspace{-6mm}
\caption{\small System architecture for a common three-tier IoT networks running FL applications with heterogeneous delays and unreliable networks.}
\vspace{-4mm}
\label{fig:intro}
\end{figure}

For real-world IoT networks, we recognize that the diverse nature of the overall system prevents an efficient and robust FL deployment. A large number of real deployments are organized in a hierarchical manner, for example, NYCMesh~\cite{nycmesh}, HPWREN~\cite{hpwren}. % and the collection of networks in the Internet Topology Zoo~\cite{knight2011internet}. 
All of these architectures can be simplified to the three-tier structure of cloud server, gateway aggregators, and edge devices as shown in Fig.~\ref{fig:intro}.
The cloud layer offers powerful servers with abundant resources and effective processing capabilities. The gateway layer includes base stations and routers, acting as an intermediate hub connecting the cloud and edge devices. The bottom layer of edge devices refers to small mobile systems like sensors, smartphones, drones, etc., usually subject to limited resources and energy.
%The \textit{heterogeneity} of this hierarchical network structure comes from various aspects and imposes the following challenges for deploying FL on large-scale IoT networks:
%can significantly degrade the performance of Federated Learning:

Deploying FL on heterogeneous and hierarchical IoT networks faces the following challenges:
\begin{itemize} %[label=(C\arabic*)]
    \item[(C1)] \textbf{Heterogeneous data distribution}: the distribution of local data on edge devices can be largely different due to environmental variations or users' specifics. Non-independent and identically distributed (non-iid) data has been shown to slow down or prevent FL convergence without careful and tailored management~\cite{li2020fedprox,wang2020tackling,karimireddy2020scaffold}.
    \item[(C2)] \textbf{Heterogeneous system characteristics}:
    The edge devices are equipped with various CPU chips, memory storage and communication technologies.
    As a result, in Fig.~\ref{fig:intro}, the computational delay on each layer and the communication delay between two layers  can be largely different. Applying synchronous and semi-asynchronous FL are subject to longer waiting time.
    \item[(C3)] \textbf{Unexpected stragglers (or device dropout)}: Stragglers are common in every layer of IoT networks, due to energy shortage, circuit shortage or wireless interference. Without careful management, the learning procedure might be delayed or completely hang up due to stragglers.
    \item[(C4)] \textbf{Scalability}: \revise{Na\"ively extending a two-tier algorithm to hierarchical networks (3 tiers and more) can lead to significant performance degradation, e.g., unconverged model, significant communication load~\cite{liu2020client}. How to preserve the positive gains while avoiding undesired degradation during scaling to hierarchical architectures remains an active research topic.}
    %The hierarchical architecture is a common way to organize large-scale IoT networks~\cite{liu2020client,hosseinalipour2020multi}. Enabling FL on a heterogeneous and hierarchical IoT network remains an active research topic.
\end{itemize}
%IoT networks utilize a variety of physical- and MAC-layer techniques, for example, using different spectrum or collision detection strategies.
%Latest benchmarks have shown that running the same machine learning model on different off-the-shelf edge platforms can end up with two magnitude of difference in computation time~\cite{baller2021deepedgebench}.
    %On the network side, different techniques and network traffic lead to different networking properties.
%One device can connect to another gateway upon link failures as a remedy.
%\vspace{-1mm}
While previous works have studied how to improve FL convergence under one or two of data heterogeneity~\cite{li2020fedprox,wang2020tackling,karimireddy2020scaffold}, system heterogeneity~\cite{chai2020tifl,lai2021oort,li2022pyramidfl}, unexpected stragglers~\cite{mitra2021linear}, and hierarchical FL for better scalability~\cite{feng2022mobility,zhong2022flee}, none of existing work provides a \textit{systematic} solution to address all challenges in a \textit{hierarchical and unreliable} IoT network.
Our work is the \textit{first} end-to-end framework that uses (i) asynchronous and hierarchical FL algorithm and (ii) system management design to enhance efficiency and robustness, for handling all challenges (C1)-(C4).
%To the best of the authors' knowledge, the problem of \textit{how to design efficient and robust FL for large-scale IoT networks} has not been addressed. %A sophisticated design that jointly optimizes the data, system and failure perspectives is required.

%The common network structure for FL is the star topology with a central server and multiple clients directly connected to the server. In recent years researchers have shown increasing interest in running FL on more complex hierarchical networks~\cite{chai2020tifl,abad2020hierarchical,liu2020client,luo2020hfel,wang2021resource}. Related terms are invented such as Fog Learning~\cite{saha2020fogfl,zhou2020privacy,wang2021network,hosseinalipour2020federated}. In this paper, we study the key question: \textit{How to design FL for large-scale IoT networks with multiple hierarchies, heterogeneous network quality and possible failures?} We make the assumption that asynchronous FL is more appropriate for hierarchical and heterogeneous IoT networks.
%Fog Learning generalizes to a multi-layer structure to align with practical network topology.

%Arguments against decentralized, or local updates at intermediate gateways: convergence is not guaranteed.

In this paper, we propose {\method}, an asynchronous and hierarchical framework for performing FL in three-tier and unreliable IoT networks. \\
%{\method} marries asynchrony and hierarchy - the former saves the waiting 
$\bullet$ On the algorithmic side, {\method} utilizes asynchronous aggregations at both the gateway and the cloud, i.e., the aggregation is performed immediately after receiving a new updated model. 
Therefore, fast edge devices do not have to wait for the slower peers and stragglers can easily catch up after downloading the latest global model. Compared to na\"ively extending existing two-tier asynchronous FL to a three-tier hierarchy, {\method} stabilizes convergence and saves communication cost by adding an intermediate gateway aggregation layer.\\
%, plus allows flexible edge device-gateway connections.
$\bullet$ Moreover, on the system management side, we propose two modules to improve the performance of asynchronous algorithm under system heterogeneities and stragglers.
%\revise{Our asynchronous hierarchical FL algorithm is different from ...state-of-the-art asynchronous algorithm, \textit{FedAsync}~\cite{xie2019asynchronous}, by adding an intermediate gateway aggregating layer and allowing flexible device-gateway association 
%\revise{In Section~\ref{sec:motivation}, using a motivation example, we demonstrate that {\method} further accelerates convergence over the semi-asynchronous scheme under non-iid data and heterogeneous network delays.}
%\revise{We will shown in Section~\ref{sec:motivation} that the asynchronous hierarchical FL itself might lead to slow convergence without a careful managing strategy.} 
We design \textit{device selection module} at the gateway level and \textit{device-gateway association module} at the cloud level. Gateway-level device selection determines which device to trigger local training in real-time, while cloud-level device-gateway association manages network topology (i.e., which device connects to which gateway) for longer-term performance. Both modules formulate and solve Integer Linear Programs to jointly consider data heterogeneity, system characteristics, and stragglers. For data heterogeneity, we define the \textit{learning utility} metric to quantify gradient affinity and diversity of devices, inspired from online coreset selection~\cite{yoon2021online}. 
For system heterogeneity, we monitor the latencies per gateway-device link and available connections.\\
$\bullet$ {\method} is different from previous works in that (i) {\method} considers \textit{finer-grained} information of gradient diversity instead of just loss values (as in state-of-the-art asynchronous client selection algorithms~\cite{zhou2021tea}), and (ii) {\method} incorporates device selection and network topology management at various tiers, which collaboratively optimize model convergence in hierarchical and unreliable IoT networks.
To minimize communication overhead, during warm up, we collect the gradients of all devices, perform Principle Component Analysis (PCA) and distribute the PCA parameters to all devices. During training, only the principle components of gradients are exchanged. 
%The information exchange between gateways and cloud is also carefully customized to minimize overhead.

%The contributions of this paper are listed as follows:
%\begin{enumerate}[label=(\arabic*)]
%    \item We formally define {\method} and prove its convergence. Since the asynchronous scheme does not wait for slow or dropped devices, {\method} presents better communication efficiency and robustness to stragglers in heterogeneous and unstable wireless networks. 
%    \item We design a distributed and reactive client selection algorithm for {\method}.
%    \item We conduct comprehensive simulations. We use a mesh network setup based on NYCMesh.
%    \item We build a real FL system by deploying Raspberry Pi 4's and 400's in a nearby area of ..km$\times$..km.
%\end{enumerate}
%For learning models, we used the latest algorithm Ditto~\cite{li2021ditto} which enables model personalization.

\revise{In summary, the contributions of {\method} are listed as follows:
\begin{itemize}
    %\item We conduct in-depth analysis on existing synchronous and semi-asynchronous FL frameworks cannot provide fast and efficient convergence in a hierarchical and unreliable IoT network (Section~\ref{sec:motivation}). 
    %Stress the performance degradation... % Previous works focused on delay distribution in a data-center setting, which is largely different from the long-tail delay distribution in real-world wireless networks. 
    \item Recognizing the unique challenges in hierarchical and unreliable IoT networks, {\method} uses asynchronous aggregations at both the gateway and cloud levels. We formally prove the convergence of {\method} under non-iid data distribution. 
    \item To quantify data heterogeneity in a finer manner, we propose the \textit{learning utility} metric based on gradient diversity to guide decision making in {\method}.
    \item To collaboratively optimize model convergence under data heterogeneity, system heterogeneity and stragglers, {\method} incorporates distributed modules of the \textit{gateway-level device selection} and the \textit{cloud-level device-gateway association}. Communication overhead is reduced by exchanging compressed gradients from PCA.
    \item We implement and evaluate {\method}'s convergence speedup and communication saving under various network characteristics using large-scale simulations based on ns-3~\cite{ns3} and NYCMesh~\cite{nycmesh}. Our results demonstrate a speedup of 1.08-1.31x in terms of \textit{wall-clock} convergence time and total communication savings of up to 21.6\% compared to state-of-the-art asynchronous FL algorithms. We further validate {\method} on a physical deployment with Raspberry Pi 4s and CPU clusters and show robust convergence under stragglers. %Comprehensive sensitivity and overhead analyses of {\method} are performed in Section~\ref{sec:sensitivity_analysis} and Section~\ref{sec:overhead_analysis} respectively.
\end{itemize}
}

\revise{\textbf{Relationship with other FL research:} {\method} focuses on addressing system variations and potential stragglers in hierarchical IoT networks, thus is orthogonal to other FL techniques of personalization~\cite{tan2022towards}, pruning~\cite{li2021hermes} and masking~\cite{li2021fedmask}. Combining {\method} with the above-mentioned techniques is feasible but is not the focus of this paper.}

The rest of the paper is organized as follows: 
Section~\ref{sec:related-work} reviews related works. Section~\ref{sec:background} introduces background and models.
Section~\ref{sec:motivation} presents a motivating study.
Section~\ref{sec:method} expands on the details of {\method}. Section~\ref{sec:evaluation} covers the experimental setups and results.
Finally, the whole paper is concluded in Section~\ref{sec:conclusion}.

%%%%%%%%%%%%%%%%%%%%%%%%%%%%%%%%%%%%%%%%%%%%%%%%%%%%%%%%%%
% Motivation
%%%%%%%%%%%%%%%%%%%%%%%%%%%%%%%%%%%%%%%%%%%%%%%%%%%%%%%%%%
\section{Related Work}
\label{sec:related-work}
%In this section, we review state-of-the-art works addressing one or a subset of the challenges we consider, and stress the necessity of an end-to-end FL framework to cope with the nature of IoT networks.
%The heterogeneity of IoT networks bring three major challenges to FL: (C1) data heterogeneity, (C2) system heterogeneity and (C3) stragglers. We recognize scalability as another challenge (C4) on how to efficiently apply FL in large-scale networks. 

In this section, we review state-of-the-art works and summarize the existing frameworks in Table~\ref{tbl:related_works} with regard to challenges (C1)-(C4).
%summarize them existing frameworks with regard to the four challenges in Table~\ref{tbl:related_works}.

%\smallskip
\textbf{Synchronous FL.}
Based on FedAvg, a large number of works have studied synchronous FL under data and system heterogeneity from both theoretical and practical perspectives~\cite{li2020fedprox,wang2020tackling,karimireddy2020scaffold,chai2020tifl,lai2021oort,mitra2021linear}.
%\begin{enumerate*}[label=(\arabic*)]
%    \item \textit{data heterogeneity}: imbalanced and non-i.i.d. data distribution at devices~\cite{li2020fedprox,wang2020tackling,karimireddy2020scaffold},
%    \item \textit{system heterogeneity}: various local computation and communication delay~\cite{chai2020tifl,lai2021oort,mitra2021linear}, and
%    \item \textit{stragglers}: unexpected device dropout~\cite{reisizadeh2020straggler,ruan2021towards}.
%\end{enumerate*}
%All above-listed works are based on the synchronous baseline FedAvg~\cite{mcmahan2017communication} The strong synchrony of FedAvg hinders convergence and robustness if deployed on wireless networks with unstable connections.
%
%A large number of existing works carefully designed the client selection mechanism to mitigate heterogeneity, 
The client (or device) selection procedure can be carefully designed to mitigate heterogeneity by leveraging various theoretical tools~\cite{wang2020optimizing,xu2021online,xu2021dynamic,khan2020federated,ribero2020communication,balakrishnan2021diverse}.
%sequential control~\cite{wang2020optimizing}, multi-armed bandits~\cite{xu2021online}, Lyapunov optimization~\cite{xu2021dynamic}, game theory~\cite{khan2020federated}, Ornstein-Uhlenbeck Process~\cite{ribero2020communication} and submodular maximization~\cite{balakrishnan2021diverse}. 
While most works only consider data and computational delay heterogeneity, TiFL~\cite{chai2020tifl}, Oort~\cite{lai2021oort} and PyramidFL~\cite{li2022pyramidfl} brought up the communication delay variation and implemented smart client selection to balance statistical and system utilities. Nevertheless, all above works consider FL performing in data centers, while long delays and stragglers in unreliable IoT networks can lead to unsatisfied performance with synchrony.
%
%The problem of selecting subset of clients in asynchronous FL is less explored, and has largely different considerations~\cite{hu2021device}. The asynchrony requires per-client rather than subset selection.
%Most existing works approached the problem via trade-offs between learning performance and resource limitations~\cite{hao2020time,imteaj2020fedar}.
%For asynchronous FL, SAFA~\cite{wu2020safa} selects clients with lower crash probabilities to boost learning efficiency in an unreliable environment.
%TEA-fed~\cite{zhou2021tea} is a time-efficient FL protocol where idle clients actively apply for training tasks and aggregate asynchronously, but the selection proceeds in a first-come-first-serve manner until a certain fraction is reached.

%\smallskip
\textbf{Asynchronous FL.}
%A recent line of works have proposed asynchronous FL algorithms which are naturally robust to heterogeneous delays and stragglers.
In contrast to synchronous FL, the asynchronous scheme leads to faster convergence under unstable networks especially with millions of devices~\cite{huba2022papaya}.
%each client commits to the global model once it finishes local updates.
%\textit{FedAsync}~\cite{xie2019asynchronous} is the baseline asynchronous algorithm using a regularized term in the loss function and staleness-aware asynchronous weight aggregation.
%Wang \textit{et al.}~\cite{wang2021asynchronous} proposed an asynchronous algorithm that adjusts model fusion depending on staleness of wireless networks.
%ASO-Fed~\cite{chen2020asynchronous} and Wang \textit{et al.}~\cite{wang2021asynchronous} improved the baseline by integrating inter-client feature representation learning and adaptation to wireless environments.
An increasing number of asynchronous FL works have been published in recent years, with focuses on client selection~\cite{hao2020time,imteaj2020fedar,zhou2021tea,chen2021towards,zhu2022online}, weight aggregation~\cite{wang2022asynchronous,you2022triple,wang2022asynchronous} and transmission scheduling~\cite{lee2021adaptive}. Semi-asynchronous mechanisms are developed to aggregate buffered updates~\cite{chai2020fedat,wu2020safa,zhang2021csafl,chen2022heterogeneous,nguyen2022federated}.
%(FedAT~\cite{chai2020fedat}, SAFA~\cite{wu2020safa}, CSAFL~\cite{zhang2021csafl}) 
%We refer the readers to~\cite{xu2021asynchronous} for a comprehensive survey on algorithmic design in asynchronous FL. 
However, how to fully utilize the asynchronous property in hierarchical and heterogeneous IoT systems have not been addressed.

%Wang \textit{et al.}~\cite{wang2020optimizing} optimized FL with non-IID data distribution using reinforcement learning, where the reward is designed based on central test accuracy. 
%Xu \textit{et al.}~\cite{xu2021online} treated the client selection problem as a multi-arm bandit program and proposed an online $\epsilon$-greedy algorithm to balance exploration and exploitation.

%clustered all clients into logical groups based on measured latency and randomly selected clients within each group. Such strategy adaptively adjusts the trade-off between heterogeneous delays and training accuracy.

%Conventional FL designs are based on \texttt{FedAvg} and a star-topology network, where local SGD is performed on each selected client and global aggregation is made at the central server~\cite{mcmahan2017communication}. Previous authors improved centralized FL from client selection~\cite{}, heterogeneity~\cite{}, communication~\cite{} and resource management~\cite{}.
%However, centralized paradigms cannot adapt to more practical and complex topology as in modern Wireless Sensor Networks.

\begin{table}%1  %{r}{0.42\textwidth}
%\small
\tabcolsep8pt
\caption{Comparing {\method} and existing works.}
\label{tbl:related_works}
\vspace{-3mm}
%\begin{center}
\begin{tabular}{ccccc} % note: no vertical bars at all
\hline
\textbf{Method} & \multicolumn{4}{c}{\textbf{Challenges}} \\
 & (C1)  & (C2)  & (C3) & (C4) \\
\hline
 Sync FL (two tier) & \Ch & \Ch & \X & \X \\ %~\cite{chai2020tifl,balakrishnan2021diverse}
 Async FL (two tier) & \Ch & \Ch & \Ch & \X \\ % ~\cite{xie2019asynchronous,chen2020asynchronous,wang2021asynchronous}
Hier. FL ($>$two tier) & \Ch & \Ch & \X & \Ch \\ %~\cite{deng2021share,wang2021resource}
 \textbf{\method} (three tier) & \Ch & \Ch & \Ch & \Ch \\
\hline
\end{tabular}
%\end{center}
%\vspace{-4mm}
\end{table}

%\smallskip
\textbf{Hierarchical FL.} 
%A hierarchical framework is required to organize IoT systems for scalability. 
In hierarchical FL, gateways perform intermediate aggregations before sending their local models to the cloud, so that the backhaul communications between gateways and the cloud are reduced~\cite{liu2020client}.
%Extending synchronous FL to the hierarchical form, 
Multiple works have formulated client association and resource allocation problems to jointly optimize computation and communication efficiency in synchronous hierarchical FL~\cite{luo2020hfel,abad2020hierarchical,liu2020client,abdellatif2022communication}. 
Recent efforts studied mobility-aware~\cite{feng2022mobility} and dynamic hierarchical aggregations for new data~\cite{zhong2022flee}.
SHARE~\cite{deng2021share} separated the device selection and device-gateway association into two subproblems, then jointly minimized communication cost and shaped data distribution at aggregators for better global accuracy.
RFL-HA~\cite{wang2021resource} adopts synchronous aggregation within each sub-cluster and asynchronous aggregation between cluster heads (gateways) and the central cloud. All above works employ synchronous aggregation in the system, thus suffering from stragglers.

The only work that studies asynchronous and hierarchical FL is~\cite{wang2022asynchronous}. 
%where a similar algorithm to ours is presented. 
In contrast, we provide a systematic framework with additional management modules to adaptively optimize convergence under (C1)-(C4) in real-time.

\section{Background}
\label{sec:background}
In this section, we provide background on learning model (Section~\ref{sec:learning-model}), system model (Section~\ref{sec:system-model}) and common techniques in existing two-tier asynchronous FL (Section~\ref{sec:fedasync}).
To help the readers get familiar with the notations, we present the list of notations in Table~\ref{tbl:notation}. 
We also depict an example deployment in Fig.~\ref{fig:system_notation} which will be referred to as we introduce the models.

\begin{table*}[t]%2
\caption{List of important notations in problem formulation.}
\label{tbl:notation}
\vspace{-2mm}
%\begin{center}
\begin{tabular}{p{3em} p{19em} | p{3em} p{19em} } 
\toprule
Symbol & Meaning & Symbol & Meaning \\
\midrule
%$[n]$ & Set of integers \{ 1,...,n \} \\
$\mathcal{C}$ & Central cloud & 
$\mathcal{G}$ & Set of $G$ gateways \\
$\mathcal{N}$ & Set of $N$ sensor nodes & 
$\bm{J}_t$ & Feasible gateway-sensor links at time $t$ \\
$\bm{I}_t$ & Connected gateway-sensor links at time $t$ &
$\tau_i^C$ & Computational delay on device $i$ \\
$\tau_{ji}^D, \tau_{ij}^U$ & Downlink and uplink delays between $j$ and $i$ &
$\tau_{ij}$ & Gateway round latency between $j$ and $i$ \\
$R_{ij}$ & Average data rate on link $i,j$ &
$R_j$ & Data rate of all selected devices at gateway $j$ \\
$B_j$ & Bandwidth limitation at gateway $j$ &
$\ell(\bm{\omega}; x, y)$ & Loss function defined on $\bm{\omega}$ and $(x,y)$ \\
$L^i(\bm{\omega})$ & Empirical loss function at device $i$ &
$L_N(\bm{\omega})$ & Empirical loss function at central cloud \\
%$f_i$ & CPU frequency of device $i$ \\
$\mathcal{D}^i$ & Data distribution at sensor node $i$ &
$n_i$ & Number of samples at sensor node $i$  \\
$\bm{\omega}_h$ & Global model weights after $h$ cloud epochs &
$\gamma$ & Learning rate at sensor nodes \\
$\bm{\omega}_{\tau, z}^j$ & Gateway model at gateway $j$ downloaded from cloud at $\tau$ and aggregated after $z$ gateway epochs &
$\bm{\omega}_{\tau, \zeta, e}^i$ & Sensor model at sensor node $i$ downloaded from gateway at $\zeta$, which the gateway downloaded from the cloud at $\tau$ and aggregated after $e$ device epochs \\
%  $\omega_{\zeta, h}^i$ & Sensor model at sensor node $i$ downloaded from gateway at $\zeta$ and aggregated after $h$ gateway epochs \\
$H, Z, E$ & Number of cloud, gateway and device epochs &
$\rho$ & Regularization weight in async FL algorithm \\
$\alpha, \beta$ & Exponential decay factor at cloud and gateways &
$s(\cdot)$ & Staleness function \\
%$e_i, e_{th}$ & Remaining energy on device $i$ and energy threshold \\
$u_i, \eta_i, v_i$ & Learning utility, gradients affinity and diversity metric for device $i$ &
$\kappa, \phi$ & Hyperparameter in device selection and association \\
%$\phi$ & Hyperparameter in device-gateway association \\
\bottomrule
\end{tabular}
%\end{center}
%\vspace{-6mm}
\end{table*}

\subsection{Learning Model}
\label{sec:learning-model}

Suppose each sensor device $i$ collects data points $(x^i, y^i) \sim \mathcal{D}^i$ where $\mathcal{D}^i$ indicates the underlying data distribution at $i$. We assume all distributions are drawn from the same domain. 
Non-iid data distribution happens when $\mathcal{D}^i \neq \mathcal{D}^j, \forall i \neq j$. 
In practice, the underlying distribution $\mathcal{D}^i$ is unknown and we only have access to a finite number of $n_{i}$ samples at each device. Note, that $n_{i}$ can be different on various devices and at various time while we omit the time subscript for simplicity.
\revise{We aim for the typical goal of FL: to learn a uniform model $\bm{\omega} \in \R^d$ to be deployed on all distributed devices. Combining with personalized models is also feasible but exceeds the scope of this paper.}
The loss function $\ell(\bm{\omega}; x^i, y^i)$ is defined as an error function of how well the model $\bm{\omega}$ performs with respect to sample $(x^i, y^i)$. 
%The objective is to find the optimal global model that minimizes the aggregated population loss of all $N$ nodes:
%\begin{equation}
%    \min_{\bm{\omega} \in \R^d} L(\bm{\omega}) = \frac{1}{N} \sum_{i=1}^{N} \mathbb{E}_{(x^i, y^i) \sim \mathcal{D}_i} \ell(\bm{\omega}; x^i, y^i),
%    \label{eq:expected_loss}
%\end{equation}
%where the expected loss of $i$ is defined on its local distribution $\mathcal{D}_i$. 
We settle for minimizing the empirical risk minimization problem (ERM) over all devices as follows:
\begin{equation}
    \min_{\bm{\omega} \in \R^d} L_N(\bm{\omega}) = \frac{1}{N} \sum_{i=1}^{N} L^i(\bm{\omega}) = \frac{1}{N} \sum_{i=1}^{N} \frac{1}{n_{i}} \sum_{k=1}^{n_{i}} \ell(\bm{\omega}; x_k^i, y_k^i),
    \label{eq:global_loss}
\end{equation}
%which is a surrogate for the expected risk minimization in Equation~\eqref{eq:expected_loss}. 
Here $L_N(\bm{\omega})$ is the global loss function at the central cloud, and $L^i(\bm{\omega})$ is the loss function at sensor device $i$.

\subsection{System Model}
\label{sec:system-model}
%To account for all heterogeneities from (C1) to (C3), we carefully build the system and learning models as a fundamental of our design. 
%Despite of the various heterogeneities in IoT systems, none of existing work has addressed them systematically for FL applications.

\textbf{Network topology.} In a hierarchical IoT network, suppose $\mathcal{C}$ denotes the central cloud. Let $[n]$ be a set of integers $\left \{ 1,...,n \right \}$. We define $\mathcal{G} = [G]$ as a set of $G$ gateways (or base stations) and $\mathcal{N}=[N]$ as a large set of $N$ static deployed sensor devices. 
%The gateways collect intermediate information from the sensor devices and combine them before forwarding to the cloud.
While all gateways should have feasible paths to the central cloud, there might be multiple gateways that are reachable from one sensor device. Reachability can be limited due to multiple reasons such as range limitation of wireless technology, failed sensor device or network backbones.
%\begin{enumerate*}[label=(\roman*)]
%    \item range limitation of wireless technology,
%    \item failed sensor device, and
%    \item inaccessible networks.
%\end{enumerate*}
We combine all above factors into one notation, matrix $\mathbf{J}_{t} \in \mathbb{Z}^{N \times G}$, which indicates the feasible sensor-gateway pairs at wall-clock time $t$:
\begin{equation}
    \mathbf{J}_{t,ij} = \left\{
    \begin{array}{ll}
    1 & \textrm{if sensor } i \textrm{ and gateway } j \textrm{ are connectable } \textrm{at time } t \\
    0 & \textrm{otherwise}.
    \end{array}\right. 
    \label{eq:reachability}
\end{equation}
During training, a sensor is associated with only one gateway at one time. The gateway triggers local training on the device, and the device needs to upload the returned model to the same gateway. But sensors can switch to another gateway between aggregations.
We use another matrix notation $\mathbf{I}_t \in \mathbb{Z}^{N \times G}$ with the same shape as $\mathbf{J}_{t}$ as decision variables for real-time sensor-gateway connections:
\begin{equation}
    \mathbf{I}_{t,ij} = \left\{
    \begin{array}{ll}
    1 & \textrm{if sensor } i \textrm{ is connected to gateway } j \textrm{ at time } t \\ 
    0 & \textrm{otherwise}.
    \end{array}\right. 
\end{equation}
For example, in Fig.~\ref{fig:system_notation}, sensor device 2 can reach both gateway 1 and 2, but communicates with gateway 1 in the current round. In this case, $\mathbf{J}_{t,21}=\mathbf{J}_{t,22}=1, \mathbf{I}_{t,21}=1, \mathbf{I}_{t,22}=0$.

\iffalse
\begin{figure}[t]
  \centering
  %\vspace{-2mm}
 \includegraphics[width=0.4\textwidth]{fig/system_notations.png}
  \vspace{-4mm}
  \caption{\small An example of hierarchical FL deployment.}
  \vspace{-4mm}
  \label{fig:system_notation}
\end{figure}
\fi

\begin{figure}%2   %{r}{0.42\textwidth}
%\centering
%\vspace{-4mm}
\includegraphics[width=0.9\columnwidth]{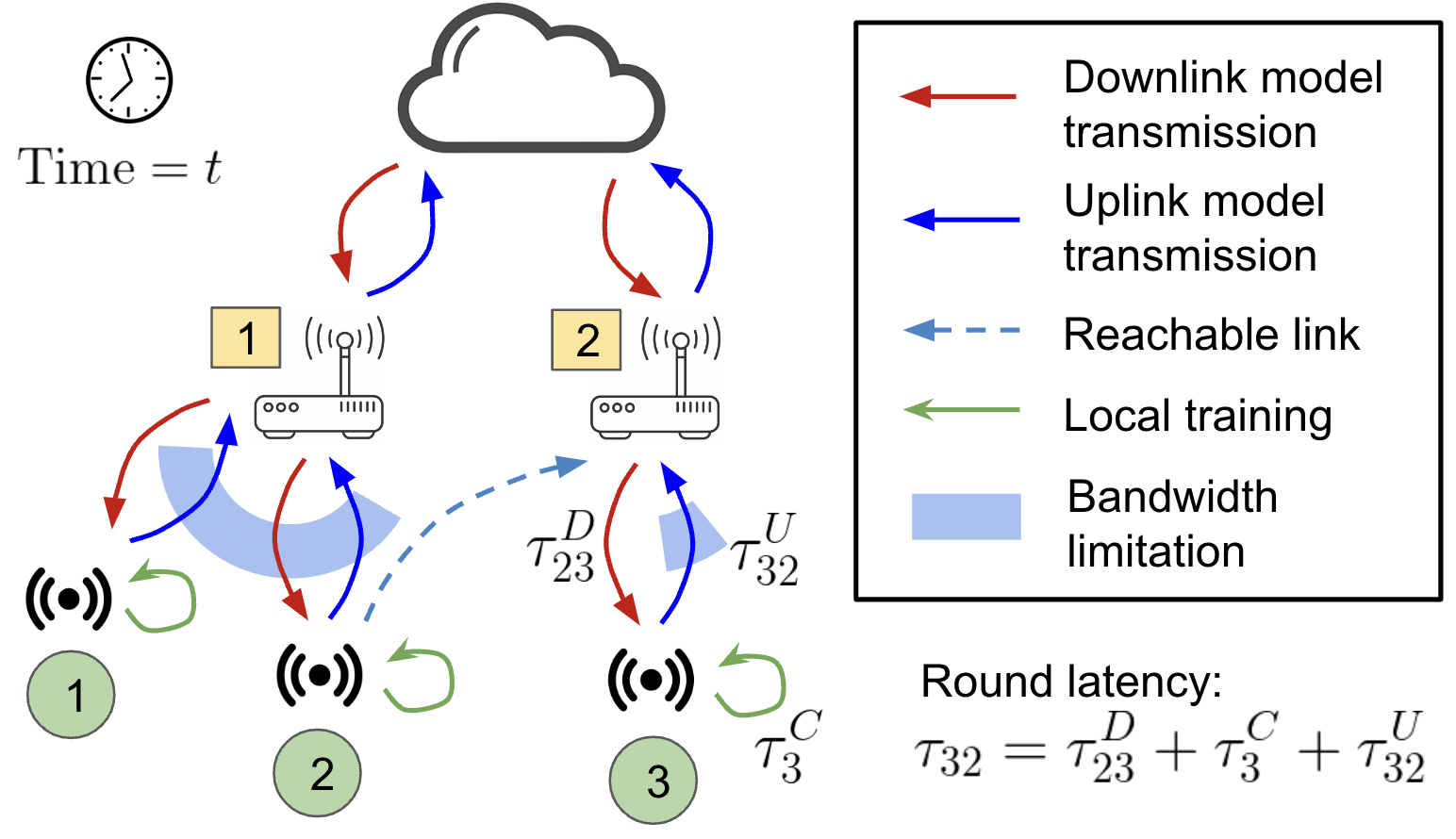}
%\vspace{-8mm}
\caption{\small An example of hierarchical FL deployment.}
\vspace{-2mm}
\label{fig:system_notation}
\end{figure}

\textbf{Computational and Communication Models.} %Each device in our model can have different computational and communicational capabilities which result in various delay. 
%Previous works have shown that computational delays can be modeled with parameters of the CPU frequency and the number of local epochs, etc.~\cite{chen2020convergence,luo2020hfel}
%we do not adopt the specific models to accommodate heterogeneous technologies.
%In contrast to previous works that use parameterized CPU models~\cite{chen2020convergence,luo2020hfel},  we measure the delays at runtime to keep the framework reactive. The average computational delay on device $i$ is denoted by $\tau_i^C$.
Computational and communication delays play the major role as system heterogeneities.
In this paper, we adopt general models while more specific parameterized CPU or network models (such as~\cite{chen2020convergence,luo2020hfel}) can be applied when more information is given.
%, and aggregation itself is less intensive.
%Similar as in~\cite{chen2020convergence,luo2020hfel}, we build computational delay model considering both system and FL parameters. The computational delay $\tau_i^C$ denotes the latency to complete $E$ local epochs at sensor device $i$: 
%\begin{equation}
%    \tau_i^C = a \cdot \frac{M \cdot n_i \cdot E}{f_i} + b,
%\end{equation}
%where $M$ is the number of multiply–accumulate (MAC) operations for one sample per epoch including forward and back propagation. Thus $M \cdot n_i \cdot E$ computes the total number of operations in one local round. $f_i$ is the CPU frequency of $i$. $a,b$ are device-specific constants derived by performing linear regression on real measurements. We fit the parameters for two typical edge computing platforms, Raspberry Pi 4~\cite{rpi4b} and 400~\cite{rpi400}. 
%For battery-powered edge device $i$, we denote its remaining energy as $e_i$.
%$P_i^C$ denotes the average power consumption during local training.
%\begin{subequations}
%\begin{align}
%    \tau_i^C = a_\tau \cdot \frac{M \cdot s_i \cdot E}{f_i} + b_\tau,
%    \label{eq:comp_delay} \\
%    P_i^C = a_P \cdot \frac{M \cdot s_i \cdot E}{f_i} + b_P,
%    \label{eq:comp_power}
%\end{align}
%\end{subequations}
%\textbf{Communication Model.}  While there exists specific delay models given MAC protocols (e.g., OFDMA~\cite{chen2020convergence,luo2020hfel}),
%
We are more interested in the communication delays happened on the last-hop links, where bandwidth is more limited and the transmitters (sensor devices) enjoy lower transmission power.
%We define $\tau_{j}^{c,D}$ as the variable for the downlink delay to transmit a FL model from the cloud to $j$, and $\tau_{j}^{c,U}$ as the variable for the uplink delay. 
During runtime, we measure \textit{round latency} $\tau_{ij}$ as the time to complete one gateway round between gateway $j$ and sensor $i$, which has three segments: (i) $\tau_{ji}^D$, the downlink delay to transmit a model from gateway $j$ to device $i$, (ii) $\tau_i^C$, the computational delay of device $i$ to perform local training and (iii) $\tau_{ij}^{U}$, the uplink delay to transmit the updated model in a reverse direction. In Fig.~\ref{fig:system_notation}, the downlink and uplink transmissions are represented with the red and blue arrows, while local training uses green arrow. The round latency is simply the sum of red, green and blue arrows: $\tau_{ij} = \tau_{ji}^D + \tau_i^C + \tau_{ij}^U$. 
%\begin{equation}
%    \tau_{ij} = \tau_{ji}^D + \tau_i^C + \tau_{ij}^U,
%    \label{eq:round_latency}
%\end{equation}
The computational delay at gateways and cloud for aggregation are neglected since gateways and cloud generally have more computational resource. 

\begin{figure*}[!tbp] %F3
\centering
\setlength{\tabcolsep}{0.2pt}
\begin{tabular}{cccc}    
    \includegraphics[width=0.23\textwidth, height=3cm]{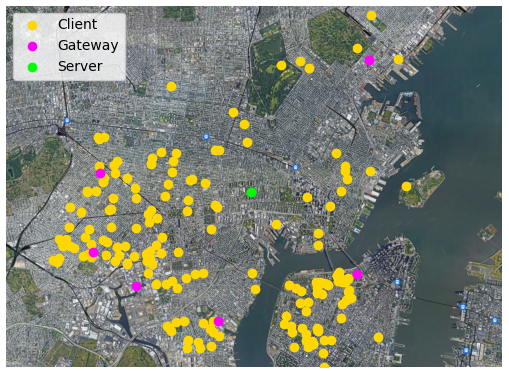} 
    &
    \includegraphics[width=0.23\textwidth, height=3cm]{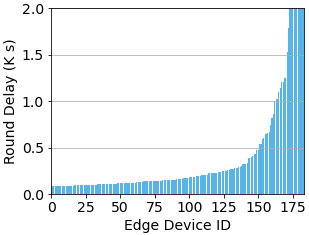}
    &
    \includegraphics[width=0.23\textwidth, height=3cm]{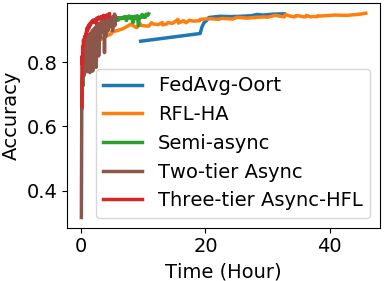}
    &
    \includegraphics[width=0.23\textwidth, height=3cm]{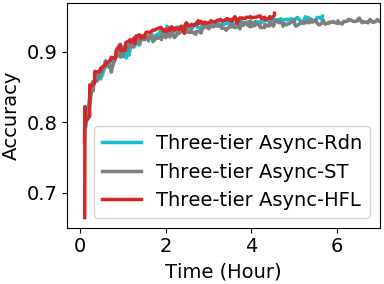}
\end{tabular}
\vspace{-4mm}
\caption{\small Left: NYCMesh topology. Second left: The round delay distribution of all edge devices. Right: Convergence performance under wall-clock time in the NYCMesh motivating study.}
\label{fig:motivation}
\vspace{-4mm}
\end{figure*}

\iffalse
\begin{figure}[!tbp]
\centering
%\vspace{-2mm}
\begin{subfigure}[t]{0.2\textwidth}
    \centering
    \includegraphics[width=3.2cm, height=2.5cm]{fig/loc2.png}
\end{subfigure}
%\vspace{-2mm}
\begin{subfigure}[t]{0.2\textwidth}
    \centering
    \includegraphics[0.2\textwidth, height=2.5cm]{fig/delays.png}
\end{subfigure}
%\vspace{-2mm}
\begin{subfigure}[t]{0.2\textwidth}
    \centering
    \includegraphics[0.2\textwidth, height=2.5cm]{fig/motivation_nycmesh_1.png}
\end{subfigure}
%\vspace{-2mm}
\begin{subfigure}[t]{0.2\textwidth}
    \centering
    \includegraphics[0.2\textwidth, height=2.5cm]{fig/motivation_nycmesh_2.png}
\end{subfigure}
\vspace{-4mm}
\caption{\small Convergence performance under wall-clock time in the NYCMesh motivating study.} %, from which we randomly select a subset to align with the number of gateways and devices in Table~\ref{tbl:datasets}.}
\label{fig:moti_results}
\vspace{-4mm}
\end{figure}
\fi

\textbf{Bandwidth Limitation.}
Bandwidth limitation places an upper bound on the throughput or data rate.
%The sensor-gateway uplink is more restricted since the sensor devices have lower transmission power.
Different from the synchronous design, we cannot accurately model the throughput at $t$ with asynchrony. % since we are not aware which device will trigger the transmission at any specific time. 
Therefore, given round latency $\tau_{ij}$ and a FL model with size $M$, we estimate the average data rate on link $i,j$ as $R_{ij} = M / \tau_{ij}$.
The total data rate of all selected devices at gateway $j$ can be computed as $R_j = \sum_{i=1}^N \bm{I}_{ij} R_{ij}$.
$B_j$ is an upper bound of \textit{average} data rate on last-hop links at gateway $j$. In Fig.~\ref{fig:system_notation}, devices 1 and 2 are subject to $B_1$, which is depicted as a blue arc.

\subsection{Two-Tier Asynchronous FL}
\label{sec:fedasync}

In asynchronous FL, each device downloads the latest global model from the cloud, runs local training, and uploads the model to the cloud where asynchronous aggregation is performed immediately. The latest asynchronous FL algorithms~\cite{xie2019asynchronous,chen2021towards} employ two common techniques as follows.
%Technical-wise, asynchronous FL faces converging challenges because the aggregation is performed per upload rather than waiting and averaging all model updates as in FedAvg. 

Firstly, in addition to the original loss term $L^i(\bm{\omega}^i)$, a \textbf{regularized loss term} penalizing the difference between current model weights $\bm{\omega}^i$ and the downloaded global model $\bm{\omega}_\tau$ is appended on device $i$:
\begin{equation}
    g^i(\bm{\omega}^i;{\bm{\omega}_\tau}) = L^i(\bm{\omega}^i) + \frac{\rho}{2} \norm{\bm{\omega}^i - \bm{\omega}_\tau}^2.
    \label{eq:loss_device}
\end{equation}
%Penalizing large changes to the global model enhances convergence.
Here $\rho$ is the regularization weight.
    
Secondly, the algorithm performs \textbf{staleness-aware weight aggregation} at the cloud:
\begin{subequations}
    \begin{align}
    \alpha_h &\gets \alpha \cdot s(h-\tau) \\
    \bm{\omega}_h &\gets (1-\alpha_h)\bm{\omega}_{h-1} + \alpha_h \bm{\omega}_{new},
    \end{align}
\end{subequations}
where $\omega_{new}$ is the newly received model weights, $h$ is the current cloud epoch and $\alpha_h$ is the staleness-aware weight calculated by multiplying a constant $\alpha$ with the staleness function $s(h-\tau)$.
Staleness refers to the difference in \textit{the number of epochs} since its last global update. For example, $h$ is the current global aggregating epoch while $\tau$ is the global epoch when the model is downloaded. Intuitively, larger staleness means the model is more outdated and thus should be given less importance.
Staleness-aware aggregation simulates an averaging process without synchrony.
%The staleness is a mature concept in machine learning community~\cite{damaskinos2018asynchronous,xie2019asynchronous}, and is completely different from the wall-clock time $t$.
The staleness function $s(h-\tau)$ determines the exponential decay factor during model aggregation. We adopt the polynomial staleness function $s_q(h-\tau) = (h-\tau+1)^{-q}$ parameterized by $q > 0$ as in~\cite{xie2019asynchronous}.

\section{A Motivating Study}
\label{sec:motivation}
%\revise{TODO: add experiments comparing FedAvg-oort, RFL-HA?, semi-async, FedAsync, random-async, short latency-async and async-HFL. Use a fixed number of selected clients, and show that client selection is important.}

%1. delay distributions of data centers vs. real iot networks
%2. comparison of sync, semi-async vs fully async
%3. the importance of client selection
\revise{In this section, we conduct a motivating study of existing FL frameworks under hierarchical and unreliable networks, justifying the design of {\method} on both algorithmic and management aspects. 
While recent works have noticed the importance of accounting the latency factor during client selection~\cite{chai2020tifl,lai2021oort,li2022pyramidfl}, they only considered the delay distribution in a \textit{data-center} setting which is significantly different from the ones in \textit{real-world wireless networks}.
Real-world measurements have shown that wireless networks follow the long-tail delay distribution and are highly unpredictable~\cite{sui2016characterizing}. 
We implement the FL frameworks based on ns3-fl~\cite{ekaireb2022ns3fl}
and extract the three-tier topology from the installed node locations in NYCMesh~\cite{nycmesh} as depicted in Fig.~\ref{fig:motivation} (left). We assume that edge devices are connected to the gateways via Wi-Fi, and the gateways are connected to the server via Ethernet. For each node, we retrieve its latitude, longitude, and height as input to the \texttt{HybridBuildingsPropagationLossModel} in ns-3 to obtain the average point-to-point latency. 
To include network uncertainties, we add a log-normal delay on top of the mean latency at each local training round.
The delay distribution of all edge devices (assuming all devices are selected) in one training round is shown in Fig.~\ref{fig:motivation} (second left). The simulated network delays mimic the measurement results in~\cite{sui2016characterizing}.
We use the human activity recognition dataset~\cite{anguita2013public}, assigning the data collected from one individual to one device thus presenting naturally non-iid data. The upper bound on the bandwidth of the gateways is set to 20KB/s for all experiments.}

In such setting, we experiment the performance of (i) \textbf{FedAvg under Oort}, the state-of-the-art latency-aware client selection algorithm~\cite{lai2021oort}, (ii) \textbf{RFL-HA}~\cite{wang2021resource} with asynchronous cloud aggregation and synchronous gateway aggregation, (iii) \textbf{Semi-async}, with synchronous cloud aggregation and semi-asynchronous gateway aggregation as in~\cite{nguyen2022federated}, (iv) \textbf{two-tier Async FL} which na\"ively extends the two-tier asynchronous algorithm~\cite{xie2019asynchronous,chen2021towards} to three-tier by letting the gateway just forward data, (v) \textbf{three-tier Async-HFL} with intermediate gateway aggregation proposed in our paper. 

\begin{table}%T3   %{r}{0.47\textwidth}
\tabcolsep9pt
%\small
%\vspace{-4mm}
\caption{\small Total communicated data size ratio (to Async-HFL) before reaching 95\% test accuracy in the motivating study.}
\label{tbl:moti_total_comm_size}
\vspace{-3mm}
%\begin{center}
\begin{tabular}{cccc} % note: no vertical bars at all
 \toprule % not \hline
 \textbf{\small Sync-Oort} & \textbf{\small RFL-HA} & \textbf{\small Semi-async} & \textbf{\small Two-tier Async} \\ 
 \midrule
 0.79x & 1.42x & 1.66x & 1.30x \\
 \bottomrule % not \hline
\end{tabular}
%\end{center}
\vspace{-4mm}
\end{table}

We report the wall-clock time convergence in Fig.~\ref{fig:motivation} (right) and the total size of the data communicated in ratio to Async-HFL in Table~\ref{tbl:moti_total_comm_size}.
The convergence of FedAvg and RFL-HA are significantly slowed down even with the state-of-the-art client selection. For Semi-async, we test the waiting period of 50, 100, 150 seconds and select the best results. Semi-async accelerates convergence but still takes 0.39x longer to reach the same 95\% accuracy than the fully asynchronous methods. Noticeably, compared to two-tier async, our three-tier Async-HFL achieves more stable and slightly faster convergence, while saving 0.3x communication load (equivalent to 426MB or 1498 multilayer perceptron models). This gain comes from the intermediate gateway aggregation. Introducing an additional ``averaging" step does not only smooth out the curve but avoids unnecessary back-and-forth transmission.

Aside from algorithmic design, framework management is also critical in hierarchical FL. We experiment with random (\textbf{Async-Random}), short-latency-first (\textbf{Async-ST}) gateway-level device selection and \textbf{Async-HFL} with full management. The convergence results are shown in Fig.~\ref{fig:motivation} (right). With carefully designed modules, Async-HFL converges 1.24x faster than the random gateway-level device selection. Yet, a poor device selection like Async-ST that ignores data heterogeneity can lead to a 2.13x slower speed to reach 95\% accuracy or even an unconverged model in the worst case.

\section{{\method} Design}
\label{sec:method}

\subsection{Overview}

\begin{figure}%F4 %{r}{0.45\textwidth}
%  \centering
%  \vspace{-4mm}
  \includegraphics[width=0.45\textwidth]{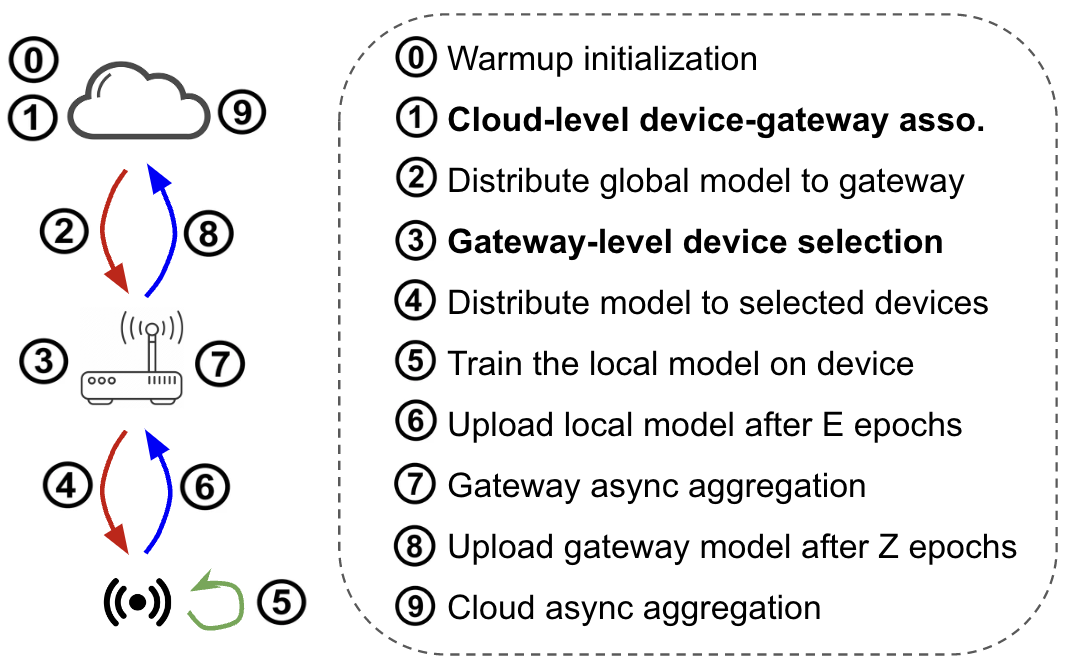}
%  \vspace{-8mm}
  \caption{\small The step-by-step procedure of {\method} in one branch of the hierarchical network.}
%  \vspace{-6mm}
  \label{fig:steps}
\end{figure}

To address all challenges (C1)-(C4) systematically, we propose an end-to-end framework {\method} for efficient and robust FL especially in hierarchical and unreliable IoT networks. The \textit{major differences} between {\method} and previous frameworks are the following: {\bf (i)} {\method} quantifies non-iid data distribution by \textit{learning utility}, which is a metric based on gradient diversity, {\bf (ii)} {\method} incorporates strategic management components, the \textbf{cloud-level device-gateway association} (\textcircled{1} in Fig.~\ref{fig:steps}) and the \textbf{gateway-level device selection} (\textcircled{3} in Fig.~\ref{fig:steps}), which are critical in jointly speeding up practical convergence under heterogeneous data and system characteristics.

Fig.~\ref{fig:steps} depicts the step-by-step procedure of {\method} in one round of cloud aggregation.
For simplicity, we only show one branch of the hierarchical network.
After the warmup initialization, we start from the cloud determining the low-level network topology, namely device-gateway association (\textcircled{1}), and then distributing the latest global model to all gateways (\textcircled{2}). Next, the gateway distributes the model to devices selected by the gateway (\textcircled{3}-\textcircled{4}). On the device, local training is performed for $E$ epochs and an updated model is returned to the gateway (\textcircled{5}-\textcircled{6}). The gateway then integrates the newly received model immediately using asynchronous aggregation (\textcircled{7}). After $Z$ gateway updates, the current gateway model is uploaded to the cloud for global asynchronous aggregation (\textcircled{8}-\textcircled{9}).

We provide more details on the design of {\method} in this section. Section~\ref{sec:algorithm} presents the detailed asynchronous hierarchical algorithm with convergence proof. Section~\ref{sec:learning_utility} presents the definition of the \textit{learning utility} metric to quantify gradient diversity. Finally, Section~\ref{sec:cs_n_ca} reveals the concrete design of device selection and device-gateway association modules.
%Namely, the aggregation is performed every time an updated model is received rather than after all selected devices have returned. With {\method}, the negative effects of long-latency devices and stragglers are minimized.

%In this section, we first give a high-level overview of our proposed framework {\method} and then explain each component in more details.
%{\method} is designed around three major components:
%\begin{enumerate*}[label=(\roman*)]
%    \item asynchronous hierarchical FL algorithm
%    \item \textit{gateway-level} device selection, and
%    \item \textit{cloud-level} device-gateway association.
%\end{enumerate*}
%Figure~\ref{fig:framwork} shows a high-level overview of the proposed framework.
\iffalse
\begin{figure}[h]
  \centering
  \vspace{-2mm}
  \includegraphics[width=0.42\textwidth]{fig/steps.png}
  \vspace{-2mm}
  \caption{\small The step-by-step procedure of {\method}.}
  \vspace{-2mm}
  \label{fig:steps}
\end{figure}
\fi

\subsection{Asynchronous Hierarchical FL Algorithm}
\label{sec:algorithm}

%In the \texttt{Updater} threads, {\method} performs  For example, $z-\zeta$ in line 21 of Algorithm~\ref{alg:async_fl} is the current staleness of \texttt{GatewayUpdater}.  indicates the gateway epoch that the sensor starts local training on (initialized in line 27). 
%
%With the above two techniques, 

In {\method}, besides the asynchronous cloud aggregation, we utilize an intermediate gateway layer and apply staleness-aware asynchronous aggregation for $Z$ epochs at the gateway. Compared to having the gateways directly forwarding asynchronous model updates to the cloud, adding intermediate gateway aggregations reduces communication burden while making sure the convergence guarantees still apply after adding minimal assumptions (as detailed later). Steps \textcircled{4}-\textcircled{7} correspond to the two-tier asynchronous FL in Fig.~\ref{fig:steps}, while our hierarchical algorithm includes steps \textcircled{2}, \textcircled{4}-\textcircled{9}, spanning all three tiers in the IoT network.

%The weight update/aggregation is performed as follows on each layer:
%\begin{subequations}
%\begin{align}
%    \bm{\omega}_{\tau,\zeta,e}^i &\gets \bm{\omega}_{\tau,\zeta,e-1}^{i} - \gamma \nabla g^i(\bm{\omega}_{\tau,\zeta,e-1}^{i};{\bm{\omega}_{\tau,z}^j}), \: \forall e \in [E] \tag*{(Device)} \\
%    \bm{\omega}_{\tau,z}^{j} &\gets (1-\beta_z^j)\bm{\omega}_{\tau,z-1}^j + \beta_z^j \bm{\omega}_{new}^i, \: \forall z \in [Z] \tag*{(Gateway)} \\
%    \bm{\omega}_t &\gets (1-\alpha_h)\bm{\omega}_{h-1} + \alpha_h \bm{\omega}_{new}^j, \: \forall h \in [H] \tag*{(Cloud)} 
%    \label{eq:async_loss}\end{align}
%\end{subequations}

The concrete algorithm implementation on cloud, gateways and devices is shown in Algorithm~\ref{alg:async_fl}. The cloud and each gateway holds a \texttt{Cloud} or \texttt{Gateway} process, which completes the initialization and asynchronously triggers the \texttt{Updater} threads for aggregation.
%The \texttt{Scheduler} thread triggers the corresponding process on the selected devices in lower layers and synchronize the latest model and epoch number.
The \texttt{Updater} thread performs aggregation until the predetermined epoch number is reached.
Each sensor device runs a \texttt{Sensor} process that locally solves a regularized optimization problem using stochastic gradient descent (SGD).
While previous works have studied adaptively adjusting local epochs according to computational resources~\cite{li2020fedprox}, namely trading model quality for faster return, such gain in {\method} might be trivial due to asynchronous aggregation and longer as well as unexpected network delays. Hence {\method} uses a fixed number of $E$ and $Z$ epochs on device and gateway-level.

\begin{algorithm}[t]
\SetKwInOut{Input}{Input}
\SetKwFunction{Cloud}{Cloud}
\SetKwFunction{CloudUpdater}{CloudUpdater}
\SetKwFunction{Gateway}{Gateway}
\SetKwFunction{GatewayUpdater}{GatewayUpdater}
\SetKwFunction{Sensor}{Sensor}
\SetKwProg{thread}{Thread}{}{}
\SetKwProg{process}{Process}{}{}
\caption{Asynchronous Hierarchical FL}
\label{alg:async_fl}
%\vspace{-5mm}
%\begin{multicols}{2}
  \DontPrintSemicolon\LinesNumbered
  \Input{$\mathcal{C}, \mathcal{G}, \mathcal{N}, H, Z, E, \alpha, \beta, s(\cdot), g^i(\cdot), \bm{\omega}_0$}
  \process{\Cloud{}}{
      % Initialize $\bm{\omega}_0$ \\
      Send $(\bm{\omega}_0,0)$ to all gateways $j\in \mathcal{G}$\\
      % Run \CloudScheduler{}, \CloudUpdater{} threads asynchronously in parallel
      Run \CloudUpdater{} asynchronously in parallel
  }
  % \thread{\CloudScheduler{}}{
  %     Periodically trigger \GatewayServer{$j$} on all gateways $\forall j \in \mathcal{G}$ and send $(\bm{\omega}_h,h)$
  % }
  \thread{\CloudUpdater{}}{
      \For{cloud epoch $h \in [H]$}{
          \If{receive $(\bm{\omega}_{new}^j, \tau)$ from gateway $j$}{
              %\textbf{Optional:} $\alpha_h \gets \alpha \times s(h-\tau)$ \\
              $\bm{\omega}_h \gets (1-\alpha_h)\bm{\omega}_{h-1} + \alpha \times s(h-\tau) \bm{\omega}_{new}^j$\\
              Send $(\bm{\omega}_h,h)$ to all gateways $j\in \mathcal{G}$
          }
      }
  }
  \process{\Gateway{$j$}}{
      \If{triggered by \Cloud{}}{
          Receive global model and timestamp $(\bm{\omega}_h, h)$\\
          Update $\tau \gets h, \bm{\omega}_{\tau,0}^{j} \gets \bm{\omega}_h$ \\
          Send $(\bm{\omega}_{\tau,z}^j,z)$ to selected sensors $i$ connected to $j$\\
  %        Run \GatewayScheduler{$j$}, \GatewayUpdater{$j$} threads asynchronously in parallel
          Run \GatewayUpdater{$j$} asynchronously in parallel
      }
  }
  % \thread{\GatewayScheduler{$j$}}{
  %     Periodically trigger \Sensor{$i$} on selected device $i$ connected to $j$ and send $(\bm{\omega}_{\tau,z}^j,z)$
  % }
  \thread{\GatewayUpdater{$j$}}{
      \For{gateway epoch $z \in [Z]$}{
          Receive $(\bm{\omega}_{new}^i, \zeta)$ from sensor node $i$ \\
          %\textbf{Optional:} $\beta_z^j \gets \beta \times s(z-\zeta)$ \\
          $\bm{\omega}_{\tau,z}^{j} \gets (1-\beta_z^j)\bm{\omega}_{\tau,z-1}^j + \beta_z^j \bm{\omega}_{new}^i$\\
          Send $(\bm{\omega}_{\tau,z}^j,z)$ to all sensor $i$ connected to $j$
      }
      Upload $(\bm{\omega}_{\tau, Z}^{j}, \tau)$ to cloud
  }
  \process{\Sensor{$i$}}{
      \If{triggered by \Gateway{$j$}}{
          Receive $(\bm{\omega}_{\tau,z}^j, z)$ from gateway $j$\\
          Update $\zeta \gets z, \bm{\omega}_{\tau,\zeta,0}^{i} \gets \bm{\omega}_{\tau,z}^{j}$\\
          \For{device epoch $e \in [E]$}{
              $\bm{\omega}_{\tau,\zeta,e}^i \gets \bm{\omega}_{\tau,\zeta,e-1}^{i} - \gamma \nabla g^i(\bm{\omega}_{\tau,\zeta,e-1}^{i};{\bm{\omega}_{\tau,z}^j})$
          }
          Upload $(\bm{\omega}_{\tau,\zeta,E}^i, \zeta)$ to gateway $j$
      }
  }
%\end{multicols}
\end{algorithm}

\textbf{Convergence Analysis.}
%\label{sec:proof}
We now establish the theoretical convergence of {\method}'s algorithm. We set the staleness function $s(\cdot) = 1$ throughout this section.
We require certain regularity conditions on the loss function, namely $L$-smoothness and $\mu$-weak convexity and bounded gradients. 
Note that $\mu$-weak convexity allows us to handle non-convex loss functions ($\mu > 0$), convex functions ($\mu=0$) and strongly convex functions ($\mu<0$). We list the additional assumptions and the convergence result below\footnote{The complete proof is included in the supplementary material or can be found at \url{https://arxiv.org/abs/2301.06646}}.
%, which we define below.
%\begin{definition}[Smoothness]
%\label{def:smooth}
%$\ell$ is $L$-smooth if for $\forall x, y \in \R^d, \ell(y) - \ell(x) \leq \lin{\nabla \ell(x), y - x} + \frac{L}{2}\norm{y - x}^2$.
%\end{definition}

%\begin{definition}[Weak Convexity]
%\label{def:cvx}
%$\ell$ is $\mu$-weakly convex if for $\forall x, y \in \R^d, \ell(y) - \ell(x) \leq \lin{\nabla \ell(x), y - x} - \frac{\mu}{2}\norm{y - x}^2$. This also means that the function $g$ with $g(x)=\ell(x)+\frac{\mu}{2}\norm{x}^2$ is convex, where $\mu > 0$. 
%\end{definition}

\begin{assumption}[Bounded gradients]
\label{assump:grad_bdd}
The loss function at the cloud,
$L_N$, and the regularized loss function at each device $g_i,\, \forall i \in \mathcal{N}$, have bounded gradients bounded by
\begin{align*}
    &\norm{\nabla L_N(\bm{\omega})}^2 \leq V_1 , \quad \forall \bm{\omega} \in \R^d\\
    &\norm{\nabla g^i(\bm{\omega}; \bm{\omega}')}^2 \leq V_2, \quad\forall \bm{\omega},\bm{\omega}' \in \R^d, \forall i \in \mathcal{N}.
\end{align*}
\end{assumption}

%Additionally, to handle asynchronous updates at cloud and gateway level, we need the maximum delay to be bounded.
\begin{assumption}[Bounded Delay]
\label{assump:bdd_delay}
The delays $h -\tau$ at the cloud model, and $z - \zeta$ at the gateway model are bounded
\begin{align}
    h - \tau \leq K_c,\quad\quad\quad z - \zeta \leq K_g
\end{align}
\end{assumption}

%Further, to ensure convergence with $\mu$-weak convexity, we require the regularization to be large.
\begin{assumption}[Regularization $\rho$ is large]
\label{assump:rho_large}
$\rho$ is large enough such that for some fixed constant $c>0$, $\forall \tau,\zeta >0, h \geq 1 , i\in[N]$, % j \in[N_i]
\begin{align}
-(1 + 2\rho + c)V_1 +\bigg(\rho^2  - \frac{\rho}{2}\bigg)\E[\norm{\bm{\omega}_{\tau,\zeta,h-1}^{i} - \bm{\omega}_{\tau,\zeta}^{i}}^2] \geq 0 
\end{align}
\end{assumption}

\begin{theorem}
\label{thm:async_fl_1}
For $L$-smooth and $\mu$-weakly convex loss function $\ell$, under Assumptions~\ref{assump:grad_bdd}-\ref{assump:rho_large}, with $\gamma \leq L^{-1},\alpha \leq K_c^{-3/2}$ and $\beta \leq K_g^{-3/2}$, after running Algorithm~\ref{alg:async_fl} for $H,Z$ and $E$ cloud, gateway and device epochs, we obtain 
\begin{equation}
    \min_{h=0}^{H-1}\E[\norm{\nabla L_N(\bm{\omega}_h)}^2] \leq \frac{\E[L_N(\bm{\omega}_0) - L_N(\bm{\omega}_{H})]}{\alpha\beta\gamma c H Z E}   + \Xi
\end{equation}
%All constant terms have been merged into $\Xi$ for clarity.
\end{theorem}

%Note that for weakly convex functions, which are non-convex, we are able to ensure $\cO(1/(HZE))$ convergence to a stationary point of the loss function at the cloud $L_N$, which matches the convergence rates for non-convex functions when there is no hierarchy or asynchrony. 
Theorem~\ref{thm:async_fl_1} extends the proof of the two-tier asynchronous FL algorithm in~\cite{xie2019asynchronous} to three tiers, namely the device-gateway-cloud architecture. Adding the extra gateway level only requires a bounded delay assumption at the gateway and adds constant terms in $\Xi$ due to the gateway, without sacrificing the convergence rate.
We are able to ensure convergence in spite of mild assumptions, for instance, constant staleness $s(\cdot)=1$ and weak convexity, and using stronger assumptions would enable even stronger results both theoretically and empirically.

\subsection{Learning Utility}
\label{sec:learning_utility}
%Even with the theoretical convergence guarantee, it is still necessary to design heterogeneity-aware device selection and device-gateway association algorithms to boost the practical convergence. 
%The device selection problem for synchronous FL can be rigorously formulated as selecting a subset of devices, namely combinatorial optimization, thus a variety of theoretical tools can be leveraged to search for or approximate the optimal solution~\cite{wang2020optimizing,xu2021online,xu2021dynamic,khan2020federated,ribero2020communication,balakrishnan2021diverse}. However, the asynchrony breaks such convenience and requires making decisions at any possible time stamp. To combat the challenge, we propose the learning utility metric per individual device, which can be combined with other system metrics to make asynchronous decisions.
%Though formulating client selection in asynchronous FL as a discrete-time control problem is possible, the resulted problem is non-trivial to solve.

To quantify the learning-wise contribution of aggregating each device's local model to the global model, we need a metric that accounts for data heterogeneity. State-of-the-art asynchronous client selection algorithms~\cite{hu2021device,hu2023scheduling} use local loss values to indicate the learning contribution of a device. %While loss values can indicate the importance of updates, it fails to account finer-grained gradient distribution.
%In a non-iid data distribution, a single loss value can be insufficient to reveal the uniqueness of updates.
Here in {\method}, we propose the \textit{learning utility} metric which takes into account gradient diversity. Compared to loss values, gradient diversity contains finer-grained information about data heterogeneity.
%While we do not have access to the local data on device, the weight updates after $E$ local training epochs can indicate the data distribution. Devices with similar data end up with similar weight updates, presenting clustered patterns as shown in~\cite{wang2020optimizing}.
%In this paper, Learning utility should take into account both the norm and the direction of weight updates, which stand for the absolute impact and the data heterogeneity respectively.

We extract the latest gradient on device $i$: $\nabla g^i(\bm{\mathbf{\omega}}_{\tau, \zeta, E}^i, \bm{\mathbf{\omega}}_{\tau, z}^i)$, and global gradient as a sum of all latest gradients: $\nabla L_N = \frac{1}{N}\sum_{i=1}^N \nabla g^i$. 
The \textit{learning utility} metric $u_i$ is defined for each device $i$ based on the gradient affinity with the global gradient, $\eta_i$, and the gradient diversity with the other devices, $\nu_i$:
\begin{subequations}
\begin{align}
    u_i &= \eta_i + \nu_i, \label{eq:learning_utility} \\
    \eta_i &= \nabla g^{i \top} \nabla L_N, \label{eq:gradient_affinity} \\
    \nu_i &= \frac{-1}{N-1} \sum_{j \neq i}^{N-1}  \nabla g^{i \top} \nabla g^j. \label{eq:gradient_disimil}
\end{align}
\end{subequations}

\begin{figure}%F5 %{r}{0.38\textwidth}
%  \centering
%  \vspace{-4mm}
  \includegraphics[width=0.8\columnwidth]{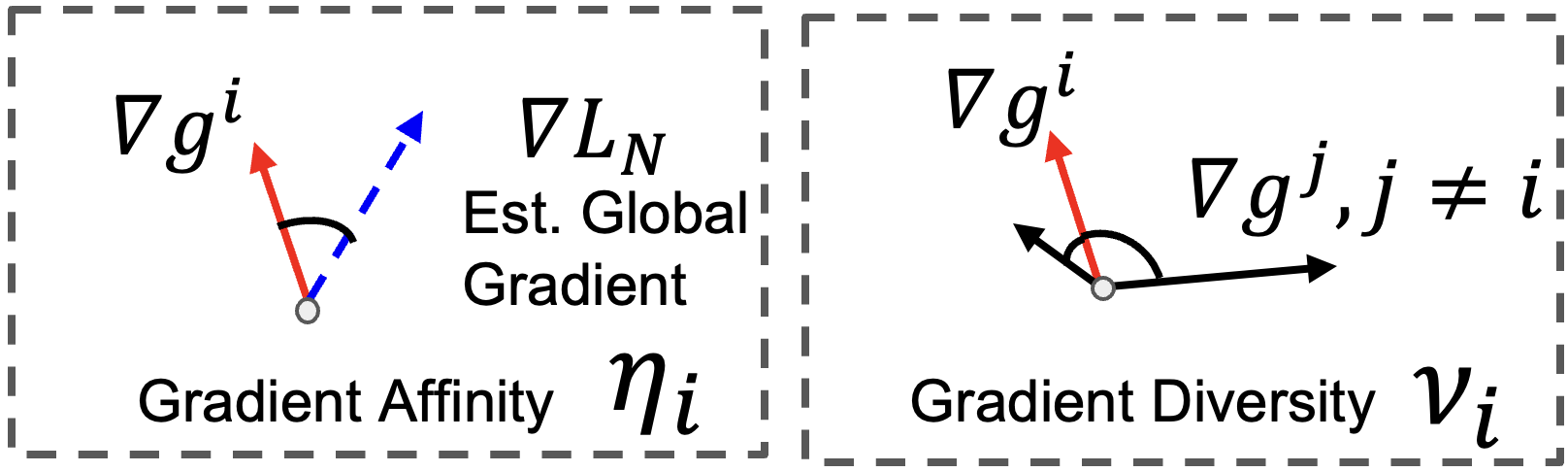}
%  \vspace{-8mm}
  \caption{\small Visualization of the defined learning utility metric combining gradient affinity and gradient diversity.}
 % \vspace{-6mm}
  \label{fig:learning_utility}
\end{figure}

\noindent To assist understanding, a visualization example is shown in Fig.~\ref{fig:learning_utility}.
The gradient affinity, $\eta_i$, evaluates the similarity between device $i$'s gradients and the global gradients, taking the dot product of $\nabla g^i$ and $\nabla L_N$ (Equation~\eqref{eq:gradient_affinity}). On the other hand, the term $\nu_i$ sums up the pairwise dissimilarity between $i$ and all other devices to evaluate the diversity of gradients (Equation~\eqref{eq:gradient_disimil}). 
By combining $\eta_i$ and $\nu_i$, the learning utility $u_i$ is a device-specific metric that favors the devices with close-to-global or largely diverse data distribution.
The idea of learning utility is inspired from online coreset selection~\cite{yoon2021online}, where the goal is to select a finite number of individual samples that preserve the maximal knowledge about data distribution to store in memory. 
%The selected coreset should bear sufficient affinity or diversity, which is similar to our case in FL device selection.
We stress that our learning utility metric jointly considers the norm and the distribution of gradients, thus integrating more information than just the norm of gradients or loss value. %If $\bm{\omega}^i$ leads to a large local loss, it increases the norm of $\Delta \bm{\omega}^i$ which will be reflected by increased $\eta_i$ and $v_i$.

\subsection{Device Selection and Device-Gateway Association}
\label{sec:cs_n_ca}

\begin{figure}%F6   %{r}{0.55\textwidth}
%  \centering
%  \vspace{-6mm}
  \includegraphics[width=0.95\columnwidth]{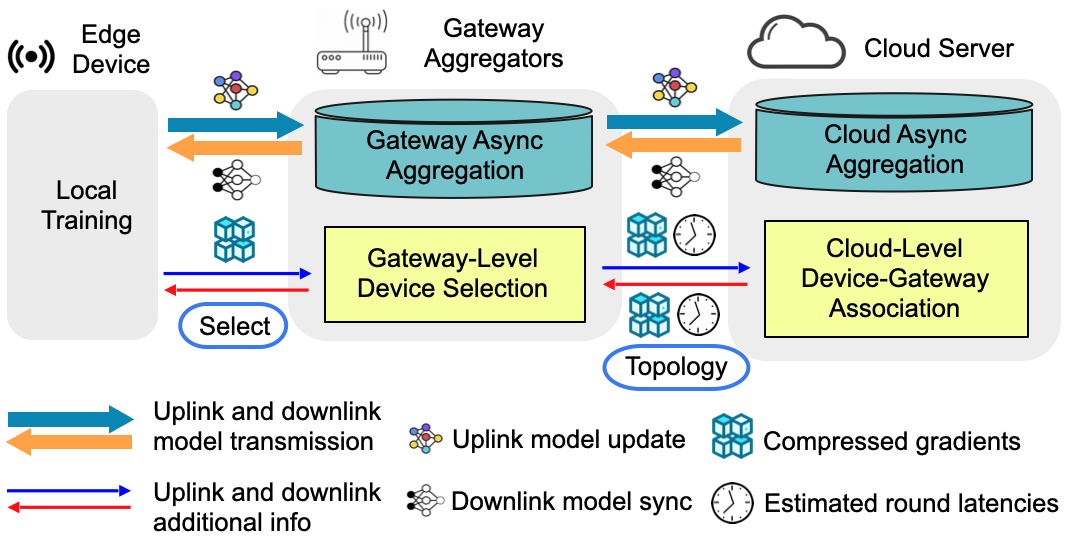}
%  \vspace{-6mm}
  \caption{\small The overview of the distributed design of gateway-level device selection and cloud-level device-gateway association. The thick arrows represent necessary communications for FL while the thin arrows stand for communication overhead.}
%  \vspace{-5mm}
  \label{fig:distributed_design}
\end{figure}

After identifying the learning utility metric to model data heterogeneity, in this section, we present the design of gateway-level device selection and cloud-level device-gateway association to enhance practical convergence of {\method}.
Both modules are designed to account diverse data distribution, heterogeneous latencies and unexpected stragglers.
Fig.~\ref{fig:distributed_design} presents an overview of the design and the necessary information to be collected and exchanged.
As shown, the gateway-level device selection module executes after the previous gateway aggregation, and thus adjusts device participation in real time. The device-gateway association module is fired less frequently, once after a certain number of cloud epochs, and thus manages network topology for longer-term performance.

%In general, the \textit{gateway-level} selections adjust device participation in real-time and with finer granularity, while the \textit{cloud-level} decisions . Both components are designed to account diverse data distribution, heterogeneous latencies and unexpected stragglers.

\textbf{Gateway-Level Device Selection.}
%The output of selection determines the selected devices in the \texttt{Gateway} process (line 13 in Algorithm~\ref{alg:async_fl}).
Given the current set of devices connected to gateway $j$ at time $t$ ($\mathbf{I}_{t,ij} = 1$), we select a subset to trigger asynchronous training. Suppose $d_i=1$ denotes that device $i$ is selected and the latest model is transmitted from the gateway to the device, but the new updated model has not returned from the device.
Once returned, the gateway-level device selection module records the compressed gradients information $\nabla g^i$ from the device, from which the \textit{learning utility} $u_i$ is updated. We also keep track of the moving average of device's round latency $\tau_{ij}$ at gateway $j$. 
In real-time selection, devices presenting large \textit{learning utility} and short round latency are preferred, as these devices are able to contribute significantly to the convergence in a fast manner.
We model the selection problem as an Integer Linear Program with variables of device selecting status $d_i$:
\begin{subequations}%Eq.10
%\small
%\label{eq:problem_selection}
\begin{align}
    \textrm{(Device Selection at} j) \: & \max \sum_{\mathbf{I}_{t,ij} = 1} d_i u_i \left (1/\tau_{ij} \right )^\kappa \label{eq:cs_obj} \\
    \textrm{s.t.} \quad & d_i R_{ij} \leq B_j, \quad \forall i \in \left \{i | \mathbf{I}_{t,ij} = 1 \right \} \label{eq:cs_comm_valid} \\
    & d_i \in \left \{ 0, 1 \right \} \quad \forall i \in \left \{i | \mathbf{I}_{t,ij} = 1 \right \} \label{eq:cs_var}
\end{align}
\end{subequations}
Equation~\eqref{eq:cs_obj} defines the objective combining \textit{learning utility} and round latency, with $\kappa$ as a hyperparameter that curves the contribution of round latency. Equation~\eqref{eq:cs_comm_valid} imposes the bandwidth upper bound at gateway $j$. 
The problem has at most $O(|N|)$ variables and linear constraints.

\textbf{Cloud-Level Device-Gateway Association.}
%While device selection decides which device to trigger at the gateway level, device-gateway association balances the learning contributions and throughputs with a global view at the cloud level. 
Given the feasible links $\mathbf{J}_t$ at time $t$, we need to determine the device-gateway association $\mathbf{I}_t$ used in the following cloud epochs.
We remind the readers that $\mathbf{J}_t$ denotes the real-time link availability, so unexpected device or link failures are reflected in $\mathbf{J}_t$ and our association solver is able to consider them timely.
At the cloud, we retrieve the gradients and round latency information from the corresponding gateways, and send back the decided association $\mathbf{I}_t$. 
Previous works have shown empirically that a stronger similarity of the gateway data distribution to the global distribution leads to a faster method convergence~\cite{deng2021share}.
%Additionally, the bandwidth resources at each gateway are bounded by $B_j$.
To ``shape'' the gateway distribution while fully utilizing bandwidth, we formulate a multi-objective optimization problem:
%To apply the above considerations to hierarchical IoT networks with \method, there are two issues to be solved:
%\begin{enumerate*}[label=(\arabic*)]
%    \item Because not all devices are selected and the aggregation is performing asynchronously, we need to approximate $\nabla{L_N(\bm{\omega_\tau})}$ during evaluating $\mathcal{S}(\Theta)$ as in Equation~\eqref{eq:simil}.
%    \item Distributed client selection decisions need to be made at each gateway with coordination.
%\end{enumerate*}
%In this section, we describe our design of the distributed client selection algorithm for {\method} which addresses the above two issues.
\begin{subequations}
%\small
\label{eq:problem_association}
\begin{align}
    \textrm{(Association at cloud)} \: & \max u_{slack} - \phi R_{slack} \label{eq:ca_obj} \\
    \textrm{s.t.} \: &\sum_{i=1}^{N} \mathbf{I}_{t,ij} \; u_i \geq u_{slack}, \quad \forall j \in \mathcal{G} \label{eq:slack_learning_utility} \\
    & \sum_{i=1}^N \mathbf{I}_{t,ij} \; R_{ij} / B_j \leq R_{slack}, \quad \forall j \in \mathcal{G} \label{eq:slack_data_rate}  \\
    & \mathbf{I}_{t,ij} \leq \mathbf{J}_{t,ij}, \quad \forall i \in \mathcal{N}, j \in \mathcal{G} \label{eq:comm_feasible} \\
    & \sum_{j=1}^{G} \mathbf{I}_{t,ij} \leq 1, \quad \forall i \in \mathcal{N} \label{eq:comm_valid} \\
    & \mathbf{I}_{t,ij} \in \left \{ 0, 1 \right \}, \quad \forall i \in \mathcal{N}
\end{align}
\end{subequations}
The total objective in Equation~\eqref{eq:ca_obj} balances the learning utility and the throughput of all associated devices at each gateway. Using slack variables, we are able to disassemble the max-min operation thus keep the problem an Integer Linear Program.
The first objective $u_{slack}$ is a slack variable defined as the minimal \textit{learning utility} among all gateways (Equation~\eqref{eq:slack_learning_utility}). The second objective $R_{slack}$ is a slack variable for the maximal associated throughput ratio ($R_j/B_j$) among all gateways (Equation~\eqref{eq:slack_data_rate}). 
Our goal is to make a balanced allocation of devices (considering both learning utility and data rate) which are proportional to the gateways' bandwidth limitations.
$\phi$ is used to tune the importance ratio between sub-objectives. 
%The last two equations guarantee the validity of the solution.
Equation~\eqref{eq:comm_feasible} limits $\mathbf{I}_t$ to use feasible links defined by $\mathbf{J}_t$. Equation~\eqref{eq:comm_valid} forces each device to connect to at most one gateway.
The problem has $O(|G||N|)$ variables and linear constraints.
%The problem has $|G||N|+2|G|$ variables and $|G||N|+|N|+2|G|$ linear constraints.

%, which can be solved within polynomial time in most cases, using the simplex algorithm~\cite{klee1972good}.
The two Integer Linear Programs are in the form of 0-1 Knapsack problem~\cite{freville2004multidimensional} which can be approached by a large number of algorithms ranging from optimal solver, greedy heuristics to meta-heuristics. In this paper, we implement both problems in the Gurobi solver~\cite{gurobi} and show the computation overhead is negligible in a 200-node network compared to the savings in convergence time. %We will study efficient heuristic for very large scale networks in our future work.

\textbf{Minimizing Communication Overhead.} 
The thin arrows in Fig.~\ref{fig:distributed_design} show the communication overhead, including latest gradients $\nabla g^i$, round latencies $\tau_{ij}$ and network topology $\bm{J}_t, \bm{I}_t$. Among these meta information for management, gradients act as the major source of overhead.
%To assist distributed decision-making, the latest device weight updates $\Delta \bm{\omega}$ and round latencies $\tau$ need to be shared between the gateway and cloud.
To minimize the communication overhead of {\method}, we first collect all devices' gradients during warmup, then perform Principle Component Analysis (PCA) on these gradients. Afterwards, we distribute the PCA parameters to all local devices. During the real training session, only the principle components of gradients are exchanged with gateways and clouds.
%Only the minimal information of associated devices is exchanged. For example, as shown in Figure~\ref{fig:distributed_design}, the gateway only uploads the weight updates of its connected devices, i.e., $\{\mathbf{\Delta \omega}^i | \mathbf{I}_{t,ij} = 1\}$. Similarly, the cloud will send the new-associated devices' weight updates history to the gateway, i.e., $\{\mathbf{\Delta \omega}^i | \mathbf{I}_{t,ij} = 0, \mathbf{I}_{t+1,ij}=1\}$. 
An overhead analysis of {\method} is presented in Section~\ref{sec:overhead_analysis}.

%%%%%%%%%%%%%%%%%%%%%%%%%%%%%%%%%%%%%%%%%%%%%%%%%%%%%%%%%%
% Evaluation
%%%%%%%%%%%%%%%%%%%%%%%%%%%%%%%%%%%%%%%%%%%%%%%%%%%%%%%%%%
\section{Evaluation}
\label{sec:evaluation}

\subsection{Datasets and Models}

To simulate heterogeneous data distribution, we retrieve non-iid datasets for four typical categories of IoT applications. The information of the datasets from each category, the partition settings and the models are summarized below.

\begin{table*}%4  %[h]
%\vspace{-4mm}
\small
\caption{\small Statistics of federated datasets. Model size refers to the size of the packet that contains all weights in the model.}
\label{tbl:datasets}
\vspace{-3mm}
%\begin{center}
\begin{tabular}{cccccc} % note: no vertical bars at all
\toprule % not \hline
\small
\textbf{Dataset} & \textbf{Devices} & \textbf{Avg. Samples/Device} & \textbf{Data Partitions} & \textbf{Models} & \textbf{Size}\\ % & \textbf{Tasks} \\
\midrule % not \hline
MNIST & 184 & 600 & Synthetic (assign 2 classes to each device) & CNN & 1.6MB \\ % & 10-class classification \\
FashionMNIST & 184 & 600 & Synthetic (assign 2 classes to each device) & CNN & 1.7MB \\ % & 10-class classification \\
CIFAR-10 & 50 & 1000 & Synthetic (assign 2 classes to each device) & ResNet-18 & 43MB \\ % & 10-class classification \\
%FEMNIST & 100 & 10 & & & &  Natural (each device is a writer) & CNN & 10-class classification\\ 
Shakespeare & 143 & 2892.5 & Natural (each device is a speaking role) & LSTM & 208KB \\ % & Next-character predicton \\
HAR & 30 & 308.5 & Natural (each device is a human subject) & MLP & 285KB \\ % & 6-class classification \\
HPWREN & 26 & 6377.4 & Natural (each device is a station) & LSTM & 292KB \\ %& Time-series prediction \\
\bottomrule % not \hline
\end{tabular}
%\end{center}
%\vspace{-6mm}
\end{table*}

\textbf{Application \#1: Image Classification.} %Image classification is a popular edge application on camera-mounted devices like smartphones. The goal is to predict the correct class category given raw images.
We select MNIST~\cite{deng2012mnist},  FashionMNIST~\cite{xiao2017fashionmnist}, CIFAR-10~\cite{krizhevsky2009learning} datasets for evaluation. %MNIST is a commonly-used handwriting image classification dataset while FashionMNIST presents a more challenging cloth classification task. Both MNIST and FashionMNIST have 10 classes and feature 28$\times$28 gray-scale images. CIFAR-10 features 32$\times$32 RGB images with 10 classes. 
We apply CNNs with two convolutional layers for MNIST and FashionMNIST, and the canonical ResNet-18~\cite{he2016deep} for CIFAR-10.
All three image classification datasets are partitioned synthetically with 2 classes randomly assigned to each device, and the local samples are dynamically updated from the same distribution. %, while the class appearance follows the normal distribution.

\textbf{Application \#2: Next-Character Prediction.} %Natural language processing is useful for edge scenarios like text auto-completion. 
We adopt the Shakespeare~\cite{caldas2018leaf} dataset, where the goal is to correctly predict the next character given a sequence of 80 characters. %This dataset is retrieved from \textit{The Complete Works of William Shakespeare} by separating different roles’ dialogues.
Local data is partitioned by assigning the dialogue of one role to one device.
We apply a two-layer LSTM classifier containing 100 hidden units with an 8D embedding layer, which is the base model for this application.

\textbf{Application \#3: Human Activity Recognition.}
%Human activity recognition is a typical task in smart home. 
We use the HAR dataset~\cite{anguita2013public} collected from 30 volunteers. %Each person performed six activities (walking, walking upstairs, walking downstairs, sitting, standing, laying) wearing a smartphone (Samsung Galaxy S II) on the waist. 
We assign the data from one individual to one device and apply a typical multilayer perceptron (MLP) model with two fully-connected layers. 
%Using the embedded accelerometer and gyroscope, each sample consists of 561 features. In our experiment, we assign the data from one individual to one device. For the model, we apply a typical multilayer perceptron (MLP) model with two fully-connected layers.

\textbf{Application \#4: Time-Series Prediction.} %Environmental monitoring is a common IoT application. While traditional statistical methods like ARIMA have shown good results on simple temperature prediction~\cite{hyndman20158}, recent works demonstrated that the LSTM-based models can capture the underlying context more effectively thus generating better results in FL~\cite{savi2021short}. In this paper, 
We build a time-series prediction task using the historical data collected by the High Performance Wireless \& Education Network (HPWREN)~\cite{hpwren}. HPWREN is a large-scale environmental monitoring sensor network spanning 20k sq. miles and collecting readings of temperature, humidity, wind speed, etc., every half an hour. Each reading has 11 features and we combine the readings in the past 24 hours (in total 48 readings) to be one sample. Each device holds the data collected at one station. The goal is to predict the next reading. We use the mean squared error (MSE) loss and a one-layer LSTM with 128 hidden units.

\subsection{System Implementation}
%We setup on a small-scale physical deployment and large-scale simulations, both sharing the same dataset, learning models and baseline settings. 
%All code for implementation will be released upon the publication of the paper.

As each trial of FL on a large physical deployment can take up to days, we mainly use simulations to mimic practical system and network heterogeneities. We further implement and validate {\method} on a smaller physical deployment.

\textbf{Large-Scale Simulation Setup\footnote{Implementation of the large-scale simulation is available at \url{https://github.com/Orienfish/Async-HFL}.}.}
%Since the ``long-tail'' delay distribution is hard to enforce on a small physical deployment with uncertainties, we  to observe the benefit of {\method} in a larger-scale network. 
We implement our discrete event-based simulator based on ns3-fl~\cite{ekaireb2022ns3fl}, the state-of-the-art FL simulator using PyTorch for FL experiments and ns-3~\cite{ns3} for network simulations.
Note, that in contrast to most existing frameworks that simulate communication rounds~\cite{li2020fair,xie2019asynchronous,wang2020tackling,karimireddy2020scaffold}, ns3-fl simulates the \textit{wall-clock} computation and communication time based on models from realistic measurements.
Approaches showing superb convergence with regard to rounds might perform poorly under wall-clock time if failing to consider system heterogeneities.
The network topology is configured based on NYCMesh as described in Section~\ref{sec:motivation} with 184 edge devices, 6 gateways, and 1 server.
%To include network uncertainties, we add a log-normal delay on top of the mean latency to produce a ``long-tail'' distribution similar as in Figure~\ref{fig:long_tail}.

%Note, that asynchronous FL simulation has much higher complexity than its synchronous counterpart. That is because, first, each asynchronous aggregating event depends on the order of previous aggregations, hence the simulation has to be run sequentially while parallel acceleration cannot be applied. Secondly, unlike synchronous FL where we only need to record a single model on gateways and cloud (which is updated after each synchronous aggregation), in asynchronous FL, we have to record the models at various asynchronous aggregation time. As a result, such complexity limits the scale of the experiments we are able to run.

\textbf{Physical Deployment Setup\footnote{Implementation of the physical deployment is available at \url{https://github.com/Orienfish/FedML}.}.}
%We further validate {\method} on a physical deployment. 
We implement {\method} on Raspberry Pi (RPi) 4B and 400 based on the state-of-the-art framework FedML~\cite{chaoyanghe2020fedml}. The physical deployment consists of 7 RPi 4Bs and 3 RPi 400s, distributing in 7 different houses and all connecting to the home Wi-Fi router. We ensure the variances of networking conditions by setting up some RPis in the farther end of the backyard, some in the bedroom in the vicinity of the router. We stress that the networking conditions may also be affected by the Wi-Fi traffic in real time. For example, the network delay could be longer if the residents are streaming a movie in the meantime. Such setup mimics the real-world scenarios where our application shares the bandwidth with other traffic and the network latency enjoys high diversity. In addition, during our experiment, we observe that RPis may fail to connect from the beginning, or (with rare probabilities) encounter unexpected suspension in the middle of one trial. Creating two virtual clients on each RPi, we are able to obtain a total of 20 clients on the RPi setup.

Apart from the RPis, we set up 20 more clients by requesting 1, 2 or 4 CPU cores from a CPU cluster and each accompanied by 4 GB RAM. Different from the RPi clients, where networking conditions vary largely, the CPU cluster has a stable internet connection but the computational delay varies depending on the requested resource.
Since we do not have access to the home Wi-Fi router, we deploy the implementation of gateways and the cloud on an Ethernet-connected desktop, with an Intel Core i7-8700@3.2GHz, 16GB RAM and a NVIDIA GeForce GTX 1060 6GB GPU.

%with the \texttt{Environment} set as \texttt{UrbanEnviroment} and the \texttt{CitySize} set as \texttt{LargeCity}.
\vspace{-3mm}
\subsection{Experimental Setup}
\vspace{-2mm}

%\begin{table}[t]
%\caption{ns-3 Simulation Parameters}
%\label{tbl:ns3}
%\small
%\begin{tabular}{ c | c }
%\hline
% Parameter & Wi-Fi \\ 
% \hline
% Routing Protocol & Static Routing \\
% \hline 
% MAC Layer & 802.11n 5GHZ \\
% \hline
% Traffic Type & TCP \\
% \hline
% Client Data Rate & 80-2048~kbps \\
% \hline
% Server Data Rate & 100~Mbps \\
% \hline
% Packet Size & 1024~bytes \\
%\hline
%Loss Model & HybridBuildingsPropagationLossModel \\
%\hline
%Error Rate Model & YansErrorRateModel \\
%\hline
%\end{tabular}
%\end{table}

\begin{table}%5  %{r}{0.49\textwidth}
\small
%\vspace{-4mm}
\caption{\small Important parameters setup on various datasets.}
\label{tbl:parameters}
\vspace{-3mm}
\begin{tabular}{c|ccccc} % note: no vertical bars at all
 \toprule % not \hline
 \small
 \textbf{Dataset} & \textbf{Target Acc./Err.} & \textbf{Gateway} & $\gamma$ & $\rho$ \\ 
% & \textbf{Accuracy} (\%) & \textbf{Bandwidth} (B/s) & & \\
 \midrule % not \hline
 MNIST & 95 & 1M & 0.01 & 0.1 \\
 FashionMNIST & 75 & 1M & 0.01 & 0.1 \\
 CIFAR-10 & 50 & 20M & 0.001 & 1.0 \\
 Shakespeare & 35 & 20k & 0.01 & 0.2 \\
 HAR & 95 & 20K & 0.003 & 0.1 \\
 HPWREN & 1.5e-5 (pred. err.) & 20K & 0.001 & 0.1 \\
 \bottomrule % not \hline
\end{tabular}
%\vspace{-6mm}
\end{table}

\begin{table*}[tb]
\vspace{-4mm}
\small
\caption{\small Convergence  speedup on large-scale simulations and various datasets. Bolded numbers reflect the best baseline result on each dataset.}
\label{tbl:large_converge}
\vspace{-4mm}
\begin{center}
\begin{tabular}{c|cccccccc} % note: no vertical bars at all
 \toprule % not \hline
 \textbf{Dataset} & \multicolumn{8}{|c}{\textbf{Convergence time speedup of {\method} with respect to baselines}} \\ 
 & Async-HL & Async-Random & Semi-async & RFL-HA & Sync-Oort & Sync-TiFL & Sync-DivFL & Sync-Random \\
 \midrule % not \hline
 MNIST & \textbf{1.11x} & 1.27x & 6.2x & 32.5x & 40.0x & 27.13x & 63.4x & 67.3x \\
 FashionMNIST & \textbf{1.08x} & 1.49x & 8.3x & 36.7x & 20.5x & 32.8x & 73.4x & 96.8x \\
 CIFAR-10 & \textbf{1.09x} & 1.40x & 2.3x & 12.3x & 44.3x & 59.0x & 62.0x & 61.7x \\
 Shakespeare & 1.19x & 1.79x & 0.59x & 0.71x & \textbf{0.31x} & 2.39x & 5.87x & 5.46x \\
 HAR & 1.31x & \textbf{1.22x} & 2.7x & 7.4x & 10.3x & 21.6x & 22.5x & 24.1x \\
 HPWREN & \textbf{1.11x} & 1.48x & 2.4x & 12.8x & 19.5x & 26.5x & 27.7x & 31.4x \\
 \bottomrule % not \hline
\end{tabular}
\end{center}
\vspace{-3mm}
\end{table*}

\begin{figure*}[!tbp] %F7
\centering
\begin{tabular}{cc}    
    \includegraphics[width=0.47\textwidth]{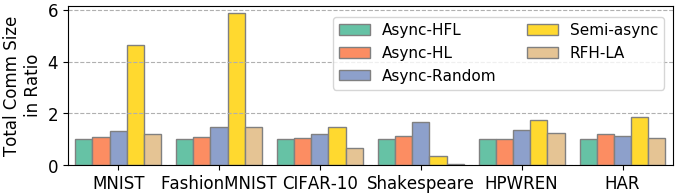} 
    &
    \includegraphics[width=0.47\textwidth]{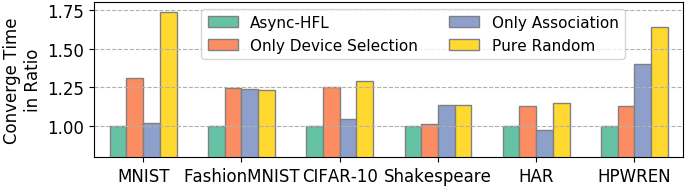}
\end{tabular}
\vspace{-4mm}
\caption{\small Left: Total communicated data size in ratio compared to {\method} on large-scale simulations and various datasets. Right: Convergence time in ratio compared to {\method} using various combinations of device selection and device-gateway association.}
\label{fig:add_sim_results}
%\vspace{-4mm}
\end{figure*}

\iffalse
\begin{figure}[tb]
  \centering
  %≈
  \includegraphics[width=0.47\textwidth]{fig/total_comm_size.png}
  \vspace{-4mm}
  \caption{\small Total communicated data size in ratio compared to {\method} on large-scale simulations and various datasets.}
  \vspace{-2mm}
  \label{fig:total+comm_size}
\end{figure}
\fi

\iffalse
\begin{figure}[tb]
  \centering
  \includegraphics[width=0.47\textwidth]{fig/ablation.png}
  \vspace{-4mm}
  \caption{\small Convergence time in ratio compared to {\method} using various combinations of device selection and device-gateway association modules.}
  \vspace{-4mm}
  \label{fig:ablation}
\end{figure}
\fi

\smallskip
\noindent \textbf{Baselines.} Given that the major design of {\method} is around device selection and association, we adopt state-of-the-art client-selection methods from the synchronous, hybrid, semi-asynchronous, and asynchronous FL schemes to compare. We add the prefix \textbf{sync} to the baselines using synchronous aggregations at both gateway and cloud. Conversely, \textbf{async} indicates asynchronous aggregations at both gateway and cloud. The only two baselines not following this naming rule are RFL-HA and Semi-async as follows. %The naming rule of baselines is visualized in Figure~\ref{fig:}.
\begin{itemize}
    \item \textbf{Sync-Random/TiFL/Oort/DivFL} makes random device-gateway association while device selections are made via random selection, TiFL~\cite{chai2020tifl}, Oort~\cite{lai2021oort} and DivFL~\cite{balakrishnan2021diverse} respectively. TiFL groups devices with similar delays to one tier and greedily selects high-loss devices in one tier until reaching the throughput limit. Oort uses a multi-arm bandits based algorithm to balance loss and latency. DivFL utilizes a greedy method to maximize a submodular function which takes the diversity of gradients into account.
    \item \textbf{RFL-HA~\cite{wang2021resource}} uses synchronous aggregation at gateways and asynchronous aggregation at cloud. While applying a random device selection at the gateway level, RFL-HA utillizes a re-clustering heuristic to adjust device-gateway associations.
    \item \textbf{Semi-async} performs semi-asynchronous aggregations at gateways as in~\cite{nguyen2022federated} and synchronous aggregations at cloud. Random choices are applied for device selection and device-gateway association. We experiment with the semi-period of 50, 100, 150 seconds and pick the best results.
    \item \textbf{Async-Random/HL} uses random device-gateway association and random or high-loss first device selection under the asynchronous scheme. Prioritizing the nodes with high loss or large gradients' norm is the state-of-the-art approach for asynchronous FL~\cite{hu2021device,hu2023scheduling}. We did not compare with~\cite{zhu2022online} as their algorithm depends on completely different metrics.
\end{itemize}
%FAVOR~\cite{wang2020optimizing}, TiFL, RFL-HA, 

\noindent \textbf{Evaluation Metrics.}
%To evaluate the efficiency of FL methods under heterogeneous system characteristics, 
For the simulation, we compare the convergence time, i.e., the wall-clock time to reach a predetermined accuracy or test loss (i.e., loss on the test dataset). Detailed parameters setup are listed in Table~\ref{tbl:parameters}. The target accuracy or loss is close to the optimal value that is reached by FedAvg. We also compare the total communicated data size to account communication efficiency. For the physical deployment, we quantify convergence using the accuracy or test loss at the same wall-clock elapsed time. 
%Note, that the majority of existing FL works evaluate the accuracy or test loss at various communication rounds~\cite{li2020fair,xie2019asynchronous,wang2020tackling,karimireddy2020scaffold}. Approaches showing superb convergence with regard to rounds might perform poorly under wall-clock time due to the failure of considering heterogeneous delays and other system characteristics.
We also study the execution time, which is indicative of energy consumption on real platforms.

\label{sec:physical_results}
\begin{figure*}[!tbp]%F8
\centering
\begin{tabular}{cccc}    
    \multicolumn{4}{c}{\includegraphics[width=0.43\textwidth]{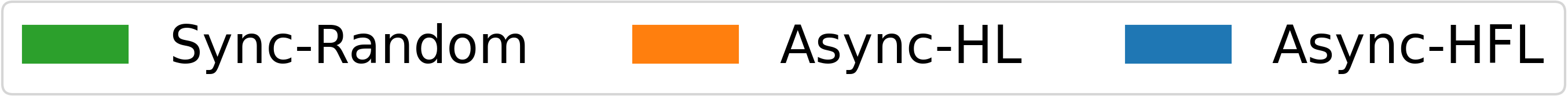}} \\
    \includegraphics[width=0.23\textwidth,height=3.1cm]{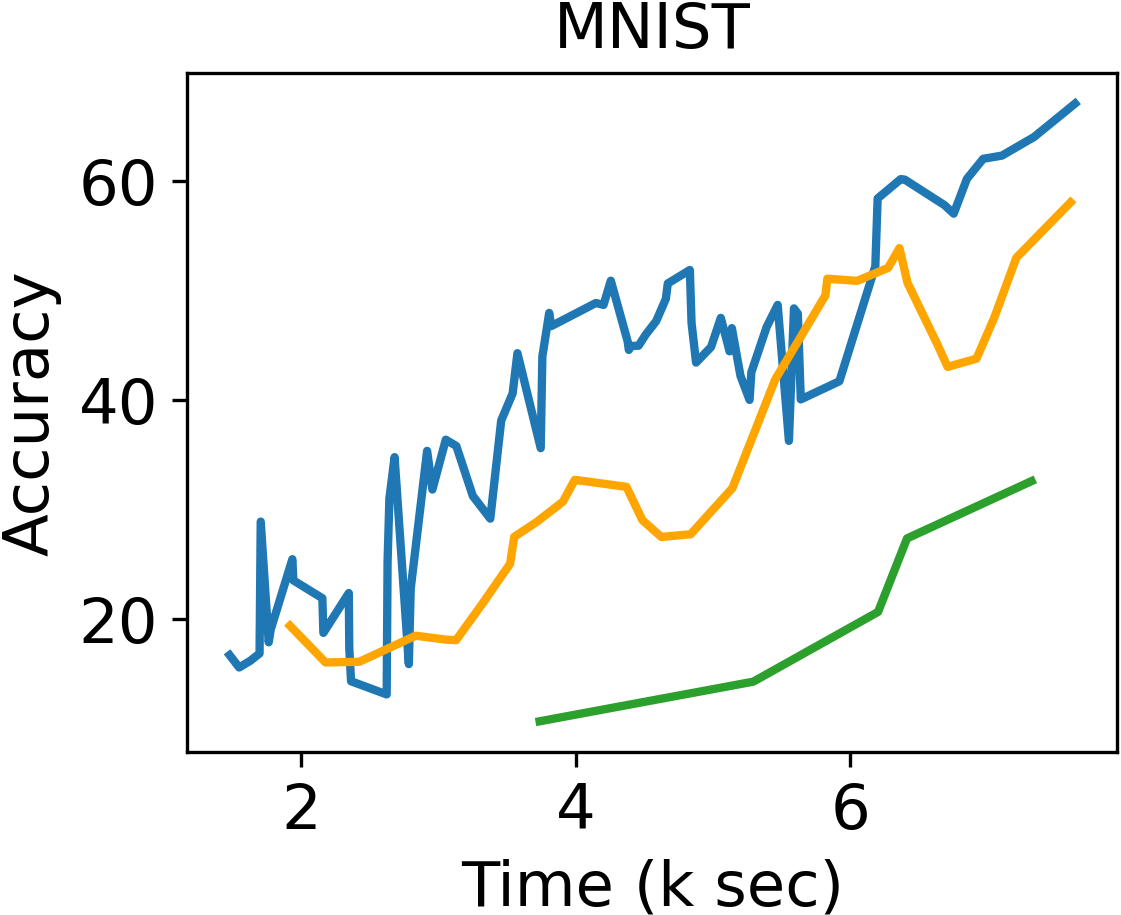}
    & 
    \includegraphics[width=0.23\textwidth,height=3.1cm]{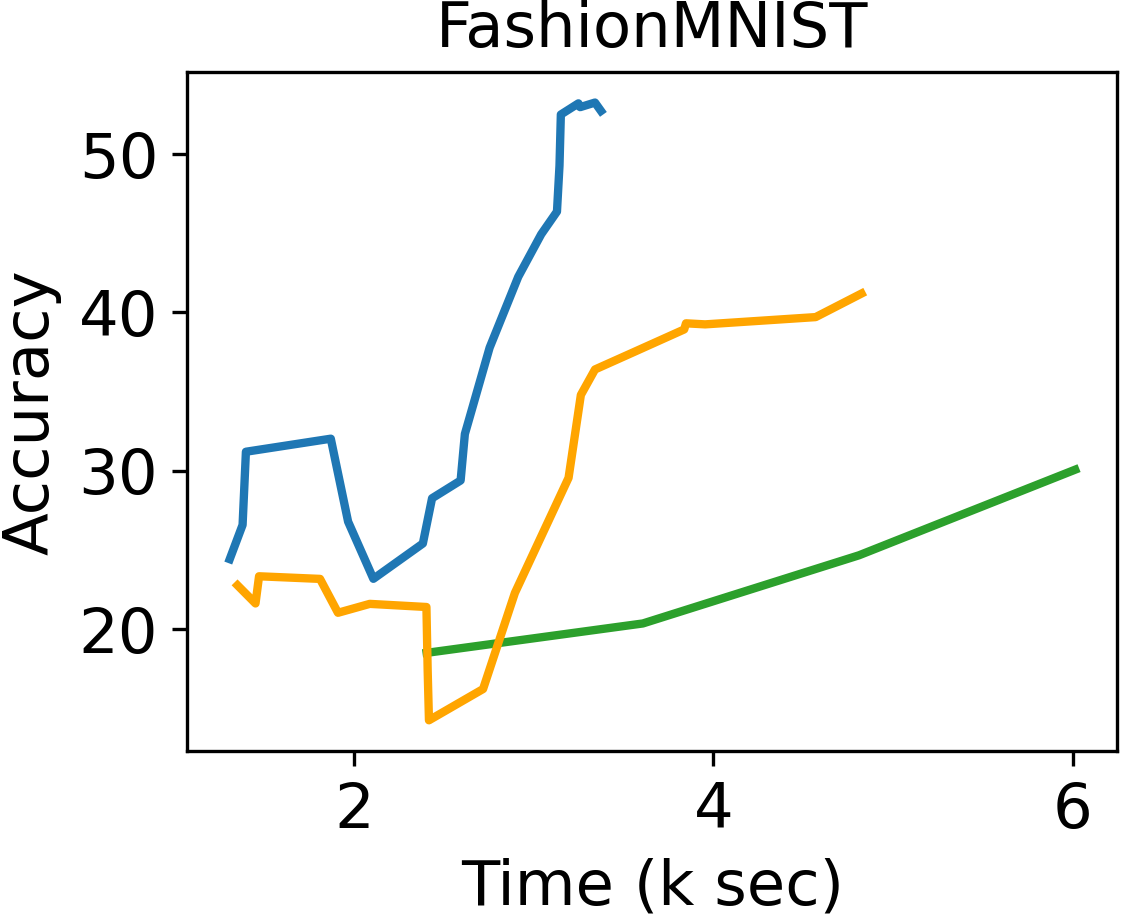}
    & 
    \includegraphics[width=0.23\textwidth,height=3.1cm]{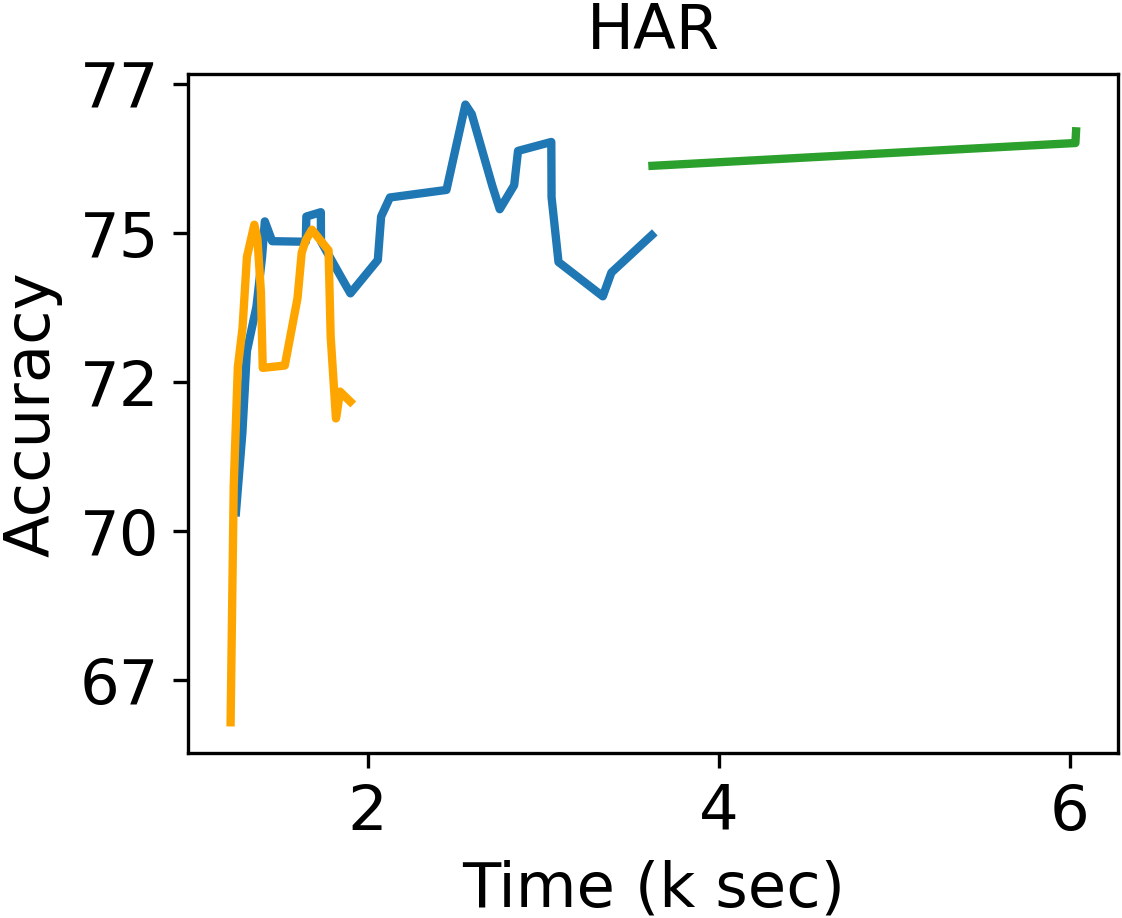}
    &
    \includegraphics[width=0.23\textwidth,height=3.1cm]{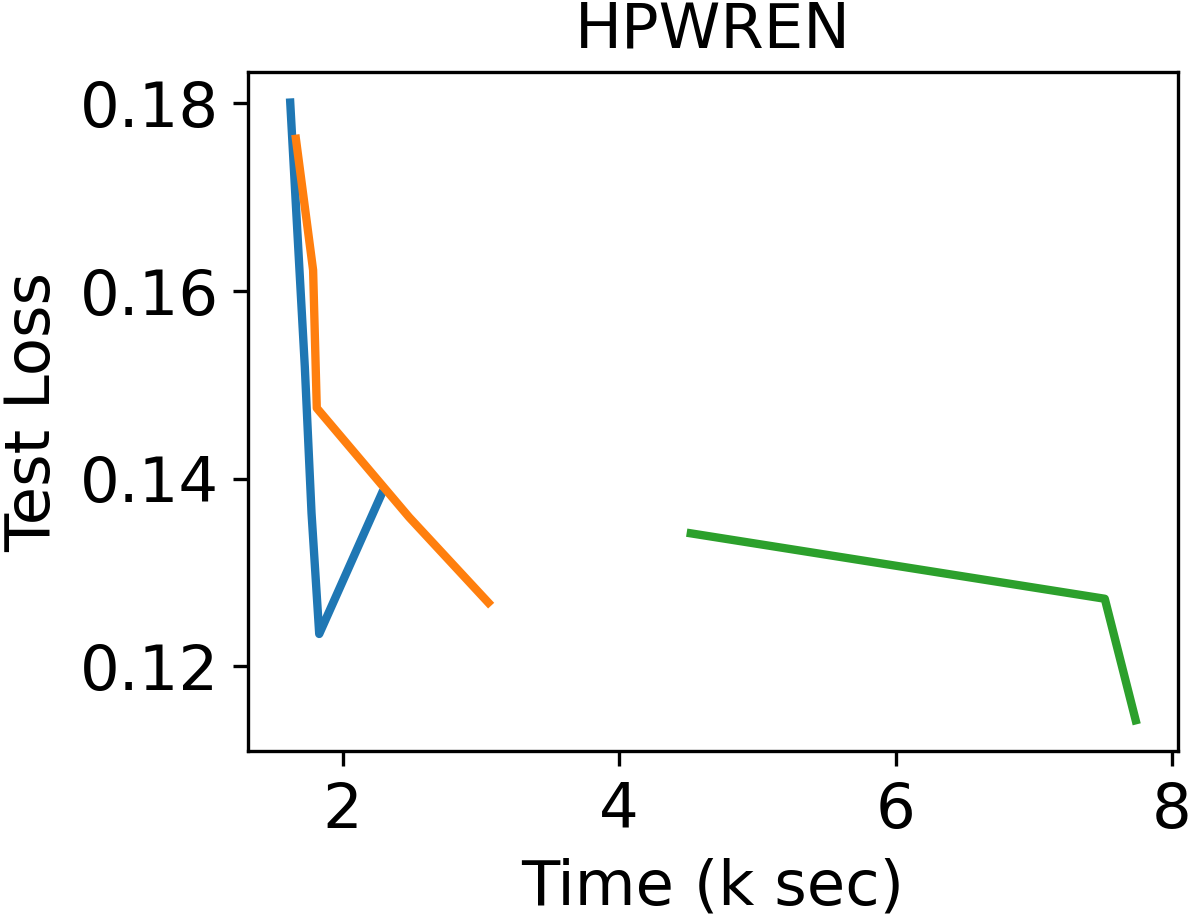}
\end{tabular}
\vspace{-4mm}
\caption{\small Convergence results under wall-clock time on the physical deployment.}
\label{fig:physical_result}
%\vspace{-4mm}
\end{figure*}

\subsection{Results on Large-Scale Simulations}
\label{sec:simulation_results}
\noindent \textbf{Convergence Results.}
We first report the simulated convergence time on all datasets using NYCMesh topology. %The target accuracy and gateway bandwidth settings are reported in Table~\ref{tbl:large_converge}, along with the convergence time of {\method} and its speedup with respect to various baselines.
We set device epochs $E=5$ and gateway epochs $Z=20$ for asynchronous, $Z=5$ for semi-asynchronous and synchronous gateway aggregations.  
Each method is tested with three random trials and we report the average convergence time and the corresponding speedup in Table~\ref{tbl:large_converge}. On all three image classification datasets, {\method} achieves a minimum 1.11x, 1.08x, 1.09x speedup over the best baseline. On HAR and HPWREN datasets (with smaller number of devices, see Table~\ref{tbl:datasets}) {\method} surpasses the best baseline by 0.22x and 0.11x respectively.
Hence the design of {\method} to balance learning utility and system characteristics works under both synthetic and nature data partition.
The only exception is Shakespeare, where Sync-Oort, RFL-HA and semi-async reach the target accuracy faster than {\method}. 
Nevertheless, we emphasize that {\method} still has 1.19x speedup compared to the state-of-the-art asynchronous FL. 
We speculate that the relative slower convergence of all asynchronous methods on Shakespeare roots in the essential converging difficulty of two-tier asynchronous FL algorithms. In the convergence curve of Shakespeare, we observe the first test accuracy increase with asynchronous methods after around 100 cloud epochs, while the synchronous baseline improves test accuracy at the first cloud epoch. We will refine the algorithmic design of {\method} for efficient convergence on Shakespeare in our future work.
In general, synchronous methods take much longer to reach the target accuracy due to the long waiting time, even with delay-aware client selection methods such as Oort and TiFL.
Both RFL-HA and Semi-async leverage hybrid aggregation scheme, thus converge slower than the fully asynchronous methods while faster than the synchronous approaches, except on Shakespeare. 
%The hybrid baseline, RFL-HA, converges at most 44.87x slower. This is because although RFL-HA uses asynchronous aggregation at cloud level, the gateway aggregations are synchronous thus intra-gateway latency heterogeneities significantly slow down the process.
The convergence speedup of {\method} over state-of-the-art asynchronous methods is 1.08-1.31x on all datasets.
%is less significant than the others. However, we emphasize that the absolute time savings in seconds are more than 1000 seconds in average - 1091, 2278 and 1934 seconds for MNIST, FashionMNIST and CIFAR-10 respectively.
%The specific speedup is determined by various factors including data distribution, bandwidth and delay distribution.

%Shakespeare, HAR and HPWREN are naturally partitioned datasets. 
%HAR and HPWREN have less than 30 devices.
%{\method} surpasses the majority of baselines except Async+HL on Shakepeare, Async+Random on HAR and Async+HL, Async+Random, RFL-HA on HPWREN. Such outcome could be attributed to the naturally non-iid data distribution and small device numbers.
%The design of {\method} depends on the learning utility metric where the global weight updates $\bm{\Delta \omega}_N$ plays an important role in evaluating the affinity of devices. With a naturally non-iid data distribution and small device number, the estimation of $\bm{\Delta \omega}_N$ might be biased and lead to suboptimal device preference.
%Nevertheless, {\method} significantly exceeds all synchronous baselines on all datasets.

The total communicated data size of non-synchronous methods compared to {\method} is shown in Fig.~\ref{fig:add_sim_results} (left). We are only interested in comparing non-synchronous methods as synchronous aggregation trades long waiting time for less cloud epochs and communication savings. Therefore, synchronous methods take much less communication to converge but the slowdown is usually unacceptable as shown in Table~\ref{tbl:large_converge}.
{\method} saves total communicated data size by 2.6\%-21.6\% and 14.5\%-66.8\% compared to Async-HL and Async-Random on all datasets. The total transmitted data size of semi-async and RFL-HA on Shakespeare is significantly lower than asynchronous approaches due to their fast convergence. Note, that on CIFAR-10, although RFL-HA consumes only 0.65x exchanged data compared to {\method}, it takes 11.3x longer to reach the same accuracy. The hybrid schemes of RFL-HA and Semi-async do not end up with faster convergence nor communication savings on MNIST, FashionMNIST, HPWREN and HAR.

\noindent \textbf{Performance Breakdown.}
\label{sec:ablation_studies}
{\method} includes two modules to boost the FL performance: gateway-level device selection and cloud-level device-gateway association. 
To evaluate the contribution of each component separately, we compare the convergence time on all datasets using {\it (i)} pure random selections, {\it (ii)} only the device selection, {\it (iii)} only the device-gateway association, and {\it (iv)} the complete {\method} as shown in Fig.~\ref{fig:add_sim_results} (right). The target accuracy and bandwidth are set to the same as in Table~\ref{tbl:parameters}.
%The benefits of {\method} reveals better under larger datasets and networks.
%The device selection and association components contribute collaboratively 
Each module contributes various extents on different datasets.
On MNIST, CIFAR-10 and HAR, the device-gateway association improves convergence more significantly by balancing the network topology, achieving 1.71x, 1.23x and 1.17x speedup by itself. On FashionMNIST, using one module does not change much, but applying both modules leads to a 1.23x speedup. On Shakespeare, the speedup is mainly supported by our coreset device selection with a 1.12x speedup by itself. On HPWREN, applying a single device selection or device-gateway association module leads to a 1.45x or 1.17x speedup, while using both modules contributes to a 1.64x total speedup.  Hence both device selection and association components are necessary in {\method} to deal with various data and system heterogeneities.

\subsection{Results on Physical Deployment}

\noindent \textbf{Convergence Results.}
We validate {\method} on the physical deployment running  MNIST, FashionMNIST, HAR and HPWREN datasets. The accuracy or test loss under wall-clock time are summarized in Fig.~\ref{fig:physical_result}. We run {\method} and Async-HL for 30 cloud rounds, Sync-Random for 3 cloud rounds, unless the system is suspended due to stragglers. Note, that Async-HL is the state-of-the-art asynchronous baseline and presents the second best result in simulations. {\method} ends up with 70\%, 56\% and 75\% accuracies on MNIST, FashionMNIST and HAR, while Async-HL only reaches 62\%, 36\% and 73\% at similar time (after 7.6K, 3.4K and 2K seconds).
For the synchronous baseline, we are only able to obtain very limited traces due to straggler effects. After setting a timeout limit for synchronous aggregations, we acquire the curves in Fig.~\ref{fig:physical_result} with very slow convergence.
The HPWREN dataset is very computational challenging for all methods and a lot of devices drop off due to no communication for a long time. {\method} strives for convergence within 2K seconds, while Async-HL and Sync-Random reach similar test loss after 3K and 7.7K seconds. 
While a small-scale physical deployment can be largely affected by uncertainties, we are able to observe consistently better convergence using {\method} over the baselines on all four datasets.
The results demonstrate the robustness of {\method} under delay heterogeneities and stragglers. This is because the {\method} performance is dynamically guided by its two modules:  (i) the gateway-level device selection module, which timely adjusts device participation, and  (ii) the cloud-level device-gateway association, which considers device dropouts via taking $\mathbf{J}_t$ as input.

\begin{figure}%F9  %{r}{0.48\textwidth}
\centering
%\vspace{-4mm}
%\setlength{\tabcolsep}{0.2pt}
\begin{tabular}{@{}cc}%    
    \includegraphics[width=0.23\textwidth]{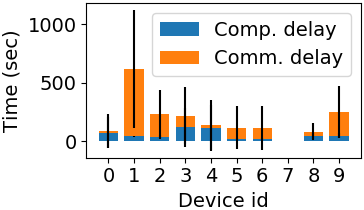}
    & 
    \includegraphics[width=0.23\textwidth]{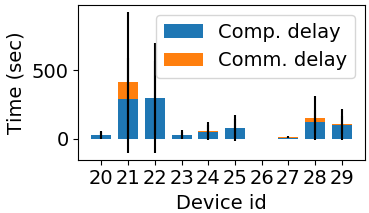}
\end{tabular}
\vspace{-4mm}
  \caption{\small Round latency results on the physical deployment running MNIST. Left: Time breakups on RPis. Right: Time breakups on CPUs.}
  \vspace{-4mm}
  \label{fig:physical}
\end{figure}

\noindent \textbf{Round Latency.}
Fig.~\ref{fig:physical} displays the round latency measurements of our practical setup, which demonstrates how challenging our physical deployment is. To remind the reader, round latency is the time to complete one gateway round of downloading the model to device, training the model on device, then returning the updated model back to the gateway.
%One gateway round could be as fast as 10 seconds, or cost more than 1000 seconds in the marginal cases, presenting an overall ``long tail'' pattern similar as Figure~\ref{fig:long_tail}.
%Similar as Figure~\ref{fig:long_tail}, very long round latency can occur in rare cases hence the overall distribution follows the ``long-tail'' pattern. 
The measurement supports our argument that IoT networks present heterogeneous system and network characteristics.
In more details, Fig.~\ref{fig:physical} left and right show the round latency breakup on ten representatives of RPis and CPU clusters respectively. The missing columns indicate failed devices. %Figure~\ref{fig:energy}~(d) reports the estimated energy breakup on RPis based on our pre-built power models from energy profiling experiments. % The power model takes CPU frequency as input and estimates computation and communication power with less than 5\% error. 
Our physical deployment setup covers two typical scenarios with very different breakups. For RPis, the major heterogeneity comes from the network side, as we setup the RPis at various places with different distances to the router. For CPU clusters, the computational delay rather than communication delay presents more variations due to varying number of requested CPU cores.
For FL, both heterogeneities cause the largely varied and unstable round latency distribution that {\method} targets to address.

\subsection{Sensitivity Analysis.}
\label{sec:sensitivity_analysis}
%The benefits of simulation is that we can easily tune the system parameters and observe the impacts. For the heterogeneous IoT networks we consider, there are two major factors that affect the outcome of {\method}: the gateway bandwidth limit and the underlying round latency distribution. We adjust both parameters and compare the convergence speedup of {\method} with regard to the synchronous and asynchronous random baselines. 

\noindent \textbf{Bandwidth Limitation.} Fig.~\ref{fig:sensitivity} (left) shows the convergence time to reach 95\% on MNIST when altering bandwidth limits at all gateways. The speedup over Async-Random is more significant under more restricted bandwidth (3.47x under 0.5MB/s vs. 0.56x under 2MB/s), as the benefit of intelligently selecting subset of devices reveals more with limited resource. Compared to the Sync-Random baseline, the speedup gap closes under 0.5MB/s bandwidth. When only a limited number of devices can be selected, FedAvg (Sync-Random) gets a stable convergence via averaging the models from multiple devices.

\iffalse
\begin{wrapfigure}{r}{0.48\textwidth}
\vspace{-4mm}
\setlength{\tabcolsep}{0.2pt}
\begin{tabular}{cc}
    \multicolumn{2}{c}{\includegraphics[width=0.48\textwidth]{fig/sensitivity_legend.png}} \\ 
    \includegraphics[width=0.24\textwidth]{fig/sensitivity_bw.png}
    & 
    \includegraphics[width=0.24\textwidth]{fig/sensitivity_target_acc.png}
\end{tabular}
\vspace{-4mm}
\caption{\small Convergence time in ratio compared to {\method} running MNIST under various bandwidth limits at gateways (left) and target accuracy (right) at cloud.}
\vspace{-5mm}
\label{fig:sensitivity}
\end{wrapfigure}
\fi

\iffalse
\begin{wrapfigure}{r}{0.48\textwidth}
\setlength{\tabcolsep}{0.2pt}
\begin{tabular}{ccc}
    \includegraphics[width=0.16\textwidth]{fig/kappa.png}
    & 
    \includegraphics[width=0.16\textwidth]{fig/phi.png} 
    &
    \includegraphics[width=0.16\textwidth]{fig/pca.png}
\end{tabular}
\vspace{-6mm}
\caption{\small Sensitivity experiments of {\method} on HAR dataset.}
\vspace{-6mm}
\label{fig:sensitivity_params}
\end{wrapfigure}
\fi

\noindent \textbf{Target Accuracy.}
Fig.~\ref{fig:sensitivity} (right) shows the convergence time to reach various target accuracies on MNIST with the same set of other settings. The speedup over Async-Random is 1.99x, 1.50x and 1.34x for reaching 85\%, 90\% and 95\%. Under the same settings, the speedup over Sync-Random is 106x, 78x and 51x. The results demonstrate {\method}'s fast convergence in the early stage, which can be attributed to prioritizing diverse and fast devices in {\method}'s management design.

\begin{figure}%F10  %{r}{0.48\textwidth}
%\begin{minipage}[b]{0.48\textwidth}
%\setlength{\tabcolsep}{0.2pt}
\begin{tabular}{@{}cc}
    \multicolumn{2}{c}{\includegraphics[width=0.45\textwidth]{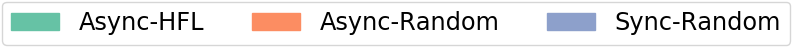}} \\ 
    \includegraphics[width=0.23\textwidth]{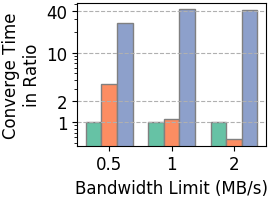}
    & 
    \includegraphics[width=0.23\textwidth]{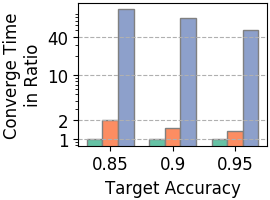}
\end{tabular}
\vspace{-4mm}
\caption{\small Convergence time in ratio compared to {\method} running MNIST under various bandwidth limits at gateways (left) and target accuracy (right) at cloud.}
\label{fig:sensitivity}
%\end{minipage}
\end{figure}

\begin{figure}%F11
%\begin{minipage}[b]{0.48\textwidth}
%\setlength{\tabcolsep}{0.2pt}
\begin{tabular}{ccc}
    \includegraphics[width=0.3\columnwidth]{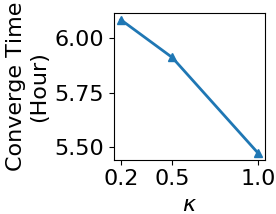}
    & 
    \includegraphics[width=0.3\columnwidth]{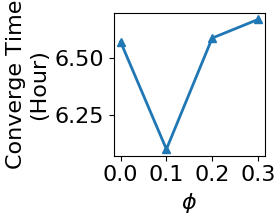} 
    &
    \includegraphics[width=0.3\columnwidth]{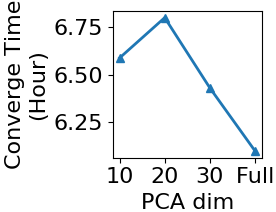}
\end{tabular}
\vspace{-5mm}
\caption{\small Sensitivity experiments of {\method} on HAR dataset.}
%\vspace{-5mm}
\label{fig:sensitivity_params}
%\end{minipage}
\end{figure}

\noindent \textbf{Hyperparameters $\kappa$ and $\phi$.}
We experiment the impact of $\kappa$ (Equation~\eqref{eq:cs_obj}) and $\phi$ (Equation~\eqref{eq:ca_obj}) on the final convergence time, as both parameters determine the balance between data heterogeneity (learning utility) and system heterogeneity (round latency). We use the HAR dataset with configurations in Table~\ref{tbl:parameters}. Fig.~\ref{fig:sensitivity_params} (left) shows the wall-clock time to reach the same accuracy using $\kappa=0.2, 0.5, 1.0$. A larger $\kappa$ increases the weight of delays during gateway-level device selection thus results in faster convergence. Fig.~\ref{fig:sensitivity_params} (middle) depicts the wall-clock convergence time using $\phi=0, 0.1, 0.2, 0.3$. $\phi=0$ means only data heterogeneity is considered, while a larger $\phi$ increases the contribution of bandwidth limitation during cloud-level device-gateway association. A proper $\phi$ (in this case, $\phi=0.1$) leads to the best convergence performance by jointly considering data and system aspects.
%Figure~\ref{fig:sensitivity} (right) shows the convergence time on MNIST using different $\sigma$ in the log-normal distribution to mimic the long-tailness.
%A larger $\sigma$ extends the ``tail'' of the distribution, hence increases the negative impact of selecting a slow device. {\method} achieves 4.15x (1.06x), 41.78x (1.10x), 374.93x (1.18x) over the synchronous (asynchronous) random baselines with $\sigma=1,2,3$. The results indicate that {\method} speeds up more significantly under heterogeneous system characteristics, as {\method} is designed to unleash the full potential of asynchronous FL.

\subsection{Overhead Analysis}
\label{sec:overhead_analysis}
As shown in Fig.~\ref{fig:distributed_design}, the major communication overhead of {\method} comes from exchanging the gradients. Using PCA to compress the gradients, the effect of various PCA dimensions on convergence time while processing the HAR dataset is shown in Fig.~\ref{fig:sensitivity_params} (right). A PCA compression of 30 dimensions introduces a communication overhead of <0.5\%, while the increase on convergence time (compared to using the full gradients) is less than 6\%.
Hence, the PCA compression strategy effectively reduces communication overhead while preserving convergence speed.
%{\method} introduces zero communication overhead on the device-gateway link, as the weight updates are used for both aggregation and learning utility.
%On the gateway-cloud link, the 
%However, the overhead is mitigated as (i) only information of re-associated devices needs to be exchanged. On a stable network, association adjustment happens very rare. (ii) {\method} uses less total number of rounds to reach the target accuracy, thus total transmitted information is also reduced.
On the computational side, the device selection algorithm consumes 1.6, 1.4, 4.3, 0.1 seconds per selection on the MNIST, FashionMNIST, CIFAR-10 and Shakespeare datasets. The time consumption of cloud-level association is 1.1, 0.9, 8.7 and 0.3 seconds per selection on the server for the above datasets.  These additional computational times are negligible on the physical deployment with an average 120.26 seconds of round latency.

%%%%%%%%%%%%%%%%%%%%%%%%%%%%%%%%%%%%%%%%%%%%%%%%%%%%%%%%%%
% Conclusion
%%%%%%%%%%%%%%%%%%%%%%%%%%%%%%%%%%%%%%%%%%%%%%%%%%%%%%%%%%
\vspace{-2mm}
\section{Conclusion}
\label{sec:conclusion}

In this paper, we propose {\method}, the first end-to-end asynchronous hierarchical Federated Learning framework which jointly considers data, system heterogeneities, stragglers and scalability in IoT networks. {\method} performs asynchronous aggregations on both gateways and cloud, thus achieving faster convergence with heterogeneous delays and being robust to stragglers. 
With the \textit{learning utility} metric to quantify gradient diversity, we design the device selection and device-gateway association modules to balance \textit{learning utility}, round latencies and unexpected stragglers, collaboratively optimizing practical model convergence.
We conduct comprehensive simulations based on ns-3 and NYCMesh to evaluate the {\method} under various network characteristics. Our results show a 1.08-1.31x speedup in terms of \textit{wall-clock} convergence time and 2.6-21.6\% communication savings compared to state-of-the-art asynchronous FL algorithms. Our physical deployment proves robust convergence under unexpected stragglers.

\begin{acks}
This work was initiated during Xiaofan Yu's  internship at Arm Research in summer of 2021.
The research was supported in part by National Science Foundation under Grants \#2112665 (TILOS AI Research Institute), \#2003279, \#1911095, \#1826967, \#2100237, \#2112167.
\end{acks}


\balance
\bibliographystyle{ACM-Reference-Format}
\bibliography{ref}


\begin{thebibliography}{62}

%%% the Web address.

\ifx \showCODEN    \undefined \def \showCODEN     #1{\unskip}     \fi
\ifx \showDOI      \undefined \def \showDOI       #1{#1}\fi
\ifx \showISBNx    \undefined \def \showISBNx     #1{\unskip}     \fi
\ifx \showISBNxiii \undefined \def \showISBNxiii  #1{\unskip}     \fi
\ifx \showISSN     \undefined \def \showISSN      #1{\unskip}     \fi
\ifx \showLCCN     \undefined \def \showLCCN      #1{\unskip}     \fi
\ifx \shownote     \undefined \def \shownote      #1{#1}          \fi
\ifx \showarticletitle \undefined \def \showarticletitle #1{#1}   \fi
\ifx \showURL      \undefined \def \showURL       {\relax}        \fi
% invisible to TeX
\providecommand\bibfield[2]{#2}
\providecommand\bibinfo[2]{#2}
\providecommand\natexlab[1]{#1}
\providecommand\showeprint[2][]{arXiv:#2}

\bibitem[hpw(2022)]%
        {hpwren}
 \bibinfo{year}{2022}\natexlab{}.
\newblock \bibinfo{title}{{High Performance Wireless Research \& Education
  Network (HPWREN)}}.
\newblock
\newblock
\urldef\tempurl%
\url{http://hpwren.ucsd.edu/}
\showURL{%
\tempurl}
\newblock
\shownote{[Online]}.


\bibitem[nyc(2022)]%
        {nycmesh}
 \bibinfo{year}{2022}\natexlab{}.
\newblock \bibinfo{title}{{New York City (NYC) Mesh}}.
\newblock
\newblock
\urldef\tempurl%
\url{https://www.nycmesh.net/map/}
\showURL{%
\tempurl}
\newblock
\shownote{[Online]}.


\bibitem[ns3(2022)]%
        {ns3}
 \bibinfo{year}{2022}\natexlab{}.
\newblock \bibinfo{title}{ns-3: a discrete-event network simulator for internet
  systems}.
\newblock
\newblock
\urldef\tempurl%
\url{https://www.nsnam.org/}
\showURL{%
\tempurl}
\newblock
\shownote{[Online]}.


\bibitem[Abad et~al\mbox{.}(2020)]%
        {abad2020hierarchical}
\bibfield{author}{\bibinfo{person}{Mehdi Salehi~Heydar Abad},
  \bibinfo{person}{Emre Ozfatura}, \bibinfo{person}{Deniz Gunduz}, {and}
  \bibinfo{person}{Ozgur Ercetin}.} \bibinfo{year}{2020}\natexlab{}.
\newblock \showarticletitle{Hierarchical federated learning across
  heterogeneous cellular networks}. In \bibinfo{booktitle}{\emph{ICASSP}}.
  IEEE, \bibinfo{pages}{8866--8870}.
\newblock


\bibitem[Abdellatif et~al\mbox{.}(2022)]%
        {abdellatif2022communication}
\bibfield{author}{\bibinfo{person}{Alaa~Awad Abdellatif},
  \bibinfo{person}{Naram Mhaisen}, \bibinfo{person}{Amr Mohamed},
  \bibinfo{person}{Aiman Erbad}, \bibinfo{person}{Mohsen Guizani},
  \bibinfo{person}{Zaher Dawy}, {and} \bibinfo{person}{Wassim Nasreddine}.}
  \bibinfo{year}{2022}\natexlab{}.
\newblock \showarticletitle{Communication-efficient hierarchical federated
  learning for IoT heterogeneous systems with imbalanced data}.
\newblock \bibinfo{journal}{\emph{Future Generation Computer Systems}}
  \bibinfo{volume}{128} (\bibinfo{year}{2022}), \bibinfo{pages}{406--419}.
\newblock


\bibitem[Ahmad and Pothuganti(2020)]%
        {ahmad2020design}
\bibfield{author}{\bibinfo{person}{Irfan Ahmad} {and}
  \bibinfo{person}{Karunakar Pothuganti}.} \bibinfo{year}{2020}\natexlab{}.
\newblock \showarticletitle{Design \& implementation of real time autonomous
  car by using image processing \& IoT}. In \bibinfo{booktitle}{\emph{ICSSIT}}.
  IEEE, \bibinfo{pages}{107--113}.
\newblock


\bibitem[Anguita et~al\mbox{.}(2013)]%
        {anguita2013public}
\bibfield{author}{\bibinfo{person}{Davide Anguita}, \bibinfo{person}{Alessandro
  Ghio}, \bibinfo{person}{Luca Oneto}, \bibinfo{person}{Xavier Parra~Perez},
  {and} \bibinfo{person}{Jorge~Luis Reyes~Ortiz}.}
  \bibinfo{year}{2013}\natexlab{}.
\newblock \showarticletitle{A public domain dataset for human activity
  recognition using smartphones}. In \bibinfo{booktitle}{\emph{ESANN}}.
  \bibinfo{pages}{437--442}.
\newblock


\bibitem[Balakrishnan et~al\mbox{.}(2021)]%
        {balakrishnan2021diverse}
\bibfield{author}{\bibinfo{person}{Ravikumar Balakrishnan},
  \bibinfo{person}{Tian Li}, \bibinfo{person}{Tianyi Zhou},
  \bibinfo{person}{Nageen Himayat}, \bibinfo{person}{Virginia Smith}, {and}
  \bibinfo{person}{Jeff Bilmes}.} \bibinfo{year}{2021}\natexlab{}.
\newblock \showarticletitle{Diverse Client Selection for Federated Learning via
  Submodular Maximization}. In \bibinfo{booktitle}{\emph{ICLR}}.
\newblock


\bibitem[Beniczky et~al\mbox{.}(2021)]%
        {beniczky2021machine}
\bibfield{author}{\bibinfo{person}{S{\'a}ndor Beniczky},
  \bibinfo{person}{Philippa Karoly}, \bibinfo{person}{Ewan Nurse},
  \bibinfo{person}{Philippe Ryvlin}, {and} \bibinfo{person}{Mark Cook}.}
  \bibinfo{year}{2021}\natexlab{}.
\newblock \showarticletitle{Machine learning and wearable devices of the
  future}.
\newblock \bibinfo{journal}{\emph{Epilepsia}}  \bibinfo{volume}{62}
  (\bibinfo{year}{2021}), \bibinfo{pages}{S116--S124}.
\newblock


\bibitem[Caldas et~al\mbox{.}(2018)]%
        {caldas2018leaf}
\bibfield{author}{\bibinfo{person}{Sebastian Caldas}, \bibinfo{person}{Sai
  Meher~Karthik Duddu}, \bibinfo{person}{Peter Wu}, \bibinfo{person}{Tian Li},
  \bibinfo{person}{Jakub Kone{\v{c}}n{\`y}}, \bibinfo{person}{H~Brendan
  McMahan}, \bibinfo{person}{Virginia Smith}, {and} \bibinfo{person}{Ameet
  Talwalkar}.} \bibinfo{year}{2018}\natexlab{}.
\newblock \showarticletitle{Leaf: A benchmark for federated settings}.
\newblock \bibinfo{journal}{\emph{arXiv preprint arXiv:1812.01097}}
  (\bibinfo{year}{2018}).
\newblock


\bibitem[Chai et~al\mbox{.}(2020a)]%
        {chai2020tifl}
\bibfield{author}{\bibinfo{person}{Zheng Chai}, \bibinfo{person}{Ahsan Ali},
  \bibinfo{person}{Syed Zawad}, \bibinfo{person}{Stacey Truex},
  \bibinfo{person}{Ali Anwar}, \bibinfo{person}{Nathalie Baracaldo},
  \bibinfo{person}{Yi Zhou}, \bibinfo{person}{Heiko Ludwig},
  \bibinfo{person}{Feng Yan}, {and} \bibinfo{person}{Yue Cheng}.}
  \bibinfo{year}{2020}\natexlab{a}.
\newblock \showarticletitle{Tifl: A tier-based federated learning system}. In
  \bibinfo{booktitle}{\emph{HPDC}}. \bibinfo{pages}{125--136}.
\newblock


\bibitem[Chai et~al\mbox{.}(2020b)]%
        {chai2020fedat}
\bibfield{author}{\bibinfo{person}{Zheng Chai}, \bibinfo{person}{Yujing Chen},
  \bibinfo{person}{Liang Zhao}, \bibinfo{person}{Yue Cheng}, {and}
  \bibinfo{person}{Huzefa Rangwala}.} \bibinfo{year}{2020}\natexlab{b}.
\newblock \showarticletitle{Fedat: A communication-efficient federated learning
  method with asynchronous tiers under non-iid data}.
\newblock \bibinfo{journal}{\emph{arXiv preprin arxiv:2010.05958}}
  (\bibinfo{year}{2020}).
\newblock


\bibitem[Chaoyang~He(2020)]%
        {chaoyanghe2020fedml}
\bibfield{author}{\bibinfo{person}{et~al. Chaoyang~He}.}
  \bibinfo{year}{2020}\natexlab{}.
\newblock \showarticletitle{Fedml: A research library and benchmark for
  federated machine learning}.
\newblock \bibinfo{journal}{\emph{arXiv preprint arXiv:2007.13518}}
  (\bibinfo{year}{2020}).
\newblock


\bibitem[Chen et~al\mbox{.}(2020)]%
        {chen2020convergence}
\bibfield{author}{\bibinfo{person}{Mingzhe Chen}, \bibinfo{person}{H~Vincent
  Poor}, \bibinfo{person}{Walid Saad}, {and} \bibinfo{person}{Shuguang Cui}.}
  \bibinfo{year}{2020}\natexlab{}.
\newblock \showarticletitle{Convergence time minimization of federated learning
  over wireless networks}. In \bibinfo{booktitle}{\emph{ICC}}. IEEE,
  \bibinfo{pages}{1--6}.
\newblock


\bibitem[Chen et~al\mbox{.}(2022)]%
        {chen2022heterogeneous}
\bibfield{author}{\bibinfo{person}{Shuai Chen}, \bibinfo{person}{Xiumin Wang},
  \bibinfo{person}{Pan Zhou}, \bibinfo{person}{Weiwei Wu},
  \bibinfo{person}{Weiwei Lin}, {and} \bibinfo{person}{Zhenyu Wang}.}
  \bibinfo{year}{2022}\natexlab{}.
\newblock \showarticletitle{Heterogeneous Semi-Asynchronous Federated Learning
  in Internet of Things: A Multi-Armed Bandit Approach}.
\newblock \bibinfo{journal}{\emph{IEEE Transactions on Emerging Topics in
  Computational Intelligence}} \bibinfo{volume}{6}, \bibinfo{number}{5}
  (\bibinfo{year}{2022}), \bibinfo{pages}{1113--1124}.
\newblock


\bibitem[Chen et~al\mbox{.}(2021)]%
        {chen2021towards}
\bibfield{author}{\bibinfo{person}{Zheyi Chen}, \bibinfo{person}{Weixian Liao},
  \bibinfo{person}{Kun Hua}, \bibinfo{person}{Chao Lu}, {and}
  \bibinfo{person}{Wei Yu}.} \bibinfo{year}{2021}\natexlab{}.
\newblock \showarticletitle{Towards asynchronous federated learning for
  heterogeneous edge-powered internet of things}.
\newblock \bibinfo{journal}{\emph{Digital Communications and Networks}}
  \bibinfo{volume}{7}, \bibinfo{number}{3} (\bibinfo{year}{2021}),
  \bibinfo{pages}{317--326}.
\newblock


\bibitem[Deng(2012)]%
        {deng2012mnist}
\bibfield{author}{\bibinfo{person}{Li Deng}.} \bibinfo{year}{2012}\natexlab{}.
\newblock \showarticletitle{The mnist database of handwritten digit images for
  machine learning research}.
\newblock \bibinfo{journal}{\emph{IEEE Signal Processing Magazine}}
  (\bibinfo{year}{2012}).
\newblock


\bibitem[Deng et~al\mbox{.}(2021)]%
        {deng2021share}
\bibfield{author}{\bibinfo{person}{Yongheng Deng}, \bibinfo{person}{Feng Lyu},
  \bibinfo{person}{Ju Ren}, \bibinfo{person}{Yongmin Zhang},
  \bibinfo{person}{Yuezhi Zhou}, \bibinfo{person}{Yaoxue Zhang}, {and}
  \bibinfo{person}{Yuanyuan Yang}.} \bibinfo{year}{2021}\natexlab{}.
\newblock \showarticletitle{SHARE: Shaping Data Distribution at Edge for
  Communication-Efficient Hierarchical Federated Learning}. In
  \bibinfo{booktitle}{\emph{ICDCS}}. IEEE, \bibinfo{pages}{24--34}.
\newblock


\bibitem[Ekaireb et~al\mbox{.}(2022)]%
        {ekaireb2022ns3fl}
\bibfield{author}{\bibinfo{person}{Emily Ekaireb}, \bibinfo{person}{Xiaofan
  Yu}, \bibinfo{person}{Kazim Ergun}, \bibinfo{person}{Quanling Zhao},
  \bibinfo{person}{Kai Lee}, \bibinfo{person}{Muhammad Huzaifa}, {and}
  \bibinfo{person}{Tajana Rosing}.} \bibinfo{year}{2022}\natexlab{}.
\newblock \showarticletitle{ns3-fl: Simulating Federated Learning with ns-3}.
  In \bibinfo{booktitle}{\emph{WNS-3}}. \bibinfo{pages}{97--104}.
\newblock


\bibitem[Feng et~al\mbox{.}(2022)]%
        {feng2022mobility}
\bibfield{author}{\bibinfo{person}{Chenyuan Feng}, \bibinfo{person}{Howard~H
  Yang}, \bibinfo{person}{Deshun Hu}, \bibinfo{person}{Zhiwei Zhao},
  \bibinfo{person}{Tony~QS Quek}, {and} \bibinfo{person}{Geyong Min}.}
  \bibinfo{year}{2022}\natexlab{}.
\newblock \showarticletitle{Mobility-aware cluster federated learning in
  hierarchical wireless networks}.
\newblock \bibinfo{journal}{\emph{IEEE Transactions on Wireless
  Communications}} \bibinfo{volume}{21}, \bibinfo{number}{10}
  (\bibinfo{year}{2022}), \bibinfo{pages}{8441--8458}.
\newblock


\bibitem[Fr{\'e}ville(2004)]%
        {freville2004multidimensional}
\bibfield{author}{\bibinfo{person}{Arnaud Fr{\'e}ville}.}
  \bibinfo{year}{2004}\natexlab{}.
\newblock \showarticletitle{The multidimensional 0--1 knapsack problem: An
  overview}.
\newblock \bibinfo{journal}{\emph{European Journal of Operational Research}}
  \bibinfo{volume}{155} (\bibinfo{year}{2004}).
\newblock


\bibitem[{Gurobi Optimization, LLC}(2022)]%
        {gurobi}
\bibfield{author}{\bibinfo{person}{{Gurobi Optimization, LLC}}.}
  \bibinfo{year}{2022}\natexlab{}.
\newblock \bibinfo{title}{{Gurobi Optimizer Reference Manual}}.
\newblock
\newblock
\urldef\tempurl%
\url{https://www.gurobi.com}
\showURL{%
\tempurl}


\bibitem[Hao et~al\mbox{.}(2020)]%
        {hao2020time}
\bibfield{author}{\bibinfo{person}{Jiangshan Hao}, \bibinfo{person}{Yanchao
  Zhao}, {and} \bibinfo{person}{Jiale Zhang}.} \bibinfo{year}{2020}\natexlab{}.
\newblock \showarticletitle{Time efficient federated learning with
  semi-asynchronous communication}. In \bibinfo{booktitle}{\emph{ICPADS}}.
  IEEE.
\newblock


\bibitem[He et~al\mbox{.}(2016)]%
        {he2016deep}
\bibfield{author}{\bibinfo{person}{Kaiming He}, \bibinfo{person}{Xiangyu
  Zhang}, \bibinfo{person}{Shaoqing Ren}, {and} \bibinfo{person}{Jian Sun}.}
  \bibinfo{year}{2016}\natexlab{}.
\newblock \showarticletitle{Deep residual learning for image recognition}. In
  \bibinfo{booktitle}{\emph{CVPR}}. \bibinfo{pages}{770--778}.
\newblock


\bibitem[Hu et~al\mbox{.}(2021)]%
        {hu2021device}
\bibfield{author}{\bibinfo{person}{Chung-Hsuan Hu}, \bibinfo{person}{Zheng
  Chen}, {and} \bibinfo{person}{Erik~G Larsson}.}
  \bibinfo{year}{2021}\natexlab{}.
\newblock \showarticletitle{Device scheduling and update aggregation policies
  for asynchronous federated learning}. In \bibinfo{booktitle}{\emph{SPAWC}}.
  IEEE, \bibinfo{pages}{281--285}.
\newblock


\bibitem[Hu et~al\mbox{.}(2023)]%
        {hu2023scheduling}
\bibfield{author}{\bibinfo{person}{Chung-Hsuan Hu}, \bibinfo{person}{Zheng
  Chen}, {and} \bibinfo{person}{Erik~G Larsson}.}
  \bibinfo{year}{2023}\natexlab{}.
\newblock \showarticletitle{Scheduling and Aggregation Design for Asynchronous
  Federated Learning over Wireless Networks}.
\newblock \bibinfo{journal}{\emph{IEEE Journal on Selected Areas in
  Communications}} (\bibinfo{year}{2023}).
\newblock


\bibitem[Huba et~al\mbox{.}(2022)]%
        {huba2022papaya}
\bibfield{author}{\bibinfo{person}{Dzmitry Huba}, \bibinfo{person}{John
  Nguyen}, \bibinfo{person}{Kshitiz Malik}, \bibinfo{person}{Ruiyu Zhu},
  \bibinfo{person}{Mike Rabbat}, \bibinfo{person}{Ashkan Yousefpour},
  \bibinfo{person}{Carole-Jean Wu}, \bibinfo{person}{Hongyuan Zhan},
  \bibinfo{person}{Pavel Ustinov}, \bibinfo{person}{Harish Srinivas},
  {et~al\mbox{.}}} \bibinfo{year}{2022}\natexlab{}.
\newblock \showarticletitle{Papaya: Practical, private, and scalable federated
  learning}.
\newblock \bibinfo{journal}{\emph{Proceedings of Machine Learning and Systems}}
   \bibinfo{volume}{4} (\bibinfo{year}{2022}), \bibinfo{pages}{814--832}.
\newblock


\bibitem[Imteaj and Amini(2020)]%
        {imteaj2020fedar}
\bibfield{author}{\bibinfo{person}{Ahmed Imteaj} {and} \bibinfo{person}{M~Hadi
  Amini}.} \bibinfo{year}{2020}\natexlab{}.
\newblock \showarticletitle{Fedar: Activity and resource-aware federated
  learning model for distributed mobile robots}. In
  \bibinfo{booktitle}{\emph{ICMLA}}. IEEE.
\newblock


\bibitem[Jasim et~al\mbox{.}(2021)]%
        {jasim2021design}
\bibfield{author}{\bibinfo{person}{Nabaa~Ali Jasim}, \bibinfo{person}{Haider
  TH}, {and} \bibinfo{person}{Salim~AL Rikabi}.}
  \bibinfo{year}{2021}\natexlab{}.
\newblock \showarticletitle{Design and implementation of smart city
  applications based on the internet of things.}
\newblock \bibinfo{journal}{\emph{iJIM}} \bibinfo{volume}{15},
  \bibinfo{number}{13} (\bibinfo{year}{2021}).
\newblock


\bibitem[Karimireddy et~al\mbox{.}(2020)]%
        {karimireddy2020scaffold}
\bibfield{author}{\bibinfo{person}{Sai~Praneeth Karimireddy},
  \bibinfo{person}{Satyen Kale}, \bibinfo{person}{Mehryar Mohri},
  \bibinfo{person}{Sashank Reddi}, \bibinfo{person}{Sebastian Stich}, {and}
  \bibinfo{person}{Ananda~Theertha Suresh}.} \bibinfo{year}{2020}\natexlab{}.
\newblock \showarticletitle{Scaffold: Stochastic controlled averaging for
  federated learning}. In \bibinfo{booktitle}{\emph{ICML}}. PMLR,
  \bibinfo{pages}{5132--5143}.
\newblock


\bibitem[Khan et~al\mbox{.}(2020)]%
        {khan2020federated}
\bibfield{author}{\bibinfo{person}{Latif~U Khan}, \bibinfo{person}{Shashi~Raj
  Pandey}, \bibinfo{person}{Nguyen~H Tran}, \bibinfo{person}{Walid Saad},
  \bibinfo{person}{Zhu Han}, \bibinfo{person}{Minh~NH Nguyen}, {and}
  \bibinfo{person}{Choong~Seon Hong}.} \bibinfo{year}{2020}\natexlab{}.
\newblock \showarticletitle{Federated learning for edge networks: Resource
  optimization and incentive mechanism}.
\newblock \bibinfo{journal}{\emph{IEEE Communications Magazine}}
  \bibinfo{volume}{58}, \bibinfo{number}{10} (\bibinfo{year}{2020}),
  \bibinfo{pages}{88--93}.
\newblock


\bibitem[Krizhevsky et~al\mbox{.}(2009)]%
        {krizhevsky2009learning}
\bibfield{author}{\bibinfo{person}{Alex Krizhevsky}, \bibinfo{person}{Geoffrey
  Hinton}, {et~al\mbox{.}}} \bibinfo{year}{2009}\natexlab{}.
\newblock \showarticletitle{Learning multiple layers of features from tiny
  images}.
\newblock  (\bibinfo{year}{2009}).
\newblock


\bibitem[Lai et~al\mbox{.}(2021)]%
        {lai2021oort}
\bibfield{author}{\bibinfo{person}{Fan Lai}, \bibinfo{person}{Xiangfeng Zhu},
  \bibinfo{person}{Harsha~V Madhyastha}, {and} \bibinfo{person}{Mosharaf
  Chowdhury}.} \bibinfo{year}{2021}\natexlab{}.
\newblock \showarticletitle{Oort: Efficient federated learning via guided
  participant selection}. In \bibinfo{booktitle}{\emph{OSDI}}.
  \bibinfo{pages}{19--35}.
\newblock


\bibitem[Lee and Lee(2021)]%
        {lee2021adaptive}
\bibfield{author}{\bibinfo{person}{Hyun-Suk Lee} {and}
  \bibinfo{person}{Jang-Won Lee}.} \bibinfo{year}{2021}\natexlab{}.
\newblock \showarticletitle{Adaptive transmission scheduling in wireless
  networks for asynchronous federated learning}.
\newblock \bibinfo{journal}{\emph{IEEE Journal on Selected Areas in
  Communications}} \bibinfo{volume}{39}, \bibinfo{number}{12}
  (\bibinfo{year}{2021}), \bibinfo{pages}{3673--3687}.
\newblock


\bibitem[Li et~al\mbox{.}(2021a)]%
        {li2021hermes}
\bibfield{author}{\bibinfo{person}{Ang Li}, \bibinfo{person}{Jingwei Sun},
  \bibinfo{person}{Pengcheng Li}, \bibinfo{person}{Yu Pu}, \bibinfo{person}{Hai
  Li}, {and} \bibinfo{person}{Yiran Chen}.} \bibinfo{year}{2021}\natexlab{a}.
\newblock \showarticletitle{Hermes: an efficient federated learning framework
  for heterogeneous mobile clients}. In \bibinfo{booktitle}{\emph{MobiCom}}.
  \bibinfo{pages}{420--437}.
\newblock


\bibitem[Li et~al\mbox{.}(2021b)]%
        {li2021fedmask}
\bibfield{author}{\bibinfo{person}{Ang Li}, \bibinfo{person}{Jingwei Sun},
  \bibinfo{person}{Xiao Zeng}, \bibinfo{person}{Mi Zhang}, \bibinfo{person}{Hai
  Li}, {and} \bibinfo{person}{Yiran Chen}.} \bibinfo{year}{2021}\natexlab{b}.
\newblock \showarticletitle{Fedmask: Joint computation and
  communication-efficient personalized federated learning via heterogeneous
  masking}. In \bibinfo{booktitle}{\emph{SenSys}}. \bibinfo{pages}{42--55}.
\newblock


\bibitem[Li et~al\mbox{.}(2022)]%
        {li2022pyramidfl}
\bibfield{author}{\bibinfo{person}{Chenning Li}, \bibinfo{person}{Xiao Zeng},
  \bibinfo{person}{Mi Zhang}, {and} \bibinfo{person}{Zhichao Cao}.}
  \bibinfo{year}{2022}\natexlab{}.
\newblock \showarticletitle{PyramidFL: A fine-grained client selection
  framework for efficient federated learning}. In
  \bibinfo{booktitle}{\emph{MobiCom}}. \bibinfo{pages}{158--171}.
\newblock


\bibitem[Li et~al\mbox{.}(2020)]%
        {li2020fedprox}
\bibfield{author}{\bibinfo{person}{Tian Li}, \bibinfo{person}{Anit~Kumar Sahu},
  \bibinfo{person}{Manzil Zaheer}, \bibinfo{person}{Maziar Sanjabi},
  \bibinfo{person}{Ameet Talwalkar}, {and} \bibinfo{person}{Virginia Smith}.}
  \bibinfo{year}{2020}\natexlab{}.
\newblock \showarticletitle{Federated optimization in heterogeneous networks}.
\newblock \bibinfo{journal}{\emph{Proceedings of Machine learning and systems}}
   \bibinfo{volume}{2} (\bibinfo{year}{2020}), \bibinfo{pages}{429--450}.
\newblock


\bibitem[Li et~al\mbox{.}(2019)]%
        {li2020fair}
\bibfield{author}{\bibinfo{person}{Tian Li}, \bibinfo{person}{Maziar Sanjabi},
  \bibinfo{person}{Ahmad Beirami}, {and} \bibinfo{person}{Virginia Smith}.}
  \bibinfo{year}{2019}\natexlab{}.
\newblock \showarticletitle{Fair resource allocation in federated learning}.
\newblock \bibinfo{journal}{\emph{arXiv preprint arXiv:1905.10497}}
  (\bibinfo{year}{2019}).
\newblock


\bibitem[Liu et~al\mbox{.}(2020)]%
        {liu2020client}
\bibfield{author}{\bibinfo{person}{Lumin Liu}, \bibinfo{person}{Jun Zhang},
  \bibinfo{person}{SH Song}, {and} \bibinfo{person}{Khaled~B Letaief}.}
  \bibinfo{year}{2020}\natexlab{}.
\newblock \showarticletitle{Client-edge-cloud hierarchical federated learning}.
  In \bibinfo{booktitle}{\emph{ICC}}. IEEE, \bibinfo{pages}{1--6}.
\newblock


\bibitem[Luo et~al\mbox{.}(2020)]%
        {luo2020hfel}
\bibfield{author}{\bibinfo{person}{Siqi Luo}, \bibinfo{person}{Xu Chen},
  \bibinfo{person}{Qiong Wu}, \bibinfo{person}{Zhi Zhou}, {and}
  \bibinfo{person}{Shuai Yu}.} \bibinfo{year}{2020}\natexlab{}.
\newblock \showarticletitle{Hfel: Joint edge association and resource
  allocation for cost-efficient hierarchical federated edge learning}.
\newblock \bibinfo{journal}{\emph{IEEE Transactions on Wireless
  Communications}} \bibinfo{volume}{19}, \bibinfo{number}{10}
  (\bibinfo{year}{2020}), \bibinfo{pages}{6535--6548}.
\newblock


\bibitem[McMahan et~al\mbox{.}(2017)]%
        {mcmahan2017communication}
\bibfield{author}{\bibinfo{person}{Brendan McMahan}, \bibinfo{person}{Eider
  Moore}, \bibinfo{person}{Daniel Ramage}, \bibinfo{person}{Seth Hampson},
  {and} \bibinfo{person}{Blaise~Aguera y Arcas}.}
  \bibinfo{year}{2017}\natexlab{}.
\newblock \showarticletitle{Communication-efficient learning of deep networks
  from decentralized data}. In \bibinfo{booktitle}{\emph{AISTATS}}. PMLR,
  \bibinfo{pages}{1273--1282}.
\newblock


\bibitem[Mitra et~al\mbox{.}(2021)]%
        {mitra2021linear}
\bibfield{author}{\bibinfo{person}{Aritra Mitra}, \bibinfo{person}{Rayana
  Jaafar}, \bibinfo{person}{George~J Pappas}, {and} \bibinfo{person}{Hamed
  Hassani}.} \bibinfo{year}{2021}\natexlab{}.
\newblock \showarticletitle{Linear convergence in federated learning: Tackling
  client heterogeneity and sparse gradients}.
\newblock \bibinfo{journal}{\emph{Advances in Neural Information Processing
  Systems}}  \bibinfo{volume}{34} (\bibinfo{year}{2021}),
  \bibinfo{pages}{14606--14619}.
\newblock


\bibitem[Nguyen et~al\mbox{.}(2022)]%
        {nguyen2022federated}
\bibfield{author}{\bibinfo{person}{John Nguyen}, \bibinfo{person}{Kshitiz
  Malik}, \bibinfo{person}{Hongyuan Zhan}, \bibinfo{person}{Ashkan Yousefpour},
  \bibinfo{person}{Mike Rabbat}, \bibinfo{person}{Mani Malek}, {and}
  \bibinfo{person}{Dzmitry Huba}.} \bibinfo{year}{2022}\natexlab{}.
\newblock \showarticletitle{Federated learning with buffered asynchronous
  aggregation}. In \bibinfo{booktitle}{\emph{AISTATS}}. PMLR,
  \bibinfo{pages}{3581--3607}.
\newblock


\bibitem[Ribero and Vikalo(2020)]%
        {ribero2020communication}
\bibfield{author}{\bibinfo{person}{Monica Ribero} {and} \bibinfo{person}{Haris
  Vikalo}.} \bibinfo{year}{2020}\natexlab{}.
\newblock \showarticletitle{Communication-efficient federated learning via
  optimal client sampling}.
\newblock \bibinfo{journal}{\emph{arXiv preprint arXiv:2007.15197}}
  (\bibinfo{year}{2020}).
\newblock


\bibitem[Sui et~al\mbox{.}(2016)]%
        {sui2016characterizing}
\bibfield{author}{\bibinfo{person}{Kaixin Sui}, \bibinfo{person}{Mengyu Zhou},
  \bibinfo{person}{Dapeng Liu}, \bibinfo{person}{Minghua Ma},
  \bibinfo{person}{Dan Pei}, \bibinfo{person}{Youjian Zhao},
  \bibinfo{person}{Zimu Li}, {and} \bibinfo{person}{Thomas Moscibroda}.}
  \bibinfo{year}{2016}\natexlab{}.
\newblock \showarticletitle{Characterizing and improving wifi latency in
  large-scale operational networks}. In \bibinfo{booktitle}{\emph{MobiSys}}.
  \bibinfo{pages}{347--360}.
\newblock


\bibitem[Tan et~al\mbox{.}(2022)]%
        {tan2022towards}
\bibfield{author}{\bibinfo{person}{Alysa~Ziying Tan}, \bibinfo{person}{Han Yu},
  \bibinfo{person}{Lizhen Cui}, {and} \bibinfo{person}{Qiang Yang}.}
  \bibinfo{year}{2022}\natexlab{}.
\newblock \showarticletitle{Towards personalized federated learning}.
\newblock \bibinfo{journal}{\emph{IEEE Trans. Neural Netw. Learn. Syst.}}
  (\bibinfo{year}{2022}).
\newblock


\bibitem[Wang et~al\mbox{.}(2020a)]%
        {wang2020optimizing}
\bibfield{author}{\bibinfo{person}{Hao Wang}, \bibinfo{person}{Zakhary Kaplan},
  \bibinfo{person}{Di Niu}, {and} \bibinfo{person}{Baochun Li}.}
  \bibinfo{year}{2020}\natexlab{a}.
\newblock \showarticletitle{Optimizing federated learning on non-iid data with
  reinforcement learning}. In \bibinfo{booktitle}{\emph{INFOCOM}}. IEEE.
\newblock


\bibitem[Wang et~al\mbox{.}(2020b)]%
        {wang2020tackling}
\bibfield{author}{\bibinfo{person}{Jianyu Wang}, \bibinfo{person}{Qinghua Liu},
  \bibinfo{person}{Hao Liang}, \bibinfo{person}{Gauri Joshi}, {and}
  \bibinfo{person}{H~Vincent Poor}.} \bibinfo{year}{2020}\natexlab{b}.
\newblock \showarticletitle{Tackling the objective inconsistency problem in
  heterogeneous federated optimization}.
\newblock \bibinfo{journal}{\emph{Adv Neural Inf Process Syst}}
  \bibinfo{volume}{33} (\bibinfo{year}{2020}), \bibinfo{pages}{7611--7623}.
\newblock


\bibitem[Wang et~al\mbox{.}(2021)]%
        {wang2021resource}
\bibfield{author}{\bibinfo{person}{Zhiyuan Wang}, \bibinfo{person}{Hongli Xu},
  \bibinfo{person}{Jianchun Liu}, \bibinfo{person}{He Huang},
  \bibinfo{person}{Chunming Qiao}, {and} \bibinfo{person}{Yangming Zhao}.}
  \bibinfo{year}{2021}\natexlab{}.
\newblock \showarticletitle{Resource-efficient federated learning with
  hierarchical aggregation in edge computing}. In
  \bibinfo{booktitle}{\emph{INFOCOM}}. IEEE, \bibinfo{pages}{1--10}.
\newblock


\bibitem[Wang et~al\mbox{.}(2022)]%
        {wang2022asynchronous}
\bibfield{author}{\bibinfo{person}{Zhongyu Wang}, \bibinfo{person}{Zhaoyang
  Zhang}, \bibinfo{person}{Yuqing Tian}, \bibinfo{person}{Qianqian Yang},
  \bibinfo{person}{Hangguan Shan}, \bibinfo{person}{Wei Wang}, {and}
  \bibinfo{person}{Tony~QS Quek}.} \bibinfo{year}{2022}\natexlab{}.
\newblock \showarticletitle{Asynchronous federated learning over wireless
  communication networks}.
\newblock \bibinfo{journal}{\emph{IEEE Transactions on Wireless
  Communications}} \bibinfo{volume}{21}, \bibinfo{number}{9}
  (\bibinfo{year}{2022}), \bibinfo{pages}{6961--6978}.
\newblock


\bibitem[Wu et~al\mbox{.}(2020)]%
        {wu2020safa}
\bibfield{author}{\bibinfo{person}{Wentai Wu}, \bibinfo{person}{Ligang He},
  \bibinfo{person}{Weiwei Lin}, \bibinfo{person}{Rui Mao},
  \bibinfo{person}{Carsten Maple}, {and} \bibinfo{person}{Stephen Jarvis}.}
  \bibinfo{year}{2020}\natexlab{}.
\newblock \showarticletitle{Safa: a semi-asynchronous protocol for fast
  federated learning with low overhead}.
\newblock \bibinfo{journal}{\emph{IEEE Trans. Comput.}} \bibinfo{volume}{70},
  \bibinfo{number}{5} (\bibinfo{year}{2020}), \bibinfo{pages}{655--668}.
\newblock


\bibitem[Xiao et~al\mbox{.}(2017)]%
        {xiao2017fashionmnist}
\bibfield{author}{\bibinfo{person}{Han Xiao}, \bibinfo{person}{Kashif Rasul},
  {and} \bibinfo{person}{Roland Vollgraf}.} \bibinfo{year}{2017}\natexlab{}.
\newblock \bibinfo{title}{Fashion-mnist: a novel image dataset for benchmarking
  machine learning algorithms}.
\newblock
\newblock


\bibitem[Xie et~al\mbox{.}(2019)]%
        {xie2019asynchronous}
\bibfield{author}{\bibinfo{person}{Cong Xie}, \bibinfo{person}{Sanmi Koyejo},
  {and} \bibinfo{person}{Indranil Gupta}.} \bibinfo{year}{2019}\natexlab{}.
\newblock \showarticletitle{Asynchronous federated optimization}.
\newblock \bibinfo{journal}{\emph{arXiv preprint arXiv:1903.03934}}
  (\bibinfo{year}{2019}).
\newblock


\bibitem[Xu et~al\mbox{.}(2021a)]%
        {xu2021online}
\bibfield{author}{\bibinfo{person}{Bo Xu}, \bibinfo{person}{Wenchao Xia},
  \bibinfo{person}{Jun Zhang}, \bibinfo{person}{Tony~QS Quek}, {and}
  \bibinfo{person}{Hongbo Zhu}.} \bibinfo{year}{2021}\natexlab{a}.
\newblock \showarticletitle{Online client scheduling for fast federated
  learning}.
\newblock \bibinfo{journal}{\emph{IEEE Wirel. Commun. Lett.}}
  \bibinfo{volume}{10}, \bibinfo{number}{7} (\bibinfo{year}{2021}),
  \bibinfo{pages}{1434--1438}.
\newblock


\bibitem[Xu et~al\mbox{.}(2021b)]%
        {xu2021dynamic}
\bibfield{author}{\bibinfo{person}{Bo Xu}, \bibinfo{person}{Wenchao Xia},
  \bibinfo{person}{Jun Zhang}, \bibinfo{person}{Xinghua Sun}, {and}
  \bibinfo{person}{Hongbo Zhu}.} \bibinfo{year}{2021}\natexlab{b}.
\newblock \showarticletitle{Dynamic client association for energy-aware
  hierarchical federated learning}. In \bibinfo{booktitle}{\emph{WCNC}}. IEEE,
  \bibinfo{pages}{1--6}.
\newblock


\bibitem[Yoon et~al\mbox{.}(2021)]%
        {yoon2021online}
\bibfield{author}{\bibinfo{person}{Jaehong Yoon}, \bibinfo{person}{Divyam
  Madaan}, \bibinfo{person}{Eunho Yang}, {and} \bibinfo{person}{Sung~Ju
  Hwang}.} \bibinfo{year}{2021}\natexlab{}.
\newblock \showarticletitle{Online coreset selection for rehearsal-based
  continual learning}.
\newblock \bibinfo{journal}{\emph{arXiv preprint arXiv:2106.01085}}
  (\bibinfo{year}{2021}).
\newblock


\bibitem[You et~al\mbox{.}(2022)]%
        {you2022triple}
\bibfield{author}{\bibinfo{person}{Linlin You}, \bibinfo{person}{Sheng Liu},
  \bibinfo{person}{Yi Chang}, {and} \bibinfo{person}{Chau Yuen}.}
  \bibinfo{year}{2022}\natexlab{}.
\newblock \showarticletitle{A triple-step asynchronous federated learning
  mechanism for client activation, interaction optimization, and aggregation
  enhancement}.
\newblock \bibinfo{journal}{\emph{IEEE Internet of Things Journal}}
  (\bibinfo{year}{2022}).
\newblock


\bibitem[Zhang et~al\mbox{.}(2021)]%
        {zhang2021csafl}
\bibfield{author}{\bibinfo{person}{Yu Zhang}, \bibinfo{person}{Morning Duan},
  \bibinfo{person}{Duo Liu}, \bibinfo{person}{Li Li}, \bibinfo{person}{Ao Ren},
  \bibinfo{person}{Xianzhang Chen}, \bibinfo{person}{Yujuan Tan}, {and}
  \bibinfo{person}{Chengliang Wang}.} \bibinfo{year}{2021}\natexlab{}.
\newblock \showarticletitle{CSAFL: A clustered semi-asynchronous federated
  learning framework}. In \bibinfo{booktitle}{\emph{IJCNN}}. IEEE,
  \bibinfo{pages}{1--10}.
\newblock


\bibitem[Zhong et~al\mbox{.}(2022)]%
        {zhong2022flee}
\bibfield{author}{\bibinfo{person}{Zhengyi Zhong}, \bibinfo{person}{Weidong
  Bao}, \bibinfo{person}{Ji Wang}, \bibinfo{person}{Xiaomin Zhu}, {and}
  \bibinfo{person}{Xiongtao Zhang}.} \bibinfo{year}{2022}\natexlab{}.
\newblock \showarticletitle{FLEE: A hierarchical federated learning framework
  for distributed deep neural network over cloud, edge and end device}.
\newblock \bibinfo{journal}{\emph{ACM TIST}} (\bibinfo{year}{2022}).
\newblock


\bibitem[Zhou et~al\mbox{.}(2021)]%
        {zhou2021tea}
\bibfield{author}{\bibinfo{person}{Chendi Zhou}, \bibinfo{person}{Hao Tian},
  \bibinfo{person}{Hong Zhang}, \bibinfo{person}{Jin Zhang},
  \bibinfo{person}{Mianxiong Dong}, {and} \bibinfo{person}{Juncheng Jia}.}
  \bibinfo{year}{2021}\natexlab{}.
\newblock \showarticletitle{TEA-fed: time-efficient asynchronous federated
  learning for edge computing}. In \bibinfo{booktitle}{\emph{ACM CF}}.
  \bibinfo{pages}{30--37}.
\newblock


\bibitem[Zhu et~al\mbox{.}(2022)]%
        {zhu2022online}
\bibfield{author}{\bibinfo{person}{Hongbin Zhu}, \bibinfo{person}{Yong Zhou},
  \bibinfo{person}{Hua Qian}, \bibinfo{person}{Yuanming Shi},
  \bibinfo{person}{Xu Chen}, {and} \bibinfo{person}{Yang Yang}.}
  \bibinfo{year}{2022}\natexlab{}.
\newblock \showarticletitle{Online client selection for asynchronous federated
  learning with fairness consideration}.
\newblock \bibinfo{journal}{\emph{IEEE Transactions on Wireless
  Communications}} (\bibinfo{year}{2022}).
\newblock


\end{thebibliography}

\appendix
\section{Proof of Theorem~\ref{thm:async_fl_1}}
\label{sec:thm_async_fl_1_proof}

In the main paper, we establish the convergence of Async-HFL under the three assumptions. 
Our analysis extends the proof in~\cite{xie2019asynchronous} from a single level of asynchrony to two levels.
The proof involves bounding the deviation of cloud loss function $L_N$ between steps in terms of its gradient norm $\nabla L_N$. We proceed from the lowest level, the device, establishing progress across device steps $e$. Then, using the progress after $E$ device steps, we can then bound progress at the next level, which is the gateway. Note that moving to the next level, the cloud is much simpler as the update steps and assumptions for cloud and gateway level are similar. We now describe the complete proof starting from the progress on any device in the next subsection.

\subsection{Device steps}
On each device, we only perform SGD steps, therefore, we can utilize Assumptions~\ref{assump:grad_bdd} and ~\ref{assump:rho_large} to bound the progress after each gradient update.

\begin{lemma}[Single-step Device]\label{lemma:device_1step}
Under Assumptions~\ref{assump:grad_bdd} and ~\ref{assump:rho_large}, the progress in terms of  $L_N$ after one device step is given by  
\begin{equation}
\begin{aligned}
    \E[L_N(\bm{\omega}_{\tau,\zeta,e}^{i})] \leq& \E[L_N(\bm{\omega}_{\tau,\zeta,e-1}^{i})]  
    -\gamma c\E[\norm{\nabla L_N(\bm{\omega}_{\tau,\zeta,e-1}^{i})}^2]\\
    &+ \gamma^2 \cO(\rho E^2 V_2) \quad \forall \tau,\zeta\geq 0, e \geq 1, i\in \mathcal{N}
\end{aligned}
\end{equation}
\end{lemma}
\begin{proof}
Since the loss is $L$-smooth, $L_N(\bm{\omega})$ and $g_{N,\bm{\omega}'}(\bm{\omega})$ are $L$-smooth and $(2L + 2\rho)$-smooth. To simplify the proof, we drop the subscript $\tau$ and superscript $i$ from the following steps.
Since $g_{N,\bm{\omega'}}(\bm{\omega}) \geq L_N(\bm{\omega}) \forall \bm{\omega},\bm{\omega'}$, we have
\begin{align*}
    \E[L_N(\bm{\omega}_{\zeta,e})] &\leq \E[g_{N,\bm{\omega}_{\zeta}}(\bm{\omega}_{\zeta,e})]\\
    &\leq \E[g_{N,\bm{\omega}_{\zeta}}(\bm{\omega}_{\zeta,e-1})]+ \gamma^2(L + \rho) \E[\norm{\nabla g_{\bm{\omega}_{\zeta}}(\bm{\omega}_{\zeta,e-1})}^2] \\
    & \quad - \gamma \E[\lin{\nabla g_{N,\bm{\omega}_{\zeta}}(\bm{\omega}_{\zeta,e-1}), \nabla g_{\bm{\omega}_{\zeta}}(\bm{\omega}_{\zeta,e-1})}]\\
    &\leq \E[L_N(\bm{\omega}_{\zeta,e-1})] + \frac{\rho}{2}\E[\norm{\bm{\omega}_{\zeta} - \bm{\omega}_{\zeta, e-1}}^2] \\
    & \quad + \gamma^2(L + \rho) \E[\norm{\nabla g_{\bm{\omega}_{\zeta}}(\bm{\omega}_{\zeta,e-1})}^2]\\
    & \quad - \gamma \E[\lin{\nabla g_{N,\bm{\omega}_{\zeta}}(\bm{\omega}_{\zeta,e-1}), \nabla g_{\bm{\omega}_{\zeta}}(\bm{\omega}_{\zeta,e-1})}]\\
    &\leq \E[L_N(\bm{\omega}_{\zeta,e-1})] + \frac{\rho \gamma^2 E^2 V_2}{2}+ \gamma^2(L + \rho)V_2 \\
    & \quad - \gamma \E[\lin{\nabla g_{N,\bm{\omega}_{\zeta}}(\bm{\omega}_{\zeta,e-1}), \nabla g_{\bm{\omega}_{\zeta}}(\bm{\omega}_{\zeta,e-1})}]
\end{align*}
For the second step, we use smoothness of $g_{N,\bm{\omega}_{\zeta}}$. For the third step, we expand $g_{N,\bm{\omega}_{\zeta}}$ and for the fourth step, we use Assumption~\ref{assump:grad_bdd} to bound $\E[\norm{\nabla g_{\bm{\omega}_{\zeta}}(\bm{\omega}_{\zeta,e-1})}^2]$ and $\E[\norm{\bm{\omega}_{\zeta} - \bm{\omega}_{\zeta, e-1}}^2]$. 

Note that, $\E[\norm{\bm{\omega}_{\zeta} - \bm{\omega}_{\zeta, e-1}}^2] =\E[\norm{\sum_{r=1}^{e-1} \gamma \nabla g_{\bm{\omega}_{\zeta}}(\bm{\omega}_{\zeta,r})}^2] \leq \gamma^2(e-1) \sum_{r=1}^{e-1}\E[\norm{\nabla g_{\bm{\omega}_{\zeta}}(\bm{\omega}_{\zeta,r})}^2]  \leq \gamma^2(e-1)^2 V_2 \leq \gamma^2 E^2 V_2$

We now utilize Assumption~\ref{assump:rho_large} to bound the remaining terms and introduce $\nabla L_N$. Consider the following inequality for some $c >0$,
\begin{align*}
    &\E[\lin{\nabla g_{N,\bm{\omega}_{\zeta}}(\bm{\omega}_{\zeta,e-1}), \nabla g_{\bm{\omega}_{\zeta}}(\bm{\omega}_{\zeta,e-1})}] - c\E[\norm{\nabla L_N(\bm{\omega}_{\zeta,e-1})}^2]\\
    &= \E[\langle\nabla L_N(\bm{\omega}_{\zeta,e-1}) + \rho(\bm{\omega}_{\zeta,e-1} - \bm{\omega}_{\zeta}), \\
    & \quad \nabla L(\bm{\omega}_{\zeta,e-1}) + \rho(\bm{\omega}_{\zeta,e-1} - \bm{\omega}_{\zeta})\rangle]  - c\E[\norm{\nabla L_N(\bm{\omega}_{\zeta,e-1})}^2]\\
    &= \E[\lin{\nabla L_N(\bm{\omega}_{\zeta,e-1}) ,\nabla L(\bm{\omega}_{\zeta,e-1})}] + \rho^2\E[\norm{\bm{\omega}_{\zeta,e-1} - \bm{\omega}_{\zeta}}^2] \\
    & \quad +\rho \E[\lin{\nabla L_N(\bm{\omega}_{\zeta,e-1}) + \nabla L(\bm{\omega}_{\zeta,e-1}),\bm{\omega}_{\zeta,e-1}- \bm{\omega}_{\zeta} }]\\
    & \quad - c\E[\norm{\nabla L_N(\bm{\omega}_{\zeta,e-1})}^2]\\
    &\geq  - \frac{1}{2}\E[\norm{\nabla L_N(\bm{\omega}_{\zeta,e-1})}^2] - \frac{1}{2}\E[\norm{\nabla L(\bm{\omega}_{\zeta,e-1})}^2] \\
    & \quad +\rho^2\E[\norm{\bm{\omega}_{\zeta,e-1} - \bm{\omega}_{\zeta}}^2]- \frac{\rho}{2}\E[\norm{\bm{\omega}_{\zeta,e-1}- \bm{\omega}_{\zeta} }^2] \\
    & \quad -\frac{\rho}{2} \E[\norm{\nabla L_N(\bm{\omega}_{\zeta,e-1}) + \nabla L(\bm{\omega}_{\zeta,e-1})}^2]- c\E[\norm{\nabla L_N(\bm{\omega}_{\zeta,e-1})}^2]\\
    &\geq  - \frac{1}{2}\E[\norm{\nabla L_N(\bm{\omega}_{\zeta,e-1})}^2] - \frac{1}{2}\E[\norm{\nabla L(\bm{\omega}_{\zeta,e-1})}^2] \\
    & \quad + \rho^2\E[\norm{\bm{\omega}_{\zeta,e-1} - \bm{\omega}_{\zeta}}^2]- \frac{\rho}{2}\E[\norm{\bm{\omega}_{\zeta,e-1}- \bm{\omega}_{\zeta} }^2] \\
    & \quad -\rho \E[\norm{\nabla L_N(\bm{\omega}_{\zeta,e-1})}^2] -\rho \E[\norm{\nabla L(\bm{\omega}_{\zeta,e-1})}^2] \\
    & \quad - c\E[\norm{\nabla L_N(\bm{\omega}_{\zeta,e-1})}^2]\\
    &\geq    -V_1(\rho + \frac{1}{2} +c)  -  V_2(\rho+  \frac{1}{2}) +     \bigl(\rho^2 - \frac{\rho}{2}\bigr)\E[\norm{\bm{\omega}_{\zeta,e-1} - \bm{\omega}_{\zeta}}^2] 
\end{align*}
In the first two steps, we expand the definitions of $g_{N,\bm{\omega}}$ and $g_{\bm{\omega}}$. In the second step and third step, we use $\lin{a,b}\geq -\frac{\norm{a}^2}{2} - \frac{\norm{b}^2}{2}$. In the fourth step, we use Assumption~\ref{assump:grad_bdd} to obtain the bounds. Finally, by Assumption~\ref{assump:rho_large}, we have a $c>0$ such that the last inequality is always $\geq 0$. Therefore,
\begin{align*}
    \E[\lin{\nabla g_{N,\bm{\omega}_{\zeta}}(\bm{\omega}_{\zeta,e-1}), \nabla g_{\bm{\omega}_{\zeta}}(\bm{\omega}_{\zeta,e-1})}] \leq  c\E[\norm{\nabla L_N(\bm{\omega}_{\zeta,e-1})}^2]
\end{align*}
Using this inequality, we have proved the lemma.
\begin{align*}
    \E[L_N(\bm{\omega}_{\tau,\zeta,e}^{i})] \leq& \E[L_N(\bm{\omega}_{\tau,\zeta,e-1}^{i})] -\gamma c\E[\norm{\nabla L_N(\bm{\omega}_{\tau,\zeta,e-1}^{i})}^2] \\
    &+ \gamma^2 \rho \cO(E^2 V_2)
\end{align*}
\end{proof}
Using the progress made in a single step, we can bound the update for all local steps at a single device, by summing the above lemma over $E$ steps.
\begin{lemma}[All-steps Device]\label{lemma:device_full}
Under Assumptions~\ref{assump:bdd_delay} and ~\ref{assump:rho_large}, after $E$ steps at the device level, the progress made in the objective function is given by 
\begin{equation}
\begin{aligned}
    \E[L_N(\bm{\omega}_{\tau,\zeta,E}^{i})] \leq& \E[L_N(\bm{\omega}_{\tau,\zeta}^{i})]  
    -\gamma c\sum_{e=1}^E\E[\norm{\nabla L_N(\bm{\omega}_{\tau,\zeta,e-1}^{i})}^2]\\
    &+ \gamma^2 \cO(\rho E^3 V_2) \quad \forall \tau,\zeta \geq 0,  i\in \mathcal{N}
\end{aligned}
\end{equation}
\end{lemma}

\newpage
\subsection{Gateway steps}
When we move a level up the hierarchy, the update is simply a convex combination of the current model at the gateway and the trained model received from a device. Note that this is the first place in the proof where we need to handle asynchrony. Using Assumption~\ref{assump:bdd_delay}, we first explicitly derive the delay recursion in the following intermediate Lemma.

\begin{lemma}[Gateway delay recursion]\label{lemma:gateway_intermediate}
Let $\bm{\omega}_{\tau,\zeta,E}^j$ be the update received at gateway $j$ at step $z$, then for $\mathcal{R}_{\tau,z}^{j} \triangleq \E[\norm{\bm{\omega}_{\tau,\zeta}^j - \bm{\omega}_{\tau,z-1}}^2]$, under Assumptions~\ref{assump:grad_bdd} and~\ref{assump:bdd_delay}, we can bound the norm of update at gateway at gateway step $z$ as,
\begin{align}
\mathcal{R}_{\tau,z}^{j} \leq & K_g \beta^2 (\gamma^2 \cO(K_g E^2 V_2) + 4\sum_{r=1}^{K_g - 1} \mathcal{R}_{\tau,z-r}^j), \, \forall \tau\geq 0, z\geq1,  j\in \mathcal{G} 
\end{align}
\end{lemma}
\begin{proof}
We drop the superscript $j$ and subscript $\tau$ and only consider the update at the given machine. From Assumption~\ref{assump:bdd_delay}, we can see that $z - 1 -\zeta \leq K_g$, therefore, there have been at most $K_g$ gateway updates between $\zeta$ and $z-1$. Utilizing this fact, we can rewrite $\mathcal{R}_z$ as
\begin{align*}
    \mathcal{R}_{z}=& \E[\norm{\sum_{r=0}^{z-\zeta-2}(\bm{\omega}_{\zeta + r+1} - \bm{\omega}_{\zeta + r})}^2]\\
    \leq & (z - \zeta -2) \sum_{r=0}^{z-\zeta-2}\E[\norm{\bm{\omega}_{\zeta + r+1} - \bm{\omega}_{\zeta + r}}^2]
\end{align*}
We utilize triangle inequality to split the terms inside the norm. 
Let $\bm{\omega}_{\zeta_{z-r},E}$ be the update received at step $z-r$ at the gateway then, we have,
\begin{align}
    \mathcal{R}_{z}\leq& (z - \zeta -1)\beta^2 \sum_{r=1}^{z-\zeta-1}\E[\norm{\bm{\omega}_{\zeta_{z-r +1},E} - \bm{\omega}_{z - r}}^2]\label{eq:agg_recursion_int}
\end{align}
Now, consider a single term in the summation given in Eq.~\eqref{eq:agg_recursion_int}.
\begin{align*}
\E[\norm{\bm{\omega}_{\zeta_{z-r +1},E} - \bm{\omega}_{z - r}}^2] &\leq 2\E[\norm{\bm{\omega}_{\zeta_{z-r+1},E} - \bm{\omega}_{\zeta_{z-r+1}}}^2] \\
& \quad + 2\E[\norm{\bm{\omega}_{z-r} - \bm{\omega}_{\zeta_{z-r+1}}}^2]\\
&\leq \gamma^2 \cO(E^2 V_2) + 2 \mathcal{R}_{z-r+1}
\end{align*}
We use the fact that $E$ steps on a device can be represented as sum of $E$ gradient updates where norm of each gradient is bounded by $\cO(V_2)$ and use a triangle inequality from Assumption~\ref{assump:grad_bdd}. 
Plugging this into Eq.~\eqref{eq:agg_recursion_int}, we obtain,
\begin{align*}
    \mathcal{R}_z \leq& (z - \zeta -2)\beta^2\sum_{r=1}^{z-\zeta - 1} (\gamma^2 \cO(E^2 V_2) + 2\mathcal{R}_{z-r})\\
    \leq& K_g \beta^2 (\gamma^2 \cO(K_g E^2 V_2) + 4\sum_{r=1}^{K_g - 1} \mathcal{R}_{z-r})
\end{align*}
We finally use the fact that delay is bounded by $K_g$ from Assumption~\ref{assump:bdd_delay}.

\end{proof}

Handling the delay recursion, by means of the above Lemma, we utilize the update step on a single gateway and the progress made by a training model at a given device, from Lemma~\ref{lemma:device_full}, to obtain the single-step progress at the gateway level.
\begin{lemma}[Single-step Gateway]\label{lemma:gateway_1step}
Under Assumptions~\ref{assump:grad_bdd},\ref{assump:bdd_delay} and ~\ref{assump:rho_large}, for a single step at the gateway, we have
\begin{equation}
\begin{aligned}
    \E[L_N(\bm{\omega}_{\tau,z}^{j}) - L_N(\bm{\omega}_{\tau,z-1}^{j})] \leq & -\gamma\beta c \sum_{e=0}^{E-1}\E[\norm{\nabla L_N(\bm{\omega}_{\tau,\zeta,e}^{i_{j,z}})}^2]+ \Delta_z\\
    &  \quad \forall \tau\geq 0,z\geq 1, j\in \mathcal{G}
\end{aligned}
\end{equation}
where the device $i_{j,z}$ was used for the $z^{th}$ gateway update and 
\begin{align*}
    \Delta_z = \beta\gamma^2\cO(\rho E^3 V_2) +\beta\cO(\sqrt{V_1})\sqrt{\mathcal{R}_{\tau,z}^j} +\beta\cO(L + \rho)\mathcal{R}_{\tau,z}^j
\end{align*}
\end{lemma}
\begin{proof}
We will again omit $\tau$ and $j$ to simplify the proof. Consider the function value after a single update at gateway at step $z$ upon receiving the update $\bm{\omega}_{\zeta,E}$ from the device.
\begin{align}
    &\E[L_N(\bm{\omega}_{z}) - L_N(\bm{\omega}_{z-1})] = \E[g_N(\bm{\omega}_{z};\bm{\omega}_{z-1}) - L_N(\bm{\omega}_{z-1})]\\
    &\leq \E[(1-\beta) g_N(\bm{\omega}_{z-1};\bm{\omega}_{z-1}) + \beta g_N(\bm{\omega}_{\zeta,E};\bm{\omega}_{z-1}) - L_N(\bm{\omega}_{z-1})]\\
    &\leq \beta\E[L_N(\bm{\omega}_{\zeta,E}) - L_N(\bm{\omega}_{z-1})] + \frac{\beta\rho}{2}\E[\norm{\bm{\omega}_{\zeta,E} - \bm{\omega}_{z-1}}^2]\\
    &\leq \beta\E[L_N(\bm{\omega}_{\zeta,E}) - L_N(\bm{\omega}_{\zeta})] + \beta\E[L_N(\bm{\omega}_{\zeta})- L_N(\bm{\omega}_{z-1})] \nonumber \\
    & \quad + \frac{\beta\rho}{2}\E[\norm{\bm{\omega}_{\zeta,E} - \bm{\omega}_{z-1}}^2]\label{eq:agg_1step_int}
\end{align}
We first use the fact that $g_N(\bm{\omega}_1;\bm{\omega}_2) \geq L_N(\bm{\omega}_1),\forall \bm{\omega}_1,\bm{\omega}_2 \in \R^d$. Then, we utilize convexity of $g_N$ followed by expanding $g_N$ into its components.  For the first term in Eq.~\eqref{eq:agg_1step_int}, we can use Lemma~\ref{lemma:device_full}, as it is the difference of function values after $E$ steps at the device.

For the second term in Eq.~\eqref{eq:agg_1step_int}, we can use the triangle inequality and bound the gradient norm from Assumption~\ref{assump:grad_bdd}, similar to the proof of Lemma~\ref{lemma:gateway_intermediate}, to obtain,
\begin{align*}
    \E[\norm{\bm{\omega}_{\zeta,E} - \bm{\omega}_{z-1}}^2] &\leq 2\E[\norm{\bm{\omega}_{\zeta,E} - \bm{\omega}_{\zeta}}^2] + 2\mathcal{R}_{z}\\
    &\leq 2\gamma^2 \cO(E^2 V_2) + 2\mathcal{R}_{z}
\end{align*}

We handle the first term in ~\eqref{eq:agg_1step_int} separately using $L$-smoothness of $L_N$.
\begin{align*}
    \E[L_N(\bm{\omega}_{\zeta})- L_N(\bm{\omega}_{z-1})] \leq& \E[\lin{\nabla L_N(\bm{\omega}_{z-1}),\bm{\omega}_{\zeta} - \bm{\omega}_{\zeta}}] \\
    & + \frac{L}{2}\E[\norm{\bm{\omega}_{\zeta} - \bm{\omega}_{\zeta}}^2]\\
    \leq& \E[\norm{\nabla L_N(\bm{\omega}_{z-1})}\norm{\bm{\omega}_{\zeta} - \bm{\omega}_{\zeta}}] + \frac{L}{2}\mathcal{R}_z\\
    \leq& \cO(\sqrt{V_1}) \E[\norm{\bm{\omega}_{\zeta} - \bm{\omega}_{\zeta}}] + \frac{L}{2}\mathcal{R}_z\\
    \leq& \cO(\sqrt{V_1}) \sqrt{\mathcal{R}_z} + \frac{L}{2}\mathcal{R}_z
\end{align*}
We use Cauchy-Schwartz inequality to bound $\lin{a,b} \leq \norm{a}\norm{b},\forall a,b\in \R^d$. Further, we utilize Assumption~\ref{assump:grad_bdd} to bound $\norm{\nabla g_N(\cdot)}$ and then use Jensen's inequality to obtain $\E[\norm{a}]\leq \sqrt{\R[\norm{a}^2]}$.

Substituting these values in Eq.~\eqref{eq:agg_1step_int}, we complete the proof.
\end{proof}
Using the above Lemmas~\ref{lemma:gateway_1step} and ~\ref{lemma:gateway_intermediate}, we can bound the progress of gateway after $Z$ steps. For this purpose, we need to bound the sum of the recursion terms in Lemma~\ref{lemma:gateway_intermediate}. The following Lemma provides this bound.
\begin{lemma}[Sum of gateway recursion]\label{lemma:agg_sum_recursion}
Under Assumptions~\ref{assump:grad_bdd} and \ref{assump:bdd_delay}, and when $\beta \leq \cO(K_g^{-3/2})$, we have,
\begin{align}
    \sum_{z=1}^Z \mathcal{R}_z \leq& \gamma^2 \beta^2 \cO(K_g^2 E^2 Z V_2 )\\
    \sum_{z=1}^Z \sqrt{\mathcal{R}_z} \leq& \gamma \beta \cO(K_g E Z\sqrt{V_2})
\end{align}
\end{lemma}
\begin{proof}
We first start with Lemma~\ref{lemma:gateway_intermediate} and sum over $z=1$ to $Z$
\begin{align}
    \sum_{z=1}^Z \mathcal{R}_z \leq& \gamma^2 \beta^2 \cO(K_g^2 E^2 Z V_2 ) + \beta^2K_g^2 \sum_{z=1}^Z \mathcal{R}_z\\  \sum_{z=1}^Z \mathcal{R}_z\leq & (1- \beta^2 K_g^2)^{-1}\gamma^2 \beta^2 \cO(K_g^2 E^2 Z V_2)
\end{align}
Here, summing over the recursion, we need to control the coefficient of the sum on RHS. If this coefficient is $<1$, then we are done, which is exactly the case when $\beta = \cO(K_g^{-3/2})$ as $\beta^2 K_g^2 = \cO(K_g^{-1})$ which can be made to be a constant less than $1$.

Now, for the sum over square roots, we first use Jensen's inequality to obtain, $\sqrt{\sum_{i=1}^n a_i} \leq \sum_{i=1}^n\sqrt{a_i},\forall a_i \geq 0$ on Lemma~\ref{lemma:gateway_intermediate}.
\begin{align}
    \sqrt{\mathcal{R}_z} \leq&  \gamma \beta \cO(K_g E \sqrt{V_2}) + \beta \sqrt{K_g}\sum_{r=1}^{K_g-1}\sqrt{\mathcal{R}_{z-r}}\\
    \sum_{z=1}^Z \sqrt{\mathcal{R}_z} \leq& \gamma \beta \cO(K_g E \sqrt{V_2}) + \beta K_g^{3/2} \sum_{z=1}^Z \sqrt{\mathcal{R}_z}\\
    \leq & (1- \beta K_g^{3/2})^{-1}\gamma \beta \cO(K_g E \sqrt{V_2})
\end{align}
Now, we again control the coefficient of the sum on RHS and for $\beta = \cO(K_g^{-3/2})$, this coefficient is a constant.

\end{proof}

\begin{lemma}[All-steps Gateway]\label{lemma:gateway_allstep}
From Lemmas~\ref{lemma:gateway_1step} and ~\ref{lemma:agg_sum_recursion}, with $\beta = \cO(K_g^{-3/2})$, we have,
\begin{align}
\E[L_N(\bm{\omega}_{\tau,Z}^{j}) - L_N(\bm{\omega}_{\tau}^{j})] \leq & -\gamma\beta c \sum_{z=1}^{Z}\sum_{e=0}^{E-1}\E[\norm{\nabla L_N(\bm{\omega}_{\tau,\zeta,e}^{i_{j,z}})}^2] + \Delta_\tau\nonumber\\
    &\quad \forall \tau\geq 0, j \in \mathcal{G}\nonumber
\end{align}
where 
\begin{align}
    \Delta_{\tau} &=\beta\gamma^2\cO(\rho E^3 Z V_2) + \gamma\beta^2\cO(K_g E Z\sqrt{V_1 V_2}) \nonumber \\
    &\quad + \gamma^2\beta^3 \cO((L + \rho)K_g^2 E^2 Z V_2)
\end{align}
\end{lemma}
\begin{proof}
We ignore $\tau$ and $j$ to simplify the proof. We can simply sum over the terms in Lemma~\ref{lemma:gateway_1step}, to obtain,
\begin{align}
        \E[L_N(\bm{\omega}_{\tau,Z}^{j}) - L_N(\bm{\omega}_{\tau}^{j})] \leq & -\gamma\beta c \sum_{z=1}^{Z}\sum_{e=0}^{E-1}\E[\norm{\nabla L_N(\bm{\omega}_{\tau,\zeta,e}^{i_{j,z}})}^2] \nonumber \\
        & + \sum_{z=1}^Z \Delta_z
\end{align}
Now, to compute the sum of $\Delta_z$ terms,
\begin{align*}
    \sum_{z=1}^Z \Delta_z &= \sum_{z=1}^Z (\beta\gamma^2\cO(\rho E^3 V_2) +\beta\cO(\sqrt{V_1})\sqrt{\mathcal{R}_{z}} +\beta\cO(L + \rho)\mathcal{R}_{z})\\
    &= \beta\gamma^2\cO(\rho E^3 Z V_2) + \beta\cO(\sqrt{V_1})\sum_{z=1}^Z \sqrt{\mathcal{R}_z} + \beta \cO(L + \rho) \sum_{z=1}^Z \mathcal{R}_z\\
    &= \beta\gamma^2\cO(\rho E^3 Z V_2) + \gamma\beta^2\cO(K_g E Z\sqrt{V_1 V_2})\\
    &\quad + \gamma^2\beta^3 \cO((L + \rho)K_g^2 E^2 Z V_2)
\end{align*}
\end{proof}

\subsection{Cloud steps}
In this section, we try to bound the progress made in a single cloud step. First, we bound certain quantities which will appear in the proof.
\begin{lemma}[Gateway iterate difference]\label{lemma:agg_allstep_iterates}
Under Assumptions~\ref{assump:grad_bdd} and \ref{assump:bdd_delay}, we have,
\begin{align}
    \E[\norm{\bm{\omega}_{\tau,Z} - \bm{\omega}_{\tau}}^2] \leq  \gamma^2\beta^2(\cO(Z^2 E^2 V_2) + \beta^2 \cO(K_g^2 E^2 Z^2 V_2))
\end{align}
\end{lemma}
\begin{proof}
Note that this term corresponds to the difference in iterates after $Z$ steps on gateway on model received at cloud step $\tau$. We can split this difference into a telescopic sum of single steps on the gateway and then use triangle inequality.
\begin{align}
    \E[\norm{\bm{\omega}_{\tau,Z} - \bm{\omega}_{\tau}}^2] &= \E[\norm{\sum_{r=1}^{Z}\bm{\omega}_{\tau,r} - \bm{\omega}_{\tau,r-1}}^2]\\
    &\leq Z \sum_{r=1}^{Z}\E[\norm{\bm{\omega}_{\tau,r} - \bm{\omega}_{\tau,r-1}}^2]\\
    &\leq Z \beta^2 \sum_{r=1}^{Z}\E[\norm{\bm{\omega}_{\tau,\zeta_r,E} - \bm{\omega}_{\tau,r-1}}^2]\label{eq:cloud_rec1_int}
\end{align}
Here, we assume that $\bm{\omega}_{\tau,\zeta_r,E}$ is the iterate received at $r^{th}$ step on the gateway. Now, we can use arguments similar to Lemma~\ref{lemma:gateway_allstep} to establish a recursion for each term in the sum in Eq.~\eqref{eq:cloud_rec1_int}.
\begin{align*}
\E[\norm{\bm{\omega}_{\tau,\zeta_r,E} - \bm{\omega}_{\tau,r-1}}^2] &\leq \E[\norm{\bm{\omega}_{\tau,\zeta_r,E} - \bm{\omega}_{\tau,\zeta_r}+ \bm{\omega}_{\tau,\zeta_r} -  \bm{\omega}_{\tau,r-1}}^2]\\
&\leq 2\E[\norm{\bm{\omega}_{\tau,\zeta_r,E} - \bm{\omega}_{\tau,\zeta_r}}^2] \\
& \quad + 2\E[\norm{\bm{\omega}_{\tau,\zeta_r} -  \bm{\omega}_{\tau,r-1}}^2]\\
&\leq 2\gamma^2\cO(E^2 V_2)+ 2 \mathcal{R}_r
\end{align*}
Plugging the above expression into Eq.~\eqref{eq:cloud_rec1_int}, and using Lemma~\ref{lemma:agg_sum_recursion} to bound $\sum_{z=1}^Z \mathcal{R}_z$, we obtain the result.
\end{proof}

\begin{lemma}[Cloud delay recursion]\label{lemma:cloud_recursion}
Under Assumptions~\ref{assump:grad_bdd},\ref{assump:rho_large} and \ref{assump:bdd_delay}, where we define $\cQ_h = \E[\norm{\bm{\omega}_\tau - \bm{\omega}_{h-1}}^2]$, where $\bm{\omega}_{\tau,Z}$ is the iterate received at cloud step $h$, then
\begin{equation}
\begin{aligned}
\cQ_h &\leq \gamma^2\alpha^2\beta^2(\cO(K_c^2 Z^2 E^2 V_2) + \beta^2 \cO(K_c^2 K_g^2 E^2 Z^2 V_2)) \\
& \quad +2\alpha^2 K_c\sum_{r=1}^{K_c  - 1}\cQ_{h-r}
\end{aligned}
\end{equation}
\end{lemma}
\begin{proof}
By Assumption~\ref{assump:bdd_delay}, $h-1-\tau \leq K_c$, therefore, there have been atmost $K_c$ cloud steps between $\bm{\omega}_\tau$ and $\bm{\omega}_{h-1}$ and we can express this difference as a telescopic sum of consecutive steps on the cloud followed by triangle inequality.
\begin{align}
    \cQ_h &= \E[\norm{\sum_{r=0}^{h-\tau}(\bm{\omega}_{h-r} - \bm{\omega}_{h-r-1}) }^2]\\
    &\leq (h-\tau)\sum_{r=0}^{h-\tau}\E[\norm{\bm{\omega}_{h -r} - \bm{\omega}_{h-r-1} }^2]\label{eq:cloud_recursion_int}
\end{align}
Now, we can analyze each term in the sum in Eq.~\eqref{eq:cloud_recursion_int} separately. Let $\bm{\omega}_{\tau_{h-r},Z}$ be the model received at the cloud at step $h-r$, then
\begin{align*}
    \E[\norm{\bm{\omega}_{h -r} - \bm{\omega}_{h-r-1}}^2]&\leq \alpha^2\E[\norm{\bm{\omega}_{\tau_{h -r},Z} - \bm{\omega}_{h-r-1}}^2]\\
    &\leq \alpha^2\E[\norm{\bm{\omega}_{\tau_{h -r},Z}-\bm{\omega}_{\tau_{h -r}}+\bm{\omega}_{\tau_{h -r}} - \bm{\omega}_{h-r-1}}^2]\\
    &\leq 2\alpha^2\bigg(\E[\norm{\bm{\omega}_{\tau_{h -r},Z}-\bm{\omega}_{\tau_{h -r}}}^2 \\
    &\quad+\E[\norm{\bm{\omega}_{\tau_{h -r}} - \bm{\omega}_{h-r-1}}^2]\bigg)\\
    &\leq 2\alpha^2\bigg(\gamma^2\beta^2(\cO(Z^2 E^2 V_2) + \beta^2 \cO(K_g^2 E^2 Z^2 V_2)) \\
    &\quad + \cQ_{h-r}\bigg)
\end{align*}
We utilize the update equation at the cloud with convexity of the $\norm{\cdot}$ along with the triangle inequality and Lemma~\ref{lemma:agg_allstep_iterates} in the above steps.

Plugging this into Eq.~\eqref{eq:cloud_recursion_int} and using Assumption~\ref{assump:bdd_delay}, we get our result.
\begin{align}
    \cQ_h &\leq 2(h-\tau)\alpha^2\bigg((h-\tau)\gamma^2\beta^2(\cO(Z^2 E^2 V_2) + \beta^2 \cO(K_g^2 E^2 Z^2 V_2)) \nonumber\\
    &\quad+\sum_{r=1}^{h-\tau}\cQ_{h-r}\bigg)\\
    &\leq \gamma^2\alpha^2\beta^2(\cO(K_c^2 Z^2 E^2 V_2) + \beta^2 \cO(K_c^2 K_g^2 E^2 Z^2 V_2)) \nonumber\\
    &\quad+2\alpha^2 K_c\sum_{r=1}^{K_c  - 1}\cQ_{h-r}
\end{align}
\end{proof}

Using these lemmas, we can analyze the progress made in a single step at the cloud.

\begin{lemma}[Single-step Cloud]\label{lemma:cloud_1step}
Under Assumptions~\ref{assump:grad_bdd},\ref{assump:rho_large} and \ref{assump:bdd_delay}, after a single step on the cloud, we have,
\begin{align}
        \E[L_N(\bm{\omega}_{h}) - L_N(\bm{\omega}_{h-1})] &\leq  -\gamma\alpha\beta c \sum_{z=0}^{Z-1}\sum_{e=0}^{E-1}\E[\norm{\nabla L_N(\bm{\omega}_{\tau,\zeta,e})}^2] \nonumber \\
        &\quad+\Xi_h , \quad \forall h > 0
\end{align}
where 
\begin{align*}
\begin{split}
\Xi_h =& \alpha\beta\gamma^2\cO(\rho E^3 Z V_2) + \alpha\gamma\beta^2\cO(K_g E Z\sqrt{V_1 V_2}) \\
    & + \alpha\gamma^2\beta^3 \cO((L + \rho)K_g^2 E^2 Z V_2)\\
    &+ \alpha\rho\gamma^2\beta^2(\cO(Z^2 E^2 V_2) + \beta^2 \cO(K_g^2 E^2 Z^2 V_2)) \\
    &+ \cO(\alpha\sqrt{V_1}) \sqrt{\cQ_h} + \alpha\cO(\rho + L)\cQ_h
\end{split}
\end{align*}

\end{lemma}
\begin{proof}
The proof is similar to that of Lemma~\ref{lemma:gateway_1step} until ~\ref{eq:agg_1step_int}, so we state the corresponding equation for a single cloud step.
\begin{align}
    &\E[L_N(\bm{\omega}_{h}) - L_N(\bm{\omega}_{h-1})] \\
    &\leq \alpha\E[L_N(\bm{\omega}_{\tau,Z}) - L_N(\bm{\omega}_{\tau})] + \alpha\E[L_N(\bm{\omega}_{\tau})- L_N(\bm{\omega}_{h-1})] \nonumber \\
    &\quad+ \frac{\alpha\rho}{2}\E[\norm{\bm{\omega}_{\tau,Z} - \bm{\omega}_{h-1}}^2]\label{eq:cloud_1step_int}
\end{align}
Now, we analyze the terms in above equation similar to Lemma~\ref{lemma:gateway_1step}
 For the second term in Eq.~\eqref{eq:cloud_1step_int}, we can use Lemma~\ref{lemma:gateway_allstep}, as it is the difference of function values after $Z$ steps at the gateway.

For the third term in Eq.~\eqref{eq:cloud_1step_int}, we can use the triangle inequality and use Lemmas~\ref{lemma:agg_allstep_iterates} and \ref{lemma:cloud_recursion},
\begin{align*}
    \E[\norm{\bm{\omega}_{\tau,Z} - \bm{\omega}_{h-1}}^2] &\leq 2\E[\norm{\bm{\omega}_{\tau,Z} - \bm{\omega}_{\tau}}^2] + 2\cQ_h\\
    &\leq \gamma^2\beta^2(\cO(Z^2 E^2 V_2) + \beta^2 \cO(K_g^2 E^2 Z^2 V_2))+ 2\cQ_{h}
\end{align*}

We  handle the first term in Eq.~\eqref{eq:cloud_1step_int} separately using $L$-smoothness of $L_N$.
\begin{align*}
    \E[L_N(\bm{\omega}_{\tau})- L_N(\bm{\omega}_{h-1})] \leq& \E[\lin{\nabla L_N(\bm{\omega}_{h-1}),\bm{\omega}_{\tau} - \bm{\omega}_{h-1}} \\
    &+ \frac{L}{2}\E[\norm{\bm{\omega}_{\tau} - \bm{\omega}_{h-1}}^2]\\
    \leq& \E[\norm{\nabla L_N(\bm{\omega}_{h-1})}\norm{\bm{\omega}_{\tau} - \bm{\omega}_{h-1}}] + \frac{L}{2}\cQ_h\\
    \leq& \cO(\sqrt{V_1}) \E[\norm{\bm{\omega}_{\tau} - \bm{\omega}_{h-1}}] + \frac{L}{2}\cQ_h\\
    \leq& \cO(\sqrt{V_1}) \sqrt{\cQ_h} + \frac{L}{2}\cQ_h
\end{align*}
We use Cauchy-Schwartz inequality to bound $\lin{a,b} \leq \norm{a}\norm{b},\forall a,b\in \R^d$. Further, we utilize Assumption~\ref{assump:grad_bdd} to bound $\norm{\nabla g_N(\cdot)}$ and then use Jensen's inequality to obtain $\E[\norm{a}]\leq \sqrt{\R[\norm{a}^2]}$.

Substituting these values in Eq.~\eqref{eq:cloud_1step_int}, we complete the proof.

\end{proof}
We now bound the sum of the recursion terms $\cQ_h$ and $\sqrt{\cQ_h}$.
\begin{lemma}[ Sum of Cloud recursion ]\label{lemma:cloud_sum_recursion}
Under Assumptions~\ref{assump:grad_bdd},\ref{assump:bdd_delay} and \ref{assump:rho_large}, and when $\alpha \leq \cO(K_c^{-3/2})$, we have,

\begin{align}
    \sum_{h=1}^H \cQ_h \leq& \gamma^2\alpha^2\beta^2(\cO(H K_c^2 Z^2 E^2 V_2)+ \beta^2 \cO(H K_c^2 K_g^2 E^2 Z^2 V_2))\\
    \sum_{h=1}^H \sqrt{\cQ_h} \leq& \gamma\alpha\beta(\cO(H K_c Z E \sqrt{V_2}) + \beta \cO(H K_c K_g E Z \sqrt{V_2}))
\end{align}
\end{lemma}
\begin{proof}
We first start with Lemma~\ref{lemma:cloud_recursion} and sum over $h=1$ to $H$
\begin{align*}
    \sum_{h=1}^H \cQ_h \leq& \gamma^2\alpha^2\beta^2(\cO(H K_c^2 Z^2 E^2 V_2) + \beta^2 \cO(H K_c^2 K_g^2 E^2 Z^2 V_2)) \\
    &\quad+2\alpha^2 K_c^2\sum_{h=1}^H\cQ_{h}\\  
    \sum_{h=1}^H \cQ_h\leq & (1- 2\alpha^2 K_c^2)^{-1}\gamma^2\alpha^2\beta^2(\cO(H K_c^2 Z^2 E^2 V_2) \\
    &\quad  + \beta^2 \cO(H K_c^2 K_g^2 E^2 Z^2 V_2))
\end{align*}
Here, summing over the recursion, we need to control the coefficient of the sum on RHS. If this coefficient is $<1$, then we are done, which is exactly the case when $\alpha = \cO(K_c^{-3/2})$ as $\alpha^2 K_c^2 = \cO(K_c^{-1})$ which can be made to be a constant less than $1$.

Now, for the sum over square roots, we first use Jensen's inequality to obtain, $\sqrt{\sum_{i=1}^n a_i} \leq \sum_{i=1}^n\sqrt{a_i},\forall a_i \geq 0$ on Lemma~\ref{lemma:cloud_recursion}.

\begin{align*}
\sqrt{\cQ_h} &\leq \gamma\alpha\beta(\cO(K_c Z E \sqrt{V_2}) + \beta \cO(K_c K_g E Z \sqrt{V_2})) \\
    & \quad +\alpha \sqrt{K_c}\sum_{r=1}^{K_c  - 1}\sqrt{\cQ_{h-r}}\\
\sum_{h=1}^H \sqrt{\cQ_h} &\leq \gamma\alpha\beta(\cO(H K_c Z E \sqrt{V_2}) + \beta \cO(H K_c K_g E Z \sqrt{V_2})) \\
    & \quad +\alpha K_c^{3/2}\sum_{h=1}^{H}\sqrt{\cQ_{h}}\\
    &\leq(1-\alpha K_c^{3/2})^{-1}\gamma\alpha\beta(\cO(H K_c Z E \sqrt{V_2}) \\
    &\quad + \beta \cO(H K_c K_g E Z \sqrt{V_2}))\\
    &\leq\gamma\alpha\beta(\cO(H K_c Z E \sqrt{V_2}) + \beta \cO(H K_c K_g E Z \sqrt{V_2}))
\end{align*}
Now, we again control the coefficient of the sum on RHS and for $\alpha = \cO(K_c^{-3/2})$, this coefficient is a constant.
\end{proof}

\subsection{Completing the proof}
The all-step progress for the cloud, after $H$ steps,  is our Theorem~\ref{thm:async_fl_1}. We now provide its proof.

% \begin{theorem}[Extended Theorem~\ref{thm:async_fl_1}]
% %\label{thm:async_fl_extended}
% For $L$-smooth and $\mu$-weakly convex loss function $\ell$, under Assumptions~\ref{assump:grad_bdd}-\ref{assump:rho_large}, with $\gamma \leq L^{-1},\alpha \leq K_c^{-3/2}$ and $\beta \leq K_g^{-3/2}$, after running Algorithm~\ref{alg:async_fl} for $H,Z$ and $E$ cloud, gateway and device steps, we obtain 
% \begin{equation}
%     \min_{h=0}^{H-1}\E[\norm{\nabla L_N(\bm{\omega}_h)}^2] \leq \frac{\E[L_N(\bm{\omega}_0) - L_N(\bm{\omega}_{H})]}{\alpha\beta\gamma c H Z E}   + \Xi
% \end{equation}
% where 
% \begin{align*}
%     \Xi &=  \gamma \cO(\rho  E^2  V_2) + \beta \cO(K_g\sqrt{V_1 V_2})\nonumber\\
%     &\quad + \gamma\beta^2 \cO((L + \rho)K_g^2  E  V_2)\\
%     &\quad+ \rho\gamma\beta (\cO(Z  E V_2) + \beta^2 \cO(K_g^2 E Z V_2))\\
%     &\quad +\alpha (\cO(H K_c Z E \sqrt{V_1V_2}) + \beta \cO( K_c K_g  \sqrt{V_1 V_2}))\\
%     &\quad + \cO(\rho + L) \gamma\alpha^2\beta(\cO( K_c^2 Z E V_2)\\
%     &\quad + \beta^2 \cO(K_c^2 K_g^2 E Z V_2))
% \end{align*}
% \end{theorem}
We first sum Lemma~\ref{lemma:cloud_1step} over $h=1$ to $H$, to obtain,
\begin{align}
        \E[L_N(\bm{\omega}_{H}) - L_N(\bm{\omega}_{0})] &\leq  -\gamma\alpha\beta c \sum_{h=1}^H\sum_{z=0}^{Z-1}\sum_{e=0}^{E-1}\E[\norm{\nabla L_N(\bm{\omega}_{\tau,\zeta,e})}^2] \nonumber \\
        & \quad + \sum_{h=1}^H\Xi_h \label{eq:cloud_allstep_int}
\end{align}
We first compute $\sum_{h=1}^H \Xi_h$ using Lemma~\ref{lemma:cloud_sum_recursion} to bound the sum of recursion terms in $\cQ_h$ and $\sqrt{\cQ_h}$.

\begin{align*}
    \sum_{h=1}^H \Xi_h &= \alpha\beta\gamma^2\cO(\rho H E^3 Z V_2) + \alpha\gamma\beta^2\cO(K_g H E Z\sqrt{V_1 V_2}) \\
    & \quad + \alpha\gamma^2\beta^3 \cO((L + \rho)K_g^2 H E^2 Z V_2)\\
    & \quad+ \alpha\rho\gamma^2\beta^2(\cO(Z^2 H E^2 V_2) + \beta^2 \cO(K_g^2 H E^2 Z^2 V_2)) + \cO(\alpha\sqrt{V_1}) \sum_{h=1}^H\sqrt{\cQ_h} \\
    & \quad + \alpha\cO(\rho + L)\sum_{h=1}^H\cQ_h\\
    &\leq \alpha\beta\gamma^2\cO(\rho H E^3 Z V_2) + \alpha\gamma\beta^2\cO(K_g H E Z\sqrt{V_1 V_2}) \\
    & \quad + \alpha\gamma^2\beta^3 \cO((L + \rho)K_g^2 H E^2 Z V_2)\\
    & \quad + \alpha\rho\gamma^2\beta^2(\cO(Z^2 H E^2 V_2) + \beta^2 \cO(K_g^2 H E^2 Z^2 V_2)) \\
    & \quad +\gamma\alpha^2\beta(\cO(H K_c Z E \sqrt{V_1V_2}) + \beta \cO(H K_c K_g E Z \sqrt{V_1 V_2}))\\
    & \quad + \cO(\rho + L) \gamma^2\alpha^3\beta^2(\cO(H K_c^2 Z^2 E^2 V_2) \\
    & \quad + \beta^2 \cO(H K_c^2 K_g^2 E^2 Z^2 V_2))
\end{align*}

Now, consider the following term,
\begin{align*}
    \min_{h=0}^{H-1}\E[\norm{\nabla L_N(\bm{\omega}_h)}^2] \leq& \frac{1}{H}\sum_{h=0}^{H-1}\E[\norm{\nabla L_N(\bm{\omega}_h)}^2]\\
    \leq& \frac{1}{H Z}\sum_{h=0}^{H-1}\sum_{z=0}^{Z-1}\E[\norm{\nabla L_N(\bm{\omega}_{\tau,z})}^2]\\
    \leq& \frac{1}{H Z E}\sum_{h=0}^{H-1}\sum_{z=0}^{Z-1}\sum_{e=0}^{E -1}\E[\norm{\nabla L_N(\bm{\omega}_{\tau,\zeta,e})}^2]
\end{align*}
We obtained the above inequality using the fact that $\min_{i \in [n]} a_i \leq \frac{1}{n}\sum_{i=1}^n a_i$ and using convexity of $\norm{\cdot}$. We can now use the results from Eq.~\eqref{eq:cloud_allstep_int} to obtain 
\begin{align}
    \min_{h=0}^{H-1}\E[\norm{\nabla L_N(\bm{\omega}_h)}^2] \leq& \frac{\E[L_N(\bm{\omega}_0) - L_N(\bm{\omega}_{H})]}{\alpha\beta\gamma c H Z E}  + \frac{\sum_{h=1}^H\Xi_h}{\alpha \beta \gamma c H Z E}
\end{align}
We set the constant term to be $\Xi$.
\end{document}